\documentclass[manuscript,screen]{arxiv_acmart}

\usepackage{times}
\usepackage{epsfig}
\usepackage{graphicx}
\usepackage{amsmath}
  
\usepackage{amssymb}
\usepackage{array}
\usepackage{tabularx}
\usepackage{threeparttable}
\usepackage{multirow}
\usepackage{makecell}
\usepackage{colortbl}
\usepackage{hhline}
\usepackage{booktabs}
\AtBeginDocument{%
  \providecommand\BibTeX{{%
    \normalfont B\kern-0.5em{\scshape i\kern-0.25em b}\kern-0.8em\TeX}}}

\setcopyright{acmcopyright}
\copyrightyear{2022}
\acmYear{2022}
\acmDOI{XXXXXXX.XXXXXXX}

\acmPrice{15.00}
\acmISBN{978-1-4503-XXXX-X/18/06}

\begin{document}

\title{Heterogeneous Federated Learning: State-of-the-art and Research Challenges}


\author{Mang Ye}
\email{yemang@whu.edu.cn}
\author{Xiuwen Fang}
\email{fangxiuwen@whu.edu.cn}
\author{Bo Du}
\email{dubo@whu.edu.cn}
\affiliation{%
  \institution{School of Computer Science, Wuhan University}
  \city{Wuhan}
  \country{China}
}

\author{Pong C. Yuen}
\email{pcyuen@comp.hkbu.edu.hk}
\affiliation{%
  \institution{Department of Computer Science, Hong Kong Baptist University}
  \city{Hong Kong}
  \country{China}
}
  \author{Dacheng Tao}
\email{dacheng.tao@gmail.com}
\affiliation{%
  \institution{The University of Sydney}
  \country{Australia}
}

\renewcommand{\shortauthors}{Ye and Fang, et al.}

\begin{abstract}
Federated learning (FL) has drawn increasing attention owing to its potential use in large-scale industrial applications. Existing federated learning works mainly focus on model homogeneous settings. However, practical federated learning typically faces the heterogeneity of data distributions, model architectures, network environments, and hardware devices among participant clients. Heterogeneous Federated Learning (HFL) is much more challenging, and corresponding solutions are diverse and complex. Therefore, a systematic survey on this topic about the research challenges and state-of-the-art is essential. In this survey, we firstly summarize the various research challenges in HFL from five aspects: statistical heterogeneity, model heterogeneity, communication heterogeneity, device heterogeneity, and additional challenges. In addition, recent advances in HFL are reviewed and a new taxonomy of existing HFL methods is proposed with an in-depth analysis of their pros and cons. We classify existing methods from three different levels according to the HFL procedure: data-level, model-level, and server-level. Finally, several critical and promising future research directions in HFL are discussed, which may facilitate further developments in this field. A periodically updated collection on HFL is available at \url{https://github.com/marswhu/HFL_Survey}.\end{abstract}



\maketitle

\section{Introduction}
With the popularization of smartphones, wearable devices, mobile networks, etc., edge devices have become ubiquitous in modern society. An effective method to better utilize the abundant private data in edge devices without compromising privacy is federated learning, which aims to collaboratively train machine learning models while keeping the data decentralized \cite{tist2019fmlconcept}, {\it i.e.}, following the data-stay-local policy. The participating devices in the federated learning system are regarded as different clients. Federated learning is a distributed machine learning framework with secure encryption technology, which enables multiple institutions to conduct joint machine learning modeling under the requirement of protecting data privacy \cite{nature2022genomics}. Federated learning has drawn increasing attention in both academia and industry owing to its potential utility in large-scale applications \cite{cvpr2022rhfl,cvpr2022fccl,tnnls2022fedpad}, which has been widely explored in the fields of healthcare \cite{nature2021exam,nature2022fedkd,npj2021privacy}, medical analysis \cite{nature2021primia, nature2021famhe}, and data security \cite{nature2021adversarial}, etc.

\begin{figure}[t]
\centering
   \includegraphics[width=\textwidth]{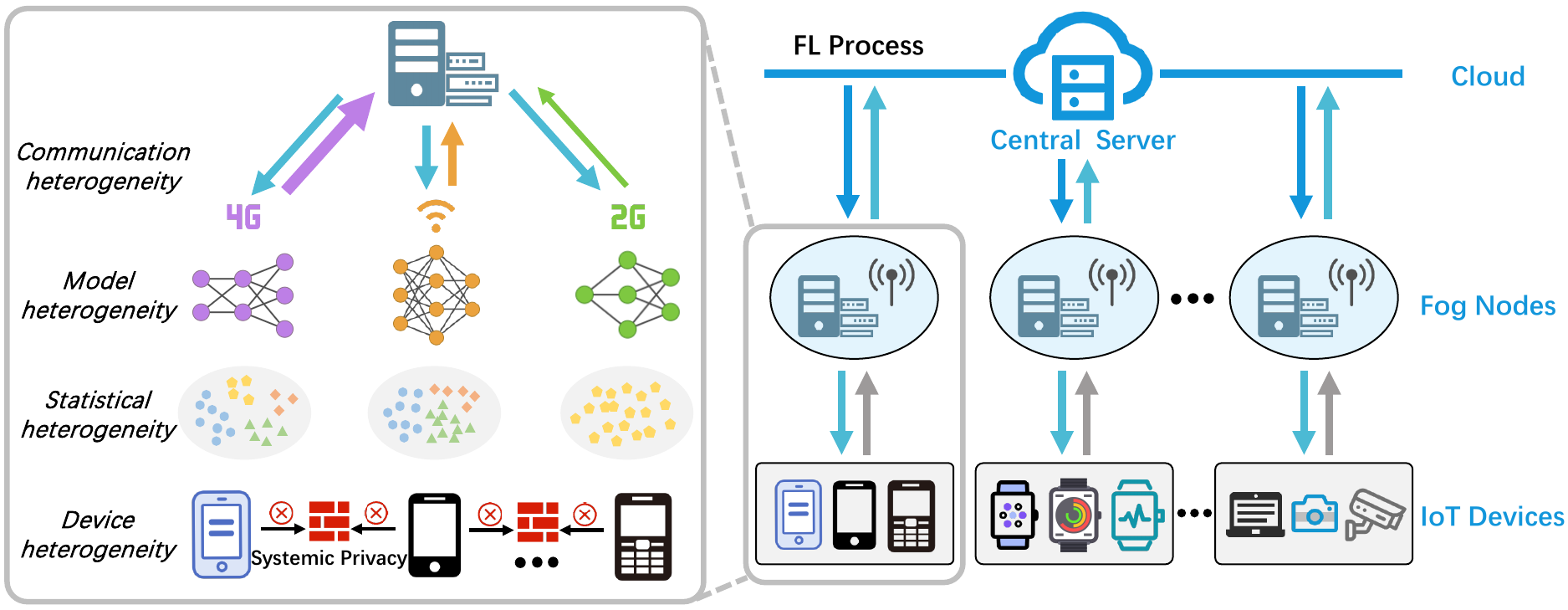}
   \caption{\small{\textcolor{black}{Schematic of heterogeneous federated learning. The figure includes cloud, fog nodes, and IoT edge devices, which constitute the multi-layer heterogeneous federated learning framework. Different devices represent participating clients with heterogeneous local models and Non-IID private data. Each client cannot access the private data from others. The clients upload model-related information to the server, and the server aggregates and broadcasts the knowledge.}}}
\label{fig:intro}
\end{figure}

Despite the great success in homogeneous federated learning, where it heavily relies on the assumption that all the participants share the same network structure and possess similar data distributions \cite{tli2020flchallenges}. However, in practical large-scale situations, there may be considerable differences between data distributions, model structures, communication networks, and system edge devices, which make it challenging to realize federated collaboration. The federated learning associated with these situations is denoted as heterogeneous federated learning, where this heterogeneity can be categorized into four classes according to the federated learning process: statistical heterogeneity, model heterogeneity, communication heterogeneity, and device heterogeneity. These are shown in Fig.~\ref{fig:intro} and detailed as follows. 
1) \textit{Statistical heterogeneity}: the collected data in different participants may be Non-Independent Identically Distributed (Non-IID) \cite{zhu2021noniidflsurvey} or unbalanced, resulting in inconsistent update optimization directions of participants. Original methods would fail in biased collaboration. Therefore, many recent approaches 
\cite{cvpr2022vitfl,icml2022vhl,icml2022fedsam,icml2022pfedbayes} 
attempt to tackle this challenge from various aspects.
2) \textit{Model heterogeneity}: Clients may have different tasks and specific requirements. Consequently, each client may expect to design its local model independently~\cite{wu2020ojcs,icml2022fedhenn}, resulting in knowledge transfer barriers among heterogeneous participants, where the widely-used model aggregation or gradients operation cannot be applied.
3) \textit{Communication heterogeneity}: Considering that clients may be deployed under varying network environments~\cite{tnnls2019robustcom}, this brings in communication inconsistency and unsynchronization, which is also ignored in existing works. This challenge might affect the learning efficiency, especially when the client number is large, which greatly limits the application in large-scale industry scenarios. 
4) \textit{Device heterogeneity}: The storage and computation capabilities of the devices for different participants may be diverse \cite{tli2020flchallenges}, which may cause faults and inactivation of some participating nodes, {\it i.e.} stragglers. There are several methods~\cite{iclr2022splitmix,icml2022bhfl,icml2022knnper} 
developed to deal with this challenge at different stages.
\textcolor{black}{Compared with traditional homogeneous federated learning environments, heterogeneous federated learning has several benefits. First and foremost, it is more flexible and adaptable to the diverse and dynamic scenarios of different clients, which may have inconsistent data distributions, model structures, communication networks, and hardware capabilities. Second, it takes full advantage of the complementary information and knowledge among heterogeneous clients, thus improving learning performance and robustness under complex and uncertain environments. Finally, it can also handle heterogeneous tasks, that is, different clients can pursue different learning tasks or objectives according to their own needs, thus improving the adaptability and flexibility of the model.}

\begin{table}[t]\small
\centering
 \caption{\textcolor{black}{\label{tab:surveys} Related federated learning survey.}}
\begin{tabular}{|m{2.3cm}|m{5.8cm}|m{5.9cm}|} 
  \cline{1-3}
  \makecell[c]{Surveys} & \makecell[c]{Key Contributions} & \makecell[c]{Differences from Our Survey} \\
  \cline{1-3}
  \makecell*[c]{Kairouz \textit{et al.}~\cite{hal2019advances}} & They discuss recent advances in federated learning and provide a survey of open problems and challenges. & This work comprehensively demonstrates the advance in federated learning, but lacks the fine-grained classification of existing methods.\\
  \cline{1-3}
  \makecell*[c]{Li \textit{et al.}~\cite{tli2020flchallenges}} & They discuss the challenges of federated learning from the perspectives of efficiency, heterogeneity and privacy, and list several future directions. & Our survey focuses on the challenges of heterogeneity and provides a more comprehensive and detailed classification of heterogeneity.\\
  \cline{1-3}
  \makecell*[c]{Wahab \textit{et al.}~\cite{wahab2021fml}} & They provide a fine-grained classification scheme of existing challenges and approaches. & Our survey focuses on the challenges of heterogeneity and provides a more comprehensive classification of heterogeneity.\\
  \cline{1-3}
  \makecell*[c]{Li \textit{et al.}~\cite{tkde2021surveyfl}} & They provide a comprehensive analysis on federated learning from systems perspective, including system components, taxonomy, summary, design, and vision. & The taxonomy proposed in this work is not based on a uniform standard.\\
  \cline{1-3}
  \makecell*[c]{Lim \textit{et al.}~\cite{lim2020flmobileedge}} & They survey federated learning in mobile edge networks. & This work is based on federated learning in mobile edge network optimization, while our work investigates from a more general perspective.\\
  \cline{1-3}
  \makecell*[c]{Niknam \textit{et al.}~\cite{commag2020flwireless}} & They discuss the applications and challenges of federated learning in wireless communication environments. & This work discusses applications of federated learning in wireless communication, while we surveys federated learning from a more general perspective.\\
  \cline{1-3}
  \makecell*[c]{Khan \textit{et al.}~\cite{ieee2021iotfladvance}} & They present advances in federated learning for IoT applications and provide a taxonomy using various parameters. & This work explores federated learning for IoT networks and identifies issues of robustness, privacy and communication cost, while our work targets heterogeneity issues\\
  \cline{1-3}
  \makecell*[c]{Nguyen \textit{et al.}~\cite{ieee2021iotflsurvey}} & They survey and analyze the services applications of federated learning in IoT networks. & This work focuses on the characteristics and requirements of IoT networks, rather than covering all possible federated learning scenarios.\\
  \cline{1-3}
  \makecell*[c]{Yang \textit{et al.}~\cite{tist2019fmlconcept}} & They divide federated learning into three categories according to the data distribution characteristics. & This work provides an overview of federated learning but lacks a detailed classification and summary of existing methods.\\
  \cline{1-3}
  \makecell*[c]{Gao \textit{et al.}~\cite{gao2022hflsurvey}} & They investigate the domain of heterogeneous FL in terms of data space, statistical, system, and model heterogeneity. & This work classifies existing methods based on problem settings and learning objectives, while our survey classifies methods based on specific techniques.\\
  \cline{1-3}
  \makecell*[c]{Zhu \textit{et al.}~\cite{zhu2021noniidflsurvey}} & They analyze the impact of non-IID data in federated learning and provide a survey of the research on handling non-IID data. & This work analyzes the impact of Non-IID data on federated learning, but ignores other challenges and related research.\\
  \cline{1-3}
  \makecell*[c]{Kulkarni \textit{et al.}~\cite{kulkarni2020worlds4}} & They point out that statistical heterogeneity can hinder federated learning and highlight the need for personalization. & This work focuses on the challenges posed by statistical heterogeneity while ignoring other issues.\\
  \cline{1-3}
  \makecell*[c]{Tan \textit{et al.}~\cite{tan2022toward}} & They explore the field of personalized federated learning and conduct a taxonomic survey of existing methods. & This work briefly explains statistical heterogeneity, but lacks a comprehensive taxonomy and analysis of the challenges in federated learning.\\
  \cline{1-3}
  \makecell*[c]{Wu \textit{et al.}~\cite{wu2020ojcs}} & They provide a personalized federated learning framework in a cloud-edge architecture for intelligent IoT applications. & This work focuses on personalized federated learning schemes, whereas our survey encompasses broader federated learning schemes.\\
  \cline{1-3}
\end{tabular}
\end{table}

\textcolor{black}{
Several surveys have been released on federated learning in general or specific aspects of federated learning. However, none of them provide a reasonable and comprehensive taxonomy of the research challenges and state-of-the-art of heterogeneous federated learning, which is an important and emerging research direction in federated learning. Heterogeneous federated learning aims to address the heterogeneity issues that arise from different aspects, such as statistical distribution, model architecture, communication setting, and device condition. Therefore, we discuss the main contributions and limitations of existing related surveys and highlight the unique contributions of our work in Tab.~\ref{tab:surveys}. 
Kairouz \textit{et al.}~\cite{hal2019advances} discuss recent advances in federated learning and provide a survey of open problems and challenges, including communication efficiency, privacy preservation, attack defense, and federated fairness.
Li \textit{et al.}~\cite{tli2020flchallenges} discuss the challenges faced by federated learning from four perspectives: communication efficiency, system heterogeneity, statistical heterogeneity, and privacy concern, and briefly listed several future research directions.
Wahab \textit{et al.}~\cite{wahab2021fml} provide a fine-grained classification scheme of existing challenges and approaches.
Li \textit{et al.}~\cite{tkde2021surveyfl} classify federated learning systems from six aspects, including data distribution, machine learning model, privacy mechanism, communication architecture, federated scale, and federated motivation.
Lim \textit{et al.}~\cite{lim2020flmobileedge} study federated learning in mobile edge networks and divided the existing methods into methods that solve the fundamental problems of federated learning and methods that use federated learning to solve edge computing problems.
Niknam \textit{et al.}~\cite{commag2020flwireless} mainly enumerate and discuss several possible applications of federated learning in 5G networks, and described the key issues faced by federated learning in wireless communication settings.
There are some surveys~\cite{ieee2021iotfladvance,ieee2021iotflsurvey} exploring federated learning for IoT Networks.
Khan \textit{et al.}~\cite{ieee2021iotfladvance} present advances in federated learning for IoT applications and provide a taxonomy using various operation modes as parameters (\textit{e.g.}, global aggregation, resource, local learning model, etc.). Besides, they identify important issues (robustness, privacy, and communication cost) and open challenges in federated learning, and propose corresponding guidelines.
Nguyen \textit{et al.}~\cite{ieee2021iotflsurvey} provide a survey and analysis of federated learning in IoT services (\textit{e.g.}, IoT data sharing, data offloading and caching, attack detection, etc.) and IoT applications (\textit{e.g.}, smart healthcare, smart transportation, unmanned aerial vehicles, etc.).
Yang \textit{et al.}~\cite{tist2019fmlconcept} divide federated learning into three categories: horizontal federated learning, vertical federated learning, and federated transfer learning according to the distribution characteristics of data. But they mainly introduce the concept and application of federated learning, lacking a detailed classification and summary of existing methods.
Gao \textit{et al.}~\cite{gao2022hflsurvey} discuss data space, statistics, system, and model heterogeneity in FL, respectively, and provide a classification and introduction of scenarios, goals, and methods under each heterogeneity problem.
Zhu \textit{et al.}~\cite{zhu2021noniidflsurvey} analyze the impact of Non-IID data in federated learning and provide a survey of the researches on handling Non-IID data, but overlook several other heterogeneity issues and related research.
Kulkarni \textit{et al.}~\cite{kulkarni2020worlds4} point out that statistical heterogeneity can deprive high-performance clients of incentives to participate in federated learning, highlight the need for personalization, and surveys work on this topic. They focus on the challenges posed by statistical heterogeneity while ignoring other issues. 
Tan \textit{et al.}~\cite{tan2022toward} explore the field of personalized federated learning, which studies the problem of learning personalized models to handle statistical heterogeneity, and conduct a taxonomic survey of existing methods. But it lacks a comprehensive taxonomy and systematic analysis of the challenges in federated learning. 
Wu \textit{et al.}~\cite{wu2020ojcs} provide a personalized federated learning framework in a cloud-edge architecture for intelligent IoT applications. But their classification of existing methods is not reasonable enough, which is only a small part of heterogeneous federated learning.  In addition, the existing methods are diverse and vary widely in their own settings without a standard setting, making it challenging for readers to keep abreast of advancements in this field.
Consequently, a comprehensive and systematic survey on the research challenges, methods, limitations, and future directions associated with heterogeneous federated learning is urgently needed.}


\begin{figure*}[t]
\centering
   \includegraphics[width=\textwidth]{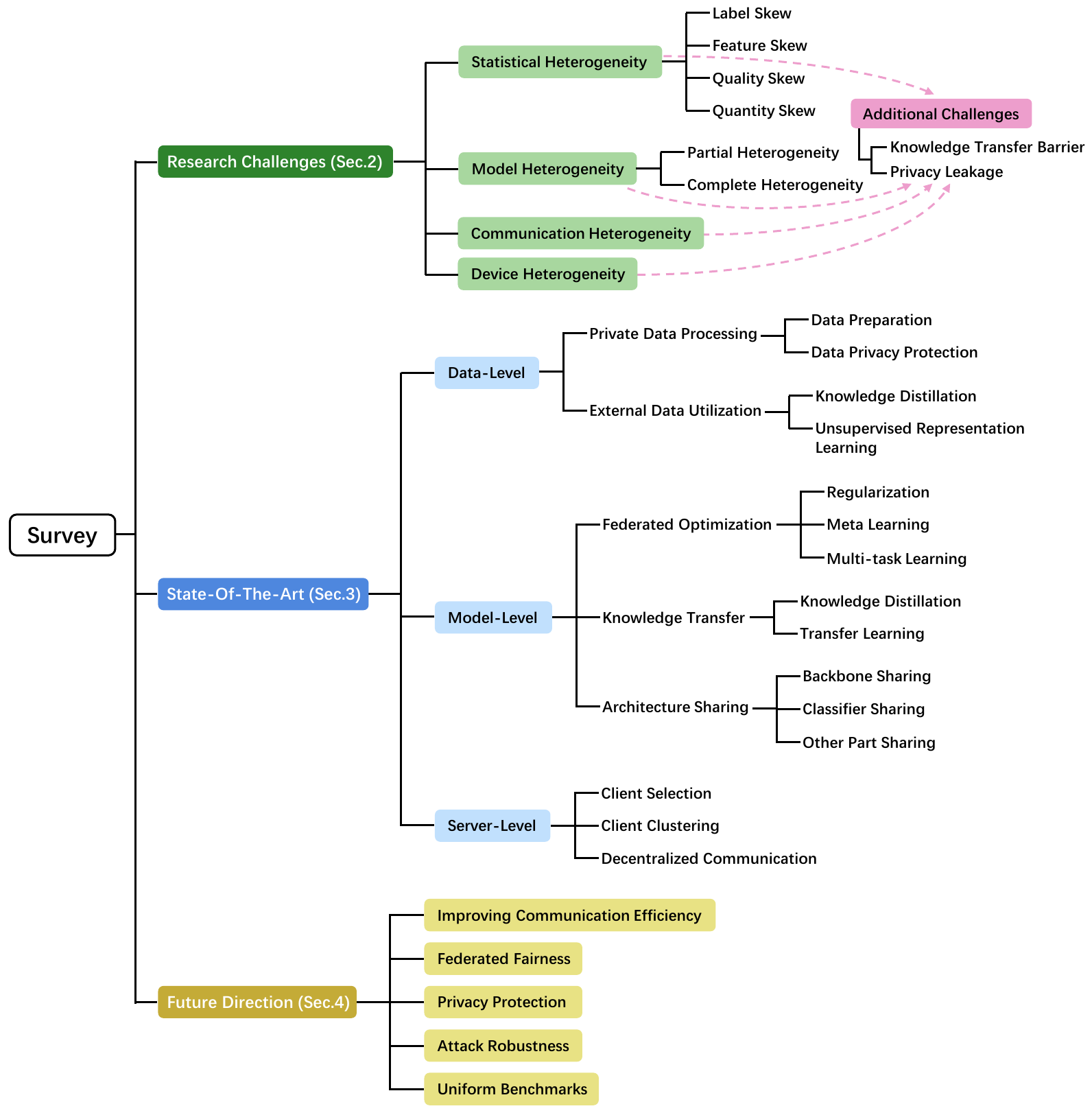}
   \caption{\small{The outline structure of our survey. It contains three different parts: Research challenges, State-of-the-Art, and Discussion of future directions.}}
\label{fig:structure}
\end{figure*}

In this paper, we survey recently published or pre-printed papers on heterogeneous federated learning from top-tier conferences and journals. In particular, we not only investigate the problem of statistical heterogeneity and model heterogeneity but also analyze the aspects of privacy preservation and storage computational capacity during federated communication, which are particularly important for heterogeneous federated learning. Unlike other related surveys \cite{tli2020flchallenges,tan2022toward}, this survey consists of three major parts (Fig.~\ref{fig:structure}). 1) We firstly systematically summarize the research challenges from five different aspects (Section \ref{sec:problems}). 2) We then review the current state-of-the-art methods with in-depth discussions about their advantages and limitations in the context of a new taxonomy (Section \ref{sec:methods}). 3) Finally, we will present a thorough outlook analysis of the unsolved issues and open problems for future development (Section \ref{sec:future}). An outline structure of our survey is illustrated in Fig.~\ref{fig:structure}.

For the research challenges of heterogeneous federated learning,  we focus on the above-mentioned five aspects, {\it i.e.}, statistical heterogeneity, model heterogeneity, communication heterogeneity, device heterogeneity, and additional challenges. Considering that the data distribution of each client may be different, we discuss the Non-IID data from four perspectives: label skew, feature skew, quality skew, and quantity skew. Model heterogeneity is divided into partial heterogeneity and complete heterogeneity, according to the architectural models trained in the federated learning process. \textcolor{black}{Communication heterogeneity refers to the differences in communication resources and environments of clients, which are affected by the bandwidth, reliability, and topology of communication channels. Device heterogeneity is mainly caused by differences in the storage and computational capability of devices. Furthermore, the above four heterogeneous challenges may exacerbate two additional challenges, namely knowledge transfer barriers and privacy leakage. Knowledge transfer barriers indicate difficulties in effectively learning from each other. Privacy leakage refers to the sensitive information of local data sources being exposed to other parties.} By extensively analyzing the research challenges in heterogeneous federated learning, the research priorities in this field can be identified.

To review the state-of-the-art methods, we introduce a new taxonomy to categorize existing heterogeneous federated learning approaches (Fig.~\ref{fig:method}) into three levels, {\it i.e.}, data-level, model-level, and server-level. 
Data-level approaches focus on smoothing the statistical heterogeneity of local data across clients at the data level to support heterogeneous federated learning, such as data augmentation \cite{iclr2021fedmix,corr2018faug,iccd2019astraea}. Model-level methods tend to operate at the model level for heterogeneous federated learning, \textit{e.g.}, sharing partial structures \cite{icml2021fedrep,nips2021ccvr}, model optimization \cite{li2020fedprox,cvpr2021moon,nips2020perfedavg}, knowledge transfer \cite{cvpr2022rhfl,cvpr2022fccl}. The server-level methods require server participation, such as participating client selection \cite{ieee2020favor,corr2020cucb}, or client clustering \cite{ieee2020cfl,acm2020furl,ghosh2020ifca}.
This new taxonomy of existing methods will facilitate the understanding of the state-of-the-art in heterogeneous federated learning, providing further guidelines for our following discussion.

Last but not least, this survey also provides several perspectives to highlight the scope for the future development of heterogeneous federated learning. For example, how to reduce resource consumption and training time while improving model performance in heterogeneous scenarios is a key challenge, that is, we need to improve the efficiency and effectiveness of federated learning by conquering the heterogeneities. Besides, the emphasis on fairness will continue to grow as practical deployments of federated learning expand to more users and enterprises. This aspect is especially important for heterogeneous participants with unequal initial states, as they may have personalized requirements and characteristics.
To ensure the privacy protection of heterogeneous federated learning, it is crucial to establish stricter and more flexible privacy constraint policies, enforcing secure federated communication in all zones. In addition, the robustness of federated systems against attacks and failures requires increasing attention, especially in cases involving heterogeneous models with varying patterns against the attacks. At present, there is a lack of widely recognized benchmark datasets and benchmark testing frameworks for heterogeneous federated learning. This highlights the need to establish systematic evaluation metrics to promote the research on and the development of heterogeneous federated learning.

\section{Problems: Research Challenges in Heterogeneous Federated Learning}\label{sec:problems}

\begin{figure*}[t]
\centering
   \includegraphics[width=\textwidth]{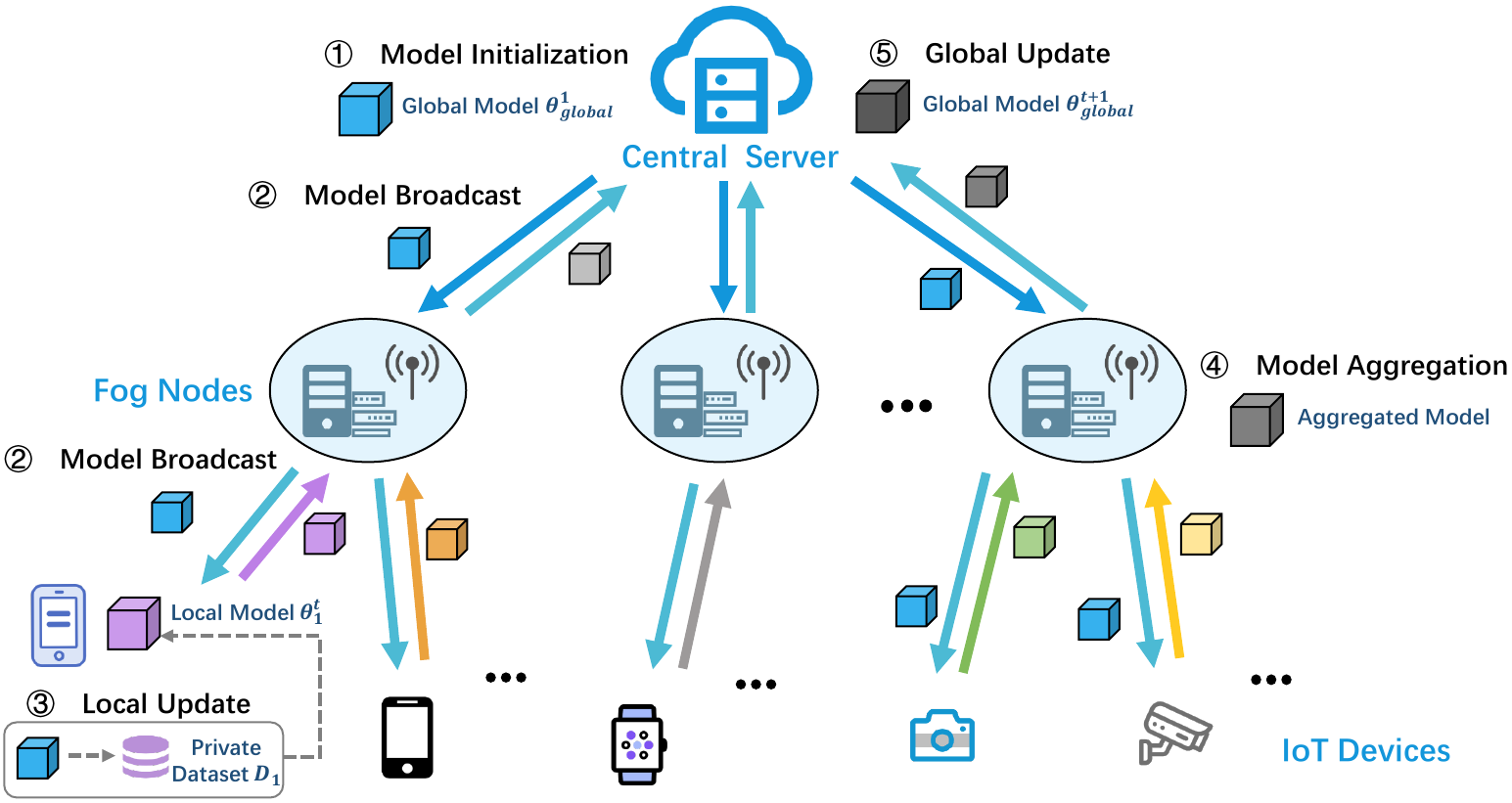}
   \caption{\small{\textcolor{black}{General multi-layer federated learning architecture diagram.}}}
\label{fig:flflow}
\end{figure*}

First, this section provides a formulaic definition of federated learning and illustrates a typical federated learning process. Additionally, we present a detailed taxonomy of the problems encountered in heterogeneous federated learning. 1) Data is the primary element in HFL. Considering that the data distribution may differ across clients, we discuss \textit{statistical heterogeneity} from the four perspectives of label skew, feature skew, quality skew, and quantity skew in \textcolor{black}{Subsection} \ref{sec:prodata}. Besides, this Non-IID phenomenon may hinder subsequent model training. 2) According to the different architectural models trained in the federated learning process, \textit{model heterogeneity} can be divided into partial heterogeneity and complete heterogeneity, as described in \textcolor{black}{Subsection} \ref{sec:promodel}. 3) For the problem of \textit{communication heterogeneity} caused by different network environments such as high communication costs and low communication efficiency, we will discuss this issue in \textcolor{black}{Subsection} \ref{sec:procommunication}. 4) \textcolor{black}{The challenges of \textit{device heterogeneity}, which result from differences in device storage and computation capabilities and may lead to stragglers, fault issues, and low communication efficiency, are discussed in Subsection \ref{sec:prodevice}.} The various forms of heterogeneities introduce additional challenges associated with knowledge transfer barriers and privacy leakage, as presented in \textcolor{black}{Subsection} \ref{sec:proadd}.

\textbf{Preliminaries.} In the typical federated learning framework, we assume that it involves $K$ participating clients $\{C_1,C_2,...,C_K\}$. For each client, the $k$-th client $C_k$ has a private dataset $D_k=\{(x_i^k,y_i^k)\}_{i=1}^{N_k}$ with $|x^k|=N_k$ and $N=\sum_{k=1}^{K}N_k$. Moreover, 
the client $C_K$ usually has a learned local network model or initialized model, denoted by $f(\theta_k)$. Therefore, $f(x^k,\theta_k)$ represents the predicted output of the private sample $x^k$ on the local model $\theta_k$. Traditional centralized machine learning frameworks are typically built on a larger centralized dataset $D_{central}=D_1\cup D_2 \cup...\cup D_K$ by directly integrating the private datasets of each client, which is then used to train a better performing centralized model $\theta_{central}$. However, owing to the constraints of data silos and data privacy, traditional centralized learning cannot be applied in real-world privacy-sensitive scenarios. As an alternative, federated learning enables each client $C_k$ to collaboratively train machine learning models without exposing the private data $D_k$ to other clients $C_{k_0\neq k}$. Here we take FedAvg~\cite{mcmahan2017fedavg} as a typical federated learning process (Fig.\ref{fig:flflow}). In the following, we show the steps involved in federated learning at the $t$-th epoch, in which the federated system iterates the steps for several epochs until the end of the federated training process.
\begin{itemize}
    \item \textbf{Model Initialization.} The server selects eligible clients $\{C_1,C_2,...,C_K\}$ as participants, and initializes the global model $\theta_{global}^1$ in the first round.
    \item \textbf{Model Broadcast.} The server sends the current global model $\theta_{global}^{t-1}$ to all participating clients as the initialization of local models $\{\theta_1^{t-1},\theta_2^{t-1},...,\theta_K^{t-1}\}$.
    \item \textbf{Local Update.} Each participating client $C_k$ utilizes the private dataset $D_k$ for local model updating as follows:
    \begin{equation}
        \theta_k^t\leftarrow \theta_k^{t-1}-\alpha \nabla_\theta \mathcal{L}_k(f(x^k,\theta_k^{t-1}),y^k),
    \end{equation}
    where $\alpha$ represents the learning rate and $\mathcal{L}_k(\cdot)$ denotes the calculated loss for each client $k$.
    \item \textbf{Model Aggregation.} The server calculates the aggregation $\sum\nolimits_{k=1}^{K} \frac{N_k}{N} \theta_k^t$ of the updated client model parameters.
    \item \textbf{Global Update.} The server updates the global model for the next epoch based on the aggregated result, as follows:
    \begin{equation}
        \theta_{global}^{t+1}\leftarrow \sum\nolimits_{k=1}^{K} \frac{N_k}{N} \theta_k^t.
    \end{equation}
    \item \textbf{Model Deployment.} The server distributes the global model to the participating clients.
\end{itemize}

\textcolor{black}{In the cloud-fog-IoT computing environment, federated learning is usually considered as decentralized and multi-layered \cite{ieee2020fogfl}, as illustrated in Fig.\ref{fig:flflow}. The participants will be divided into different layers according to their roles and capabilities, including the cloud layer, fog layers and IoT layers. The cloud layer is the central server that performs global model aggregation and updates~\cite{ieee2022fedfog}. It has high computing and storage capabilities, but the communication cost with edge nodes is also high. The fog layers are intermediate layers composed of multiple edge servers (\textit{e.g.}, base stations) that can communicate with the cloud layer and the IoT layers~\cite{csur2017fogcompute}. The IoT layers are composed of edge devices (\textit{e.g.}, smartphones, sensors), performing local model training and communicating with the fog layers. And the IoT devices are allowed to communicate with their neighbors in a peer-to-peer manner \cite{lim2020flmobileedge}. Compared with the general federated learning process, the cloud-fog-IoT federated learning introduces the middle layer of the fog server between the cloud server and the IoT device~\cite{icc2020hierfavg}. Therefore, there will be an additional step of model distribution and model aggregation in the fog layers, which can relieve the communication pressure between the cloud layer and the IoT layers \cite{ieeeiot2020flfogcompute}. In addition, the cloud-fog-IoT systems have greater flexibility and adaptability in handling various heterogeneities in federated learning.}

\subsection{Statistical Heterogeneity}\label{sec:prodata}
\begin{figure*}[t]
\centering
   \includegraphics[width=\textwidth]{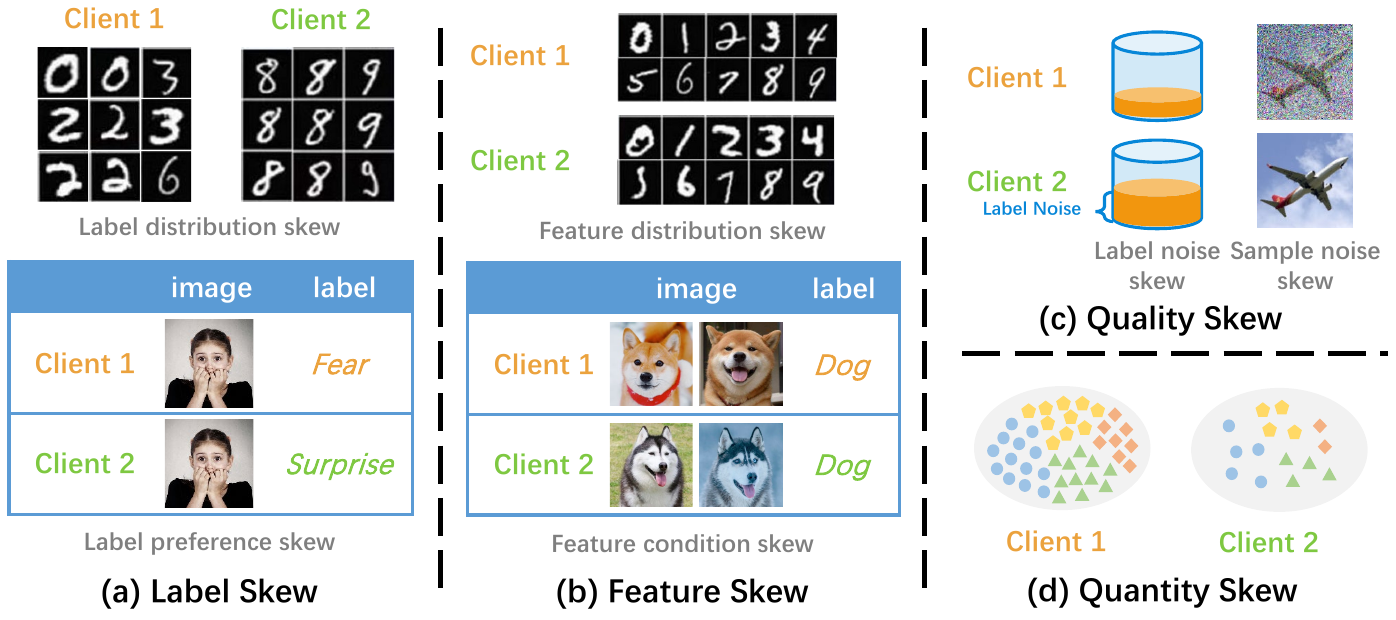}
   \caption{\small{Illustration of four different skew patterns in statistical heterogeneity.}}
\label{fig:statisticalproblem}
\end{figure*}

\textcolor{black}{Statistical heterogeneity refers to the case where the data distribution across clients in federated learning is inconsistent and does not obey the same sampling, {\it i.e.}, Non-IID.} To explore the difficulties of the statistical heterogeneity with the Non-IID phenomenon, we classify statistical heterogeneity from a distribution perspective \cite{hal2019advances,li2021noniidsilos}. Specifically, we distinguish different categories of Non-IID data in terms of four different skew patterns as shown in Fig.~\ref{fig:statisticalproblem}, including label skew, feature skew, quality skew, and quantity skew. We define two different clients $i$ and $j$. Therefore, the local data distribution of client $i$ is denoted as $P_i(x,y)$, where $x$ and $y$ represent the features and labels of the data samples, respectively. \textcolor{black}{Numerous studies~\cite{li2020fedprox,li2021noniidsilos,icml2020scaffold} indicate that the local optimization objectives of the clients are inconsistent with the global optimization objective due to the differences in the local data distribution of the clients. Therefore, statistical heterogeneity may cause local models to converge in different directions, reaching local optima rather than global optima, thus degrading the federated learning performance, which might be even worse than the local learning stage without federated communication.}

{\bf Label Skew.} It means that the label distributions across participating clients are different. 
This phenomenon is commonly encountered in practical applications in which the data collection or annotation is inconsistent. To characterize the various label skew scenarios, we introduce two different settings: \textit{label distribution skew}~\cite{icml2022fedlc} and \textit{label preference skew}~\cite{hal2019advances}. A visual example is illustrated in Fig.~\ref{fig:statisticalproblem}(a).

\textit {Label distribution skew} indicates that the label distributions may be different for different clients, {\it i.e.}, $P_i(y)\neq P_j(y)$, even if the feature distribution is shared \textcolor{black}{(the features of the data samples are similar for each label, regardless of which client they belong to)}, {\it i.e.}, $P_i(x|y)=P_j(x|y)$. 
For example, in handwriting number recognition, different users may contain different numbers.

\textit {Label preference skew} implies that even when the feature distribution is consistent across clients, {\it i.e.}, $P_i(x) = P_j(x)$, the label distribution may be different for different clients, {\it i.e.}, $P_i(y|x)\neq P_j(y|x)$. The local training datasets of different clients may overlap horizontally, leading to label preference skew, {\it i.e.}, the same features have different labels. That is, different clients may annotate the same data samples with different labels owing to individual annotation preferences. 
For example, for a visual intent understanding task~\cite{cvpr2021intentonomy}, the same image may be annotated with different labels according to the personal preferences of users.

{\bf Feature Skew.} It refers to the scenario that the feature distributions across participant clients are different.
This phenomenon often occurs in complex real-world environments, implying that the feature distribution of local data on individual clients may be significantly different \cite{li2021noniidsilos,icml2022dfl}. Feature skew can be divided into the following two settings, which are shown in Fig.~\ref{fig:statisticalproblem}(b):

\textit {Feature distribution skew} implies that the label distribution is consistent, {\it i.e.}, $P_i(y|x)=P_j(y|x)$, but the feature distribution may be different, {\it i.e.}, $P_i(x)\neq P_j(x)$. For instance, in handwriting number recognition, different users may write the same number with different styles, stroke thicknesses, etc.

\textit {Feature condition skew} means that the feature distributions may vary across clients, {\it i.e.}, $P_i(x|y)\neq P_j(x|y)$, even if $P_i(y)=P_j(y)$. The data features across clients may not fully overlap, and this situation is mainly related to vertical federated learning \cite{chen2020vafl}, which is commonly performed in medical applications.
\textcolor{black}{For example, when individual clients link regions, the Japan region contains a large number of Shiba Inu samples, while the Siberia region contains a large number of Husky samples, but their labels are all dogs.}

{\bf Quality Skew.} It demonstrates that the annotation or data collection qualities are inconsistent for different clients.
In general, federated learning typically involves a large number of clients, each of whom may have different data synthesis capabilities. There is no guarantee that all clients have data samples of the same quality, and they may hold unequal proportions of noisy data \cite{yang2020noiselabel}. Therefore, we divide the quality skew into label noise skew and sample noise skew, as shown in Fig.~\ref{fig:statisticalproblem}(c).

\textit {Label noise skew} represents that the proportion of noisy labels contained differs across clients. Owing to the differences in expertise and input costs, the quality of data annotations tends to vary widely across clients, which means that clients contain data with varying degrees of label noise. This challenge intensifies when the architectures of participating clients are different since the decision boundaries are inconsistent.

\textit {Sample noise skew} refers that each client holds private data with different qualities, where the data collection process inevitably introduces varying levels of sample noise. Owing to differences in the abilities of clients to synthesize and collect data, several clients may collect data containing redundant or noisy information, which makes communication across different clients uncertain and complex.

{\bf Quantity Skew.}
It means that the amount of local data may be extremely unbalanced across clients \cite{zhu2021noniidflsurvey}, {\it i.e.}, long-tail distributed data \cite{ijcai2022creff}. An example is illustrated in Fig.~\ref{fig:statisticalproblem}(d). In such a situation, several clients may have problems with data scarcity \cite{acmmm2022fewshot}, which reinforces the need for federated learning. However, existing methods cannot adaptively balance the contribution of each client in the server aggregation process.

\subsection{Model Heterogeneity}\label{sec:promodel}
In the widely-studied federated learning paradigm, each participating client is required to use a local model with the same architecture. Thus the network parameters of the local model can be aggregated into a global model on the server side. In practical IoT applications, each client may expect to design their own local model architecture in a unique manner owing to differences in individual requirements and hardware constraints \cite{wu2020ojcs}. Additionally, clients are often reluctant to reveal or share the details of model design, as they wish to safeguard their commercial proprietary information and privacy. For example, when healthcare institutions conduct collaborative learning without sharing patient information, they may design models with different structures to meet the properties of different tasks. Therefore, model heterogeneous federated learning requires learning knowledge from others without sharing private data or disclosing local model structure information. \textcolor{black}{The main challenge of model heterogeneity would be the difficulty of transferring knowledge between model heterogeneous clients in a model-agnostic manner.} To this end, we categorize model heterogeneity into partial heterogeneity and complete heterogeneity.

{\bf Partial Heterogeneity.}
It is commonly encountered in real-life scenarios, where some of the clients are using the same model structure while others do not. When client subsets are divided based on the local model structure, at least one client subset is expected to have no less than two elements \cite{ieee2020cfl}. In this analysis, we consider the federated system to be a partial model heterogeneity. A federated learning model needs to be trained for each client subset whose models are isomorphic. Through the clustering algorithms, participating clients can be divided into many clusters, {\it i.e.,} the structures are the same within each cluster. \textcolor{black}{Therefore, some common techniques, such as weighted averaging of model parameters, can be directly used to realize the aggregation of intra-cluster models. However, the communication of inter-cluster models requires the design of some special techniques, such as knowledge distillation.}

{\bf Complete Heterogeneity.}
It is a special case of partial heterogeneity, in which all the network structures of participant models are completely different in the federated learning framework. Complete model heterogeneity occurs when the individual clients in a federated system have local model structures that differ from each other \cite{nips2017mocha}. In this context, it is necessary to learn a unique model for each client, which can better handle different data distributions but may eventually lead to high learning costs and low communication efficiency. Ensuring communication between complete heterogeneous models is challenging because the widely-used network parameter aggregation or gradient operations cannot be performed.

\subsection{Communication Heterogeneity}\label{sec:procommunication}
\textcolor{black}{In practical IoT applications, the devices are typically deployed in different network environments and have different network connectivity settings (3G, 4G, 5G, Wi-Fi)~\cite{tli2020flchallenges,pnas2021comeff}, which leads to inconsistent communication bandwidth, latency, and reliability, {\it i.e.}, communication heterogeneity~\cite{icml2023pfedgate}. For example, a central hospital may have a high-speed fiber optic network, while a rural clinic may only have a low-speed wireless network. This leads to the problem of communication heterogeneity. When these medical institutions perform operations such as uploading and downloading with the server, delays, and failures may occur, thereby hindering the effect of federated learning \cite{mcmahan2017fedavg}.}

\textcolor{black}{Communication heterogeneity increases communication cost and complexity to some extent. Considering the differences in the network connectivity settings of IoT devices, different devices may require different amounts of data or time to connect and communicate with the server~\cite{iclr2022fedchain,iclr2022fedpara}. Some devices may connect slowly, rendering them expensive and unreliable to communicate with. Besides, in the training process, there may be offline devices due to network bandwidth and energy constraints. Communication heterogeneity may also reduce communication efficiency and effectiveness~\cite{icml2022dadaquant}. Some IoT devices have problems such as low-quality network environments, slow device connection, and limited network bandwidth. Therefore, the clients may encounter varying degrees of noise, delay, or loss during the communication process, which severely reduces communication efficiency~\cite{icml2022fedcams,icml2022ntk}.} To enhance communication efficiency, stragglers and offline devices with a significant time difference may be discarded after a sufficient number of clients have transmitted their feedback results to the server side. Notably, communication heterogeneity can be viewed as a strategy to address the differences in device computing power \cite{aaai2015hpml} by using bounded-delay assumptions to control devices latencies. \textcolor{black}{Communication heterogeneity is very prevalent in complex IoT environments, which may lead to high-cost and low-efficiency communication, thereby diminishing the effectiveness of federated learning.} Therefore, how to adaptively adjust federated communication in heterogeneous network environments is worth studying.

\subsection{Device Heterogeneity}\label{sec:prodevice}
\textcolor{black}{In practical applications, federated learning networks may involve a large number of participating IoT devices. The differences in device hardware capabilities (CPU, memory, battery life) may lead to different storage and computation capabilities \cite{tli2020flchallenges}, which inevitably lead to device heterogeneity. For example, one client is a smartphone, while another client is a smartwatch. The smartphones have larger storage capacity and stronger computing capacity than smartwatches, which brings device heterogeneity. Therefore, during federated communication, smartwatches may have longer local runtimes and may be dropped or lost, reducing system efficiency and stability.}

In federated learning, clients need to perform local updates and seed the updates back to the server side. However, the participating clients may encounter faults during this process. When the model is updated synchronously, devices with limited computation capabilities may consume a significant amount of time to update the model and may become stragglers. \textcolor{black}{Overall, device heterogeneity poses several challenges to federated learning. First, it leads to system imbalance and inefficiency, as different clients may have different computing speeds or resources, causing system lags or bottlenecks. Second, it introduces uncertainty and instability to the system, since different clients may have different device states, including states that lead to system failure or loss~\cite{tli2020flchallenges}. Therefore, device heterogeneity introduces constraints such as straggler mitigation and fault tolerance into the federated learning process, which necessitates the adaptive adjustment of the feedback for different devices in large-scale federated learning scenarios.}

\subsection{Additional Challenges}\label{sec:proadd}
Apart from the above-mentioned heterogeneities, this part also discusses some additional challenges in heterogeneous federated learning, including knowledge transfer barriers and privacy leakage. 

{\bf Knowledge Transfer Barriers.}
Federated learning is aimed at transferring knowledge between different clients to collaboratively learn models with superior performance. However, the above-mentioned heterogeneous characteristics cause knowledge transfer barriers to different degrees. The client collects data in a Non-IID way, which leads to statistical heterogeneity. Under the combined influence of multiple skew patterns, the domain knowledge of all clients cannot be sufficiently learned. When the clients have models with different structures, the general average parameter strategy cannot be used for model aggregation, resulting in barriers to knowledge exchange between heterogeneous models. Besides, communication heterogeneity and device heterogeneity in the complex Internet of Things (IoT) environments weaken the efficiency of knowledge transfer. Therefore, how to achieve efficient knowledge transfer in heterogeneous scenarios is a problem that current researches need to focus on.

{\bf Privacy Leakage.}
It is now well-understood that privacy is one of the first-order concerns in federated learning \cite{hal2019advances}, \textcolor{black}{because protecting the local data of clients from being leaked is a fundamental principle of federated learning.} In the communication process, each client never shares private data with the server or other clients to ensure basic privacy~\cite{fgcs2021privacysurvey}. \textcolor{black}{However, federated learning by itself cannot guarantee perfect data security, as there are still potential privacy risks or attacks on data privacy.} Moreover, the above-mentioned four types of heterogeneity inevitably exacerbate privacy leakage in different learning stages. \textcolor{black}{For example, when clients implement federated learning by sharing model gradient updates, logits output, etc., attackers can infer the private information of clients by injecting malicious data or models into the system, or by analyzing their model gradients. This may result in the unavoidable leakage of sensitive information to the server or other clients.} 

\textcolor{black}{Several methods have been investigated to enhance the privacy protection of federated learning, mainly containing anonymization, secure aggregation, Differential Privacy(DP), homomorphic encryption, Secure Multi-party Computation(SMC), and so on. Anonymization conceals the identity of the client by using cryptography so that model updates or gradients cannot reveal anything unique to the client. 
Data aggregation enhances the privacy of federated learning by combining data or gradients from multiple clients, reducing the information from a specific client in the shared information \cite{icml2021gaussian}. DP hides real original information by adding noise to the data or gradients before sending them to the server, including local DP \cite{acm2020ldpfed}, hybrid DP \cite{jsa2022hybrid}, shuffle model \cite{aistats2021cldpsgd}, etc. Homomorphic encryption allows a server to perform computations on encrypted data or gradients without decrypting them. Secure multi-party computing is based on the SMC encryption protocol, enabling multiple clients to jointly calculate functions applicable to their private data without sharing the original data.} Typically, they achieve privacy protection at the expense of model performance and communication efficiency. Therefore, how to ensure privacy protection without compromising the model performance is a key challenge in federated communication.

\begin{figure*}[t]
\centering
   \includegraphics[width=\textwidth]{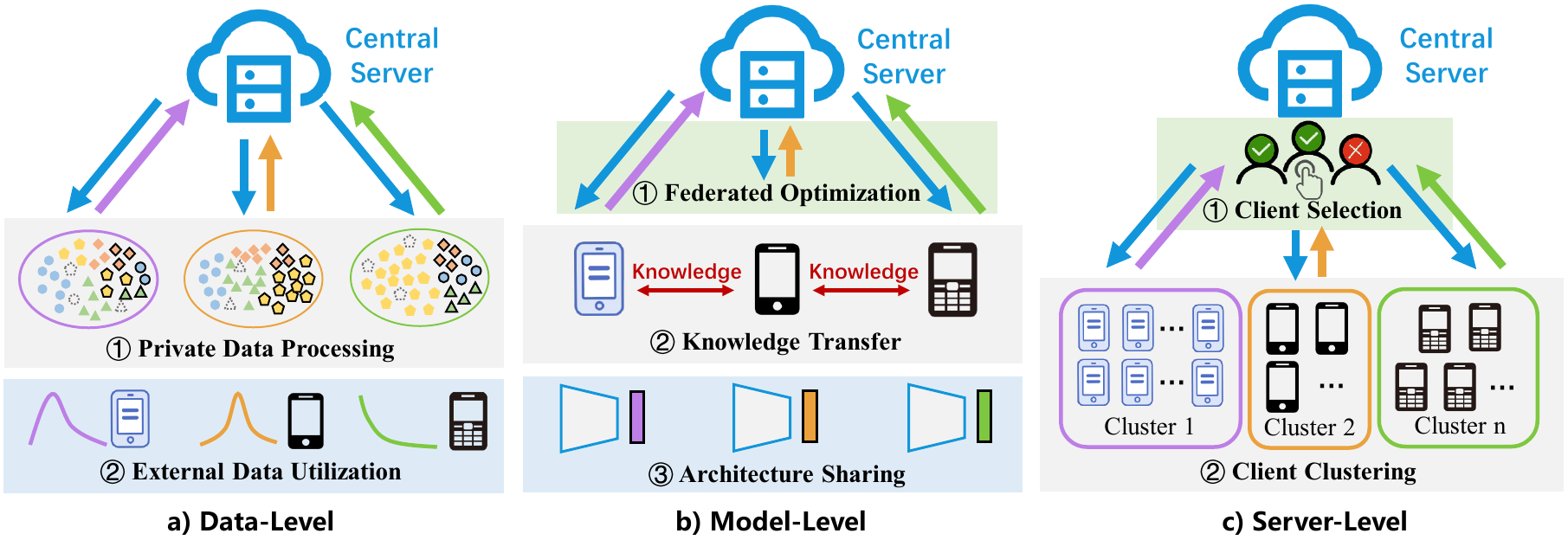}
   \caption{\small{Illustration of the state-of-the-art methods in our taxonomy at three different levels.}}
\label{fig:method}
\end{figure*}

\section{Methods: State-of-the-art}
\label{sec:methods}
This section reviews existing heterogeneous federated learning approaches by dividing them into three parts (Fig. \ref{fig:method}), {\it i.e.}, data-level, model-level, and server-level methods. The data-level methods refer to operations at the data level that smooth the statistical heterogeneity of local data across clients or improve data privacy, such as data augmentation and anonymization techniques. The model-level methods refer to operations designed at the model level, \textit{e.g.}, sharing partial structures, and model optimization. The server-level methods require server engagement, such as participating client selection, or clients clustering.

\subsection{Data-Level Methods}
\textcolor{black}{In this subsection, we introduce the classification of data-level methods and some representative methods in each category, as shown in Tab.~\ref{tab:datamethod}. Data-level methods refer to operations performed at the data level, including private data processing and external data utilization. Private data processing means that clients internally process private data to improve data quality, diversity, and security, thereby optimizing the performance of federated learning. These methods include data preparation and data privacy protection. Data preparation includes operations such as data collection, filtering, cleaning, and augmentation, which can directly alleviate statistical heterogeneity. Data privacy protection aims to ensure that the original data information is not disclosed. External data utilization refers to performing model knowledge distillation or imposing constraints on model updates by introducing additional data. Knowledge distillation is usually employed to deal with communication difficulties caused by model heterogeneity and can alleviate data heterogeneity and communication heterogeneity to some extent. Unsupervised representation learning can alleviate the statistical heterogeneity between the local data of clients.}

\begin{table}[t]\small
\centering
 \caption{\label{tab:datamethod} \textcolor{black}{Data-level methods.}}
\begin{tabular}{|c|c|m{2cm}|m{0.6cm}|m{4cm}|m{3.3cm}|} 
  \cline{1-6}
  \multicolumn{2}{|c|}{Methods} & \makecell[c]{Advantages} & Ref. & \makecell[c]{Key Contributions} & \makecell[c]{Limitations} \\
  \cline{1-6}
  &  &  & \makecell[c]{\cite{ieeetii2022safe}} & Safe detects and filters out poisoned data from attacked devices through a clustering algorithm. & Filtering poisoned data by clustering requires proper selection of attacking clients. \\
  \cline{4-6}
  &  &  & \makecell[c]{\cite{iclr2021fedmix}} & FedMix performs data augmentation based on the MixUp strategy. & Collecting local data distributions may bring potential information leakage. \\
  \cline{4-6}
  &  &  & \makecell[c]{\cite{iccd2019astraea}} & Astraea performs data augmentation based on the global data distribution, generated by collecting the local data distribution. & Uploading local data distributions may expose potential backdoors to attacks.\\
  \cline{4-6}
  &  \multirow{-11}{*}{\makecell[c]{Data \\ Preparation}} & \multirow{-10.5}{0.13\textwidth}{These methods effectively improve the quality and security of private data, and alleviate statistical heterogeneity.} & \makecell[c]{\cite{corr2018faug}} & FAug studies the trade-off between privacy leakage and communication overhead through a GAN-based data augmentation scheme. & Transferring local data to the server violates the privacy requirements of federated learning. \\
  \cline{2-6}
  &  &  & \makecell[c]{\cite{iot2020perfldprivacy}} & PLDP-PFL performs personalized differential privacy according to the sensitivity of private data. & DP can degrade model performance due to its clipping and noise-adding operations\\
  \cline{4-6}
  \multirow{-18}{*}{\makecell[c]{Private \\ Data \\ Processing \\ \ref{method:dataprocess}}} &  \multirow{-4}{*}{\makecell[c]{Data Privacy \\ Protection}} & \multirow{-4}{0.13\textwidth}{These methods ensure the local data privacy.} & \makecell[c]{\cite{choudhury2020anonymizing}} & They use anonymization technology to desensitize local private data. & Anonymization processing may may reduce data quality and availability. \\
  \cline{1-6}
  &  &  & \makecell[c]{\cite{nips2019fedmd}} & FedMD enables clients to independently design their models and implements communication between heterogeneous models. & Global knowledge cannot describe rich domain knowledge in feature skew scenarios. \\
  \cline{4-6}
  &  &  & \makecell[c]{\cite{yu2020salvaging}} & They mitigate overfitting in personalized updates by enhancing the logits similarity between the global model and the local models. & Logits Exploitation may lead to insufficient learning of local information. \\
  \cline{4-6}
  &  &  & \makecell[c]{\cite{fedgkt2020nips}} & FedGKT extracts knowledge on resource-constrained edge devices through knowledge distillation. & Uploading prediction vectors to the server may not satisfy DP guarantees. \\
  \cline{4-6}
  & \multirow{-12}{*}{\makecell[c]{Knowledge \\ Distillation}} & \multirow{-13}{0.14\textwidth}{These methods overcome model heterogeneity to achieve communication between heterogeneous models and effectively decrease communication overhead.} & \makecell[c]{\cite{cvpr2022fedftg}} & FedFTG trains a conditional generator to fit the input space of a local model, and uses it to generate pseudo data. & Training the generator and using FL-div to learn the global knowledge may increase the computation cost. \\
  \cline{2-6}
  &  &  & \makecell[c]{\cite{zhang2020fedca}} & A novel problem named FURL is proposed. And the corresponding solution algorithm, namely FedCA, is designed. & Directly sharing the features of local data may introduce potential privacy risks. \\
  \cline{4-6}
  &  &  & \makecell[c]{\cite{cvpr2021moon}} & MOON solves the Non-IID data issues through model-based contrastive learning. & Storing multiple models simultaneously may require significant additional resources. \\
  \cline{4-6}
  \multirow{-24}{*}{\makecell[c]{External \\ Data \\ Utilization \\ \ref{method:exterdata}}} & \multirow{-8}{*}{\makecell[c]{Unsupervised \\ Representation \\ Learning}} & \multirow{-8.4}{0.14\textwidth}{These approaches enable local models to learn consistent representations while keeping private data decentralized and unlabeled.} & \makecell[c]{\cite{mu2021fedproc}} & FedProc mitigates statistical heterogeneity through prototype-based contrastive learning. & Unprocessed logits transmissions may lead to privacy leakage of local data. \\
  \cline{1-6}
\end{tabular}
\end{table}

\begin{figure}[t]
\centering
   \includegraphics[width=\textwidth]{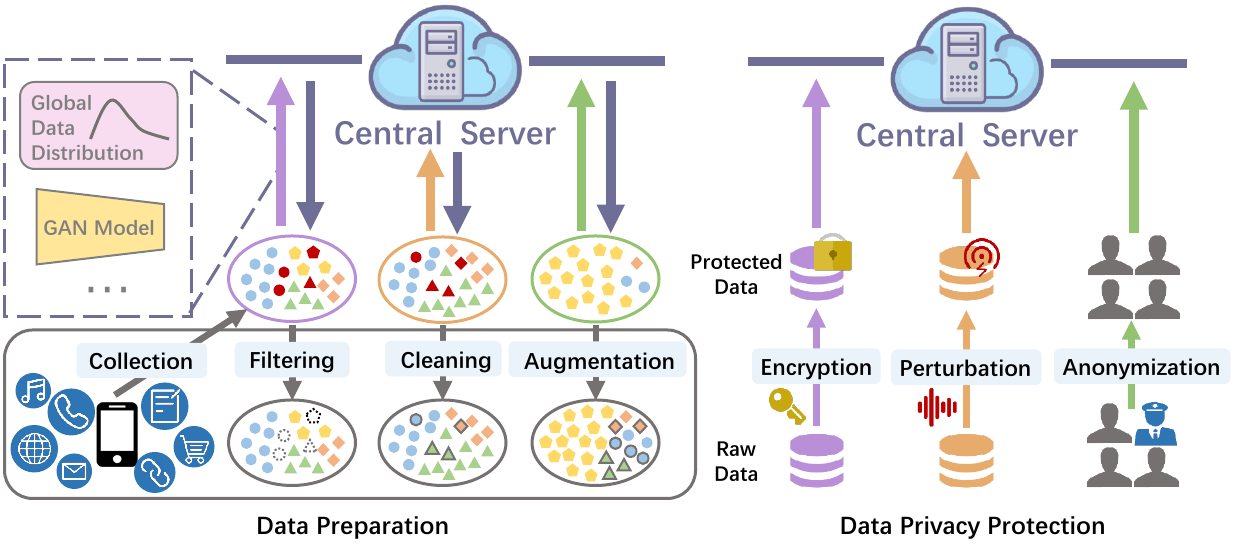}
   \caption{\small{\textcolor{black}{Illustration of private data processing methods in heterogeneous federated learning.}}}
\label{fig:dataprocess}
\end{figure}

\subsubsection{ \textcolor{black}{Private Data Processing}}\label{method:dataprocess}
\

\
\textbf{\textcolor{black}{Data Preparation.}}
\textcolor{black}{Private data preparation in federated learning includes data collection, filtering, cleaning, augmentation, and more (Fig.~\ref{fig:dataprocess}). These operations are aimed at ensuring the data quality and security on each client, thereby improving the efficiency and effectiveness of federated learning. Data collection refers to the process of obtaining data from participating clients. Data filtering refers to the removal of irrelevant data. Data cleaning refers to the correction of inaccurate data. Data augmentation is the process of enhancing or augmenting data with additional information or features, and it has been widely explored in federated learning to address statistical heterogeneity.}

\textcolor{black}{Data collection refers to the process of obtaining data from participating clients. In federated learning, the quantity, quality, and diversity of data collection determine how much useful information the client provides to the federated learning system. Therefore, local data collection is an important part to be optimized in federated learning. Data filtering is the process of removing or excluding irrelevant, noisy or malicious data from federated learning. Therefore, Li \textit{et al.}~\cite{infocom2021sampleselect} consider various data-related factors (error rate, classification distribution, content diversity and data size) that affect model performance to measure the data quality of samples. Then they select the optimal sample combination with the highest total quality under the monetary budget constraint. Besides, data filtering can effectively prevent the federated learning system from being negatively affected by malicious data. For example, Xu \textit{et al.}~\cite{ieeetii2022safe} propose a collaborative data filtering method, Safe, for data selection, which can detect and filter out poisoned data from attacked devices in a federated learning system. Specifically, Safe first clusters the local data, then measures the distance between each sample and its cluster center, and finally discards samples far from the cluster center as poisoned data. Data cleaning is the process of correcting or improving incomplete, inaccurate, or inconsistent data in federated learning, usually by applying relevant techniques locally on the client side to impute missing values, resolve conflicts, and standardize data.}

Data augmentation is a technique of artificially expanding training datasets by generating more data from limited raw data, which can effectively alleviate the problem of data deficiency in deep learning. In addition, popular data augmentation operations include flipping, rotating, scaling, cropping, shifting, Gaussian noise introduction, and MixUp~\cite{iclr2018mixup,shin2020xormixup}. Additional data can also be artificially synthesized using Generative Adversarial Networks (GAN)~\cite{nips2014gan}. However, statistical heterogeneity of client datasets is commonly encountered in federated settings, and private data augmentation techniques can be directly used to smooth the data distribution across multiple clients and mitigate statistical heterogeneity. Federated data augmentation typically requires users to upload a few local data samples, which increases the risk of data privacy breaches. To circumvent this risk, several approaches require a proxy dataset that can represent the overall data distribution of a federated system. Owing to these aspects, data augmentation in federated settings is highly challenging.

In a Non-IID environment with uneven data distributions on the clients, Yoon \textit{et al.}~\cite{iclr2021fedmix} construct a Mean Augmented Federated Learning (MAFL) framework, in which clients exchange mean local data to obtain global information while maintaining privacy requirements. Furthermore, they designed a data augmentation algorithm FedMix under the MAFL framework, which approximates the loss function of the global mixup through Taylor expansion without accessing the raw data of other clients.
Consider the client $i$ has a private local dataset $(x_i,y_i)$, and $f(,)$ is the model output. Therefore, the approximated FedMix loss can be expressed as: 
\begin{equation}\label{eq:fedmix}
\begin{split}
\mathcal{L}_{FedMix} = (1-\lambda)\mathcal{L}(f((1-\lambda)x_i),y_i)+\lambda\mathcal{L}(f((1-\lambda)x_i),\bar{y}_j)+\lambda{\frac{\partial\mathcal{L}}{\partial x}}\bar{x}_j,
\end{split}
\end{equation}
\textcolor{black}{where $\lambda$ represents the mixup rate, and $\bar{x}_j$ and $\bar{y}_j$ refer to the means of all inputs and labels received from client $j$.} However, MAFL may pose a threat to privacy security. Especially when there are insufficient local data, the averaged data contain a large amount of raw relevant information, and adequate privacy restrictions for the raw data cannot be ensured. In Astraea \cite{iccd2019astraea}, the server collects the local data distributions of the clients in the initialization phase and then performs data augmentation based on the global data distribution. To alleviate data distribution imbalance, Astraea creates mediators that rearrange the training of clients based on the KullbackLeibler (KL) divergence of the local data distributions. Federated augmentation (FAug) \cite{corr2018faug} is a data augmentation scheme using GAN. Each client can identify target labels that are lacking in data samples. Subsequently, the clients upload partial data samples of the target labels to the server, and the server oversamples the uploaded data samples to train a conditional GAN. In this manner, the clients effectively enhance the statistical homogeneity of the local data by generating missing data samples using the received GAN.

\textbf{\textcolor{black}{Data Privacy Protection.}}
\textcolor{black}{To ensure that original information about commercial encryption, user privacy, etc. is not leaked to other clients, methods for data privacy protection at the local level have been extensively studied. These methods generally fall into three categories~\cite{csur2021privacyflsurvey}, namely data encryption, perturbation and anonymization (Fig.~\ref{fig:dataprocess}).}

\textcolor{black}{Homomorphic encryption is a commonly used data encryption method that allows computations to be performed using encrypted data without decryption. Asad \textit{et al.}~\cite{appsci2020fedopt,comsec2020homoencryfl} apply homomorphic encryption to federated learning, enabling the client to encrypt its local model with a private key and then send it to the server. So the server can only get encrypted model parameters and cannot deduce any private information. DP is the most commonly used method when using data perturbation to achieve privacy protection. It protects the client's private information by clipping and adding noise to local updates. Hu \textit{et al.}~\cite{iot2020perfldprivacy} propose a federated learning method for personalized local DP, PLDP-PFL. It allows each client to choose an appropriate privacy budget for personalized differential privacy according to the sensitivity of its private data. To alleviate the model performance degradation caused by DP, Shi \textit{et al.}~\cite{cvpr2023dpfedsam} propose a federated learning framework with DP, DP-FedSAM. It leverages the sharpness aware minimization optimizer while updating locally to generate a local flatness model with better stability and robustness to weight perturbations, thus improving its robustness to DP noise. Data anonymization makes it difficult for data subjects to be identified by removing or replacing identifiable sensitive information in the data. Choudhury \textit{et al.}~\cite{choudhury2020anonymizing} make the clients generate their own anonymous data mapping according to the characteristics of the local data set, that is, convert the original data into some random numbers or symbols, so as to desensitize the original data.}

\begin{figure}[t]
\centering
   \includegraphics[width=10.8cm]{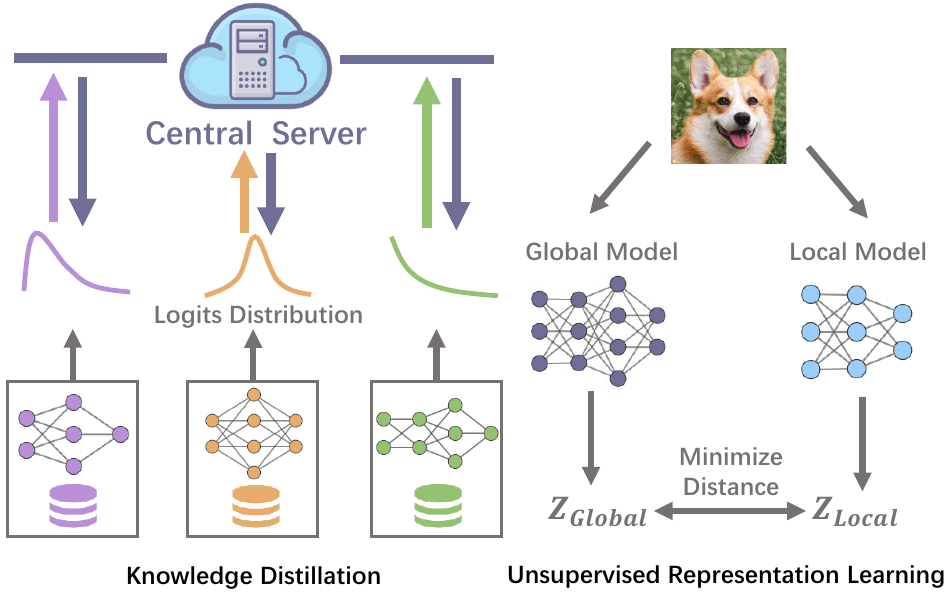}
   \caption{\small{Illustration of external data utilization methods in heterogeneous federated learning.}}
\label{fig:extdata}
\end{figure}

\subsubsection{ External Data Utilization}\label{method:exterdata}
\

\
\textbf{Knowledge Distillation from External Data.}
\textcolor{black}{This method leverages external data sources for knowledge distillation to improve federated learning performance (Fig.~\ref{fig:extdata}), where external data usually refers to public data. The idea is to use a global teacher model trained on an external dataset to help clients generate soft labels for local data. Then, clients utilize these soft labels as an additional supervision for local updates, thereby improving the generalization ability of the model and mitigating the impact of data heterogeneity. Besides, the distillation methods are often used to address model heterogeneity by utilizing external easily accessed data~\cite{cvpr2022fccl,cvpr2022rhfl}. Specifically, each client computes the output prediction distribution of the local model on external data, and sends the output as local knowledge to the server or other clients. And distillation means that other clients update their own local model parameters based on the received knowledge. In this way, clients with heterogeneous models can share local information in a blackbox manner, thereby alleviating the impact of model heterogeneity.} 

FAug \cite{corr2018faug} and FedMD \cite{nips2019fedmd} utilize Federated Distillation (FD), also known as Co-Distillation (CD), to learn knowledge from other clients. Each client stores a local model output and considers the average local model output of all clients as the global output. Huang \textit{et al.}~\cite{acmmm2022fewshot} adopt a federated communication strategy similar to FedMD, and innovatively adds a latent embedding adaptation module, which alleviates the impact of the large domain gap between public dataset and private datasets. Considering the global model as a teacher and the local model as a student, Yu \textit{et al.}~\cite{yu2020salvaging} attempt to mitigate overfitting in personalized updates by enhancing the logits similarity between the global model and the local models. FedGKT \cite{fedgkt2020nips} periodically transfers the knowledge of small CNNs on edges to the large server-side CNN through knowledge distillation, thereby decreasing the burden of edge training. Besides, several recent approaches utilize knowledge distillation to mitigate statistical heterogeneity among clients. FedFTG~\cite{cvpr2022fedftg} trains a conditional generator to fit the input space of a local model, which is then used to generate pseudo data. These pseudo data are input to the global model and the local model for knowledge distillation, and the knowledge of the local model is transferred to the global model by narrowing the Kullback-Leibler divergence between their output predictions. 

\textbf{Unsupervised Representation Learning.}
Since the private data from clients are usually difficult to annotation \cite{ye2020augmentation} and it usually involves huge manual cost, Federated Unsupervised Representation Learning (FURL) \cite{acm2020furl,icml2022orchestra} is discussed to learn a common representation model while keeping private data decentralized and unlabeled (Fig.~\ref{fig:extdata}).

FedCA \cite{zhang2020fedca} is a federated unsupervised representation learning algorithm based on contrastive loss, which can address data distribution inconsistencies and representation misalignment across clients. FedCA includes a dictionary module that aggregates sample representations from all clients and sharing them with clients, and an alignment module that aligns the representations of each client on public data. Briefly, the clients generate local dictionaries through the above two contrastive learning modules, and then the server aggregates the trained local models and integrates the local dictionaries into the global dictionary. Similarly, MOON \cite{cvpr2021moon} and FedProc \cite{mu2021fedproc} also use contrastive learning to address statistical heterogeneity in federated learning. MOON corrects the update direction by introducing a model-contrastive loss. Its objective is to drive the representation learned by the current local model to be consistent with that learned by the global model, while increasing the distance between the representation learned by the current local model and that learned by the previous local model. $z$, $z_p$, and $z_g$ denote the representations from the current local model, previous local model, and global model, respectively. Therefore, the model-contrastive loss can be defined as:
\begin{equation}\label{eq:moon}
\begin{split}
\mathcal{L}_{con} = -\log\frac{\exp(sim(z,z_g)/\tau)}{\exp(sim(z,z_g)/\tau)+\exp(sim(z,z_p)/\tau)},
\end{split}
\end{equation}
where $\tau$ represents a temperature parameter. The main idea of FedProc is to treat the global prototype as global knowledge, and use a local network architecture and a global prototype contrastive loss to constrain the training of the local model. \textcolor{black}{Different from MOON, Tan et al. \cite{aaai2022fedproto} propose FedProto, which only transfers the prototype to the client without transferring the model parameters and gradients. This method can handle various heterogeneous problems more efficiently.}

\subsection{Model-Level Methods}
\textcolor{black}{In this subsection, we categorize model-level methods, introduce representative methods in each category, and discuss their contributions and limitations, as shown in Tab.~\ref{tab:modelmethod}. Model-level methods represent methods for innovative design at the model level, mainly including federated optimization, knowledge transfer across models, and architecture sharing. Federated optimization aims to adapt the model to the local distribution while learning the global information. They can effectively realize local model personalization under statistical heterogeneity. Knowledge transfer across models enables multi-party collaboration in a model-agnostic manner and thus is usually used to solve model and communication heterogeneity. Architecture sharing realizes personalized federated learning by sharing part of the model structure and can solve statistical, model and device heterogeneity simultaneously to some extent. The data-level methods can solve the problem associated with large differences in data distribution to a certain extent and accelerate convergence by smoothing the statistical heterogeneity among data.
In contrast, the model-level methods aim to learn a local model for each client that adapts to its private data distribution, and thus such methods have been extensively researched.}

\begin{table}[h]\small
\vspace{-4mm}
\centering
 \caption{\label{tab:modelmethod} Model-level methods.}
\begin{tabular}{|c|c|m{2.1cm}|m{0.6cm}|m{3.7cm}|m{3.3cm}|} 
  \cline{1-6}
  \multicolumn{2}{|c|}{Methods} & \makecell[c]{Advantages} & \makecell[c]{Ref.} & \makecell[c]{Key Contributions} & \makecell[c]{Limitations} \\
  \cline{1-6}
  &  &  & \makecell[c]{\cite{li2020fedprox}} & FedProx is a federated optimization algorithm that adds a proximal term to FedAvg. & Adding a regularization term may result in slower convergence. \\
  \cline{4-6}
  &  &  & \makecell[c]{\cite{nips2019fedcurv}} & FedCurv uses the EWC algorithm to prevent catastrophic forgetting when transferring tasks. & It may ignore the differences in the extent to which clients are affected by catastrophic forgetting. \\
  \cline{4-6}
  & \multirow{-8}{*}{Regularization}  & \multirow{-8}{0.15\textwidth}{These methods provide convergence guarantee under statistical heterogeneity.} & \makecell[c]{\cite{nips2020pfedme}} & pFedME utilizes the Moreau envelope function as a regularized loss function. & Tuning the regularization parameters may be labour intensive. \\
  \cline{2-6}
  &  &  & \makecell[c]{\cite{Jiang2019fedmaml}} & Similarities between the MAML setting and the personalized objectives of HFL are pointed out. & The two stages of meta-training and meta-testing may introduce additional communication overhead. \\
  \cline{4-6}
  & \multirow{-5}{*}{\makecell*[c]{Meta-\\learning}}  & \multirow{-5.2}{0.15\textwidth}{These methods use meta-learning to achieve the personalized objectives under HFL.} & \makecell[c]{\cite{nips2020perfedavg}} & Per-FedAvg is a personalized variant of FedAvg algorithm based on the MAML formula. & It may not apply to scenarios with significant feature skew. \\
  \cline{2-6}
  &  &  & \makecell[c]{\cite{nips2017mocha}} & A system-aware optimization framework for FMTL is built. & It cannot be applied to non-convex deep learning models. \\
  \cline{4-6}
  \multirow{-20}{*}{\makecell[c]{Federated \\ Optimization \\ \ref{method:potialgorith}}} & \multirow{-4}{*}{\makecell[c]{Multi-task \\ Learning}} & \multirow{-4.3}{0.14\textwidth}{These methods leverage multi-task learning to learn personalized local models.} & \makecell[c]{\cite{icml2021ditto}} & Ditto is a scalable federated multi-task learning framework with two tasks: global goal and local goal. & Tuning the regularization parameters may take much effort. \\
  \cline{1-6}
  & \multirow{-1.6}{*}{\makecell*[c]{Knowledge \\ Distillation}}  & \multirow{-2.4}{0.14\textwidth}{These methods refine knowledge distribution of each client.} & \makecell[c]{\cite{cvpr2022rhfl}} & RHFL uses irrelevant data for knowledge distillation, thereby solving the problem of model heterogeneity. & Using logits output on irrelevant data as local knowledge may underutilize local information. \\
  \cline{2-6}
  &  &  & \makecell[c]{\cite{nips2021ktpfl}} & FT-pFL achieves personalized knowledge transfer via a knowledge coefficient matrix. & The logits output on public dataset may not describe rich local information. \\
  \cline{4-6}
  \multirow{-8}{*}{\makecell[c]{Knowledge \\ Transfer \\ \ref{method:knowtrans}}} & \multirow{-4}{*}{\makecell[c]{Transfer \\ Learning}} & \multirow{-4.5}{0.14\textwidth}{These methods transfer the local knowledge in a model-agnostic manner.} & \makecell[c]{\cite{ieee2020fedhealth}} & FedHealth is a federated transfer learning framework applied in the healthcare domain. & Its generality may be decreased owing to the model heterogeneity in real medical scenarios.  \\
  \cline{1-6}
  
  &  &  & \makecell[c]{\cite{aiatits2020fedper}} & FedPer combines base layers and personalized layers for federated training. & Excessive resource may be consumed owing to the activation of all clients in each round. \\
  \cline{4-6}
  & \multirow{-4}{*}{\makecell[c]{Backbone \\ Sharing}}  & \multirow{-5.4}{0.14\textwidth}{These methods decrease computation costs while satisfying personalized demands.} & \makecell[c]{\cite{icml2021fedrep}} & FedRep enables all clients jointly train global representation learning structure, and then uses private data to train their own heads. & The base layers learn a global representation that may limit the personalization. \\
  \cline{2-6}
  & \multirow{-1.4}{*}{\makecell[c]{Classifier \\ Sharing}}  & \multirow{-2.4}{0.15\textwidth}{These methods leverage the personalized layers to extract features.} & \makecell[c]{\cite{liang2020lgfedavg}} & LG-FedAvg uses personalized layers to extract high-level features and server-shared base layers for classification. & Assuming that all clients have sufficient training data may decrease the generality. \\
  \cline{2-6}
  \multirow{-12}{*}{\makecell[c]{Architecture \\ Sharing \\ \ref{methods:archishare}}} & \multirow{-1.4}{*}{\makecell[c]{Other Part \\ Sharing}} & \multirow{-2.4}{0.14\textwidth}{These methods share part of the model according to local conditions.} & \makecell[c]{\cite{iclr2021heterofl}} & HeteroFL allocates local models of different sizes according to the computational and communication capabilities of each client. & It may not satisfy practical scenarios with high model heterogeneity. \\
  \cline{1-6}
  
\end{tabular}
\end{table}

\subsubsection{Federated Optimization} \label{method:potialgorith}
\

\

\begin{figure}[t]
\centering
   \includegraphics[width=10.6cm]{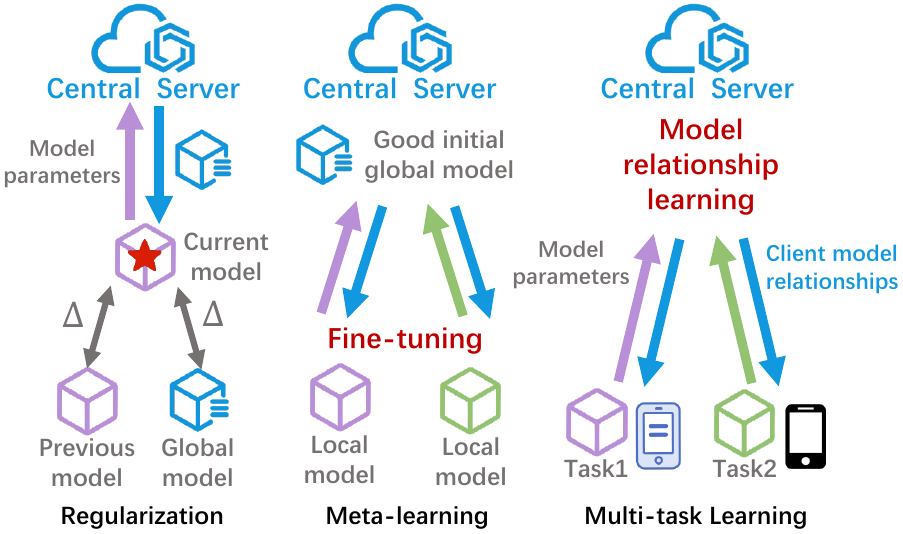}
   \caption{\small{Illustration of federated optimization in heterogeneous federated learning.}}
\label{fig:opti}
\end{figure}

\textbf{Regularization.}
Regularization is a technique that helps prevent overfitting by adding a penalty term to the loss function. This strategy decreases the model complexity by dynamically estimating the parameter values, and decreases variance by adding bias terms (Fig.~\ref{fig:opti}). Therefore, many federated learning frameworks implement regularization to provide convergence guarantees when learning under statistical heterogeneity~\cite{icml2022fedmlb}. 

FedProx \cite{li2020fedprox} adds a proximal term on the basis of FedAvg~\cite{mcmahan2017fedavg}. The server distributes the global model $\theta^t$ of the previous epoch $t$ to clients, and the client $k$ computes the local empirical risk $f_k(\theta)$ on the private dataset. Furthermore, the client $k$ approximately minimizes the objective $h_k$ as follows:
\begin{equation}\label{eq:fedprox}
\begin{split}
\min\limits_{\theta}h_k(\theta,\theta^t) = f_k(\theta)+\frac{\lambda}{2}||\theta-\theta^t||^2,
\end{split}
\end{equation}
where $\lambda$ is the regularization parameter to control the strength of $\theta^t$ to the personalized model, and FedAvg is a special case with $\lambda=0$. The essence of the proximal term is to constrain the difference between the local model and the global model, so as to effectively increase the stability of model training and accelerate model convergence. FedCurv \cite{nips2019fedcurv} and FedCL \cite{icip2020fedcl} are adaptations of the Elastic Weight Consolidation (EWC)~\cite{pnas2017ewc} algorithm to federated learning scenarios. FedCurv uses the EWC algorithm to prevent catastrophic forgetting when transferring to different learning tasks. Its main idea is to selectively penalize certain parameter vectors far from the network parameters learned in the previous task. FedCL adopts the EWC algorithm to estimate the importance weight matrix of the global model and integrates the knowledge of each client into the global model. pFedME \cite{nips2020pfedme} utilizes the Moreau envelope function as a regularized loss function, which decouples the personalized model optimization process from global model learning. Zhang \textit{et al.}~\cite{iclr2021fedfomo} highlight that the clients should consider the suitability of other models to their goals when downloading personalized weighted combinations, and thus devised a personalized federated learning framework named FedFomo. FedAMP \cite{aaai2021fedamp} builds a positive feedback loop, iteratively promotes similar client models to collaborate more strongly than dissimilar client models, and adaptively groups similar clients to promote effective collaboration. FedBN \cite{iclr2021fedbn} solves the problem of feature skew in statistical heterogeneity by adding a batch normalization layer to the local model. SCAFFOLD \cite{icml2020scaffold} uses variance reduction to correct the client-drift caused by statistical heterogeneity. Specifically, SCAFFOLD estimates the update direction of the server model and the local model and then uses the difference between the server model and the local model to correct the local model update. Instead of learning a single global model, Hanzely \textit{et al.}~\cite{hanzely2020l2gd} find a trade-off between the global and local models by adding a regularization term, and learn an implicit mixture model of the global and local models. Unlike these methods, MOON~\cite{cvpr2021moon} considers not only the regularization between the global model and the local model but also the regularization between the current local model and the previous local model. To deal with statistical heterogeneity and to improve training stability, Xu \textit{et al.}~\cite{cvpr2022fedcorr} introduce an adaptive weighted proximal regularization term based on the estimated noise level. 
To address the problem of statistical heterogeneity and device heterogeneity, Pillutla \textit{et al.}~\cite{springerml2023superquantile,ciss2021superquantile,laguel2020superquantile} improves the performance of the worse-off clients while maintaining the average performance using a risk measure known as the superquantile (or CVaR), which is able to capture the tail statistical characteristics of the client error distribution. Huang \textit{et al.}~\cite{cvpr2023pfl} propose FPL to make sample embeddings closer to cluster prototypes of the same domain and category. Meanwhile, a consistency regularization is introduced to align sample embeddings with homogeneous unbiased prototypes that do not contain domain information. Chen \textit{et al.}~\cite{cvpr2023elasticagg} design an elastic aggregation strategy that adaptively interpolates client-side models based on parameter sensitivity, measured by computing the impact of each parameter variation on the output of the overall prediction function. It is an implicit regularization method. In terms of practical technical applications, FedHumor~\cite{tist2022fedhumor} applies federated learning to personalized humor recognition in texts, and considers the distribution of humor preferences of different clients to perform domain adaptive fine-tuning training, achieving personalized federated learning. 

\textbf{Meta-learning.}
This technique is also known as "learning to learn". Previous experience is used to guide the learning of new tasks, thereby allowing a machine to learn a model by itself for different tasks (Fig.~\ref{fig:opti}). Recently, a meta-learning algorithm named Model-Agnostic Meta-Learning (MAML) \cite{icml2017maml} has attracted widespread attention, as it can be directly applied to any method based on gradient descent. In brief, MAML first trains the initialized model. When training on new tasks, a satisfactory learning performance can be achieved by performing fine-tuning based on the initial model with only a small amount of data. In this manner, the personalization ability of meta-learning can solve the problem of statistical heterogeneity in federated learning.

Jiang \textit{et al.}~\cite{Jiang2019fedmaml} point out that the MAML setting is consistent with the personalized objectives of heterogeneous federated learning. The MAML algorithm is divided into two phases: meta-training and meta-testing, corresponding to the global model training and local model personalization in federated learning, respectively. They also observe that the FedAvg algorithm is very similar to the Reptile \cite{nichol2018reptile} algorithm. Careful fine-tuning yields a global model with a high accuracy, and the local models are easy to personalize.  Per-FedAvg~\cite{nips2020perfedavg} is a personalized variant of the FedAvg algorithm based on the MAML formula. Per-FedAvg aims to learn a high-performance initial global model to ensure that each heterogeneous client can obtain a high-performance local model after personalized updating on the global model. Compared with Per-FedAvg, which performs only one-step gradient updates to obtain personalized models, pFedMe \cite{nips2020pfedme} implements multi-step updates. Chen \textit{et al.}~\cite{chen2018fedmeta} propose a federated meta-learning framework, FedMeta. An algorithm is maintained on the server and distributed to the client for training, after which the test results on the query set are uploaded to the server for algorithm update. ARUBA \cite{nips2019aruba} utilizes online convex optimization and sequence prediction algorithms to adaptively learn the task-similarity and test the federated learning performance. To combat the possible vulnerabilities of meta-learning algorithms, a federated meta-learning method named FedML \cite{icdcs2020fedml} is established based on distribution robust optimization (DRO). Zheng \textit{et al.}~\cite{ijcai2021fedmetacredit} apply federated meta-learning to fraudulent credit card detection. This method enables the collaboration between different banks through federated learning, improves the triplet loss function, and designs a meta-learning-based classifier for local model updates. \textcolor{black}{Chu \textit{et al.}~\cite{chu2022multilayerperfl} propose a multi-layer personalized federated learning method, MLPFL, to optimize the inference accuracy of different levels of device grouping criteria. MLPFL trains a personalized model with meta-gradient updates for all groups of edge devices.}

\textbf{Multi-task Learning.}
Multi-task learning enables models learned on a single task to help learn other tasks by using shared representations or models for relevant tasks. If the local model learning for each client is considered as a separate task, the idea of multi-task learning can be implemented to solve the federated learning problem. To this end, multi-task learning aims to solve different tasks on multiple clients simultaneously and train a model that jointly learns the relevant tasks (Fig.~\ref{fig:opti}). All participating clients collaboratively train their local models, thereby effectively mitigating statistical heterogeneity and yielding high-performance personalized local models. 

MOCHA \cite{nips2017mocha} is a system-aware optimization framework for federated multi-task learning(FMTL), which attempts to address the problems of high communication cost, stragglers, and fault tolerance for distributed multi-task learning. To address statistical heterogeneity and system challenges, MOCHA employs the distributed optimization method COCOA~\cite{nips2014cocoa1,icml2015cocoa2}, and trains a unique model for each client. However, MOCHA has some limitations in that it requires all clients to participate in each iteration, which is clearly impractical, and this method cannot be applied to non-convex deep learning models. To ensure that the FMTL algorithm can be applied to general non-convex models, VIRTUAL~\cite{corinzia2019virtual} considers the federation of a central server and clients as a Bayesian network and employs approximated variational inference for training. OFMTL~\cite{bigdata2019ofmtl} simulates the relationship between different devices by introducing an accuracy matrix, through which personalized models for new devices can be inferred without revisiting the original device data. Besides, Dinh \textit{et al.}~\cite{dinh2021fedu} advise a communication centralized FMTL algorithm FedU, which uses Laplacian regularization to capture the relationship between client models. Ditto~\cite{icml2021ditto} is a scalable FMTL framework with two tasks, a global objective and a local objective. This method ensures that the personalized model is close to the global optimization model by introducing a regularization term. Marfoq \textit{et al.}~\cite{nips2021fedem} study FMTL under the assumption that each local data distribution is a mixture of unknown underlying distributions. Therefore, each client can benefit from the knowledge distilled from the local data of other clients by modeling the distributions.

\subsubsection{ Knowledge Transfer across Models} \label{method:knowtrans}
\

\

\begin{figure}[t]
\centering
   \includegraphics[width=10.8cm]{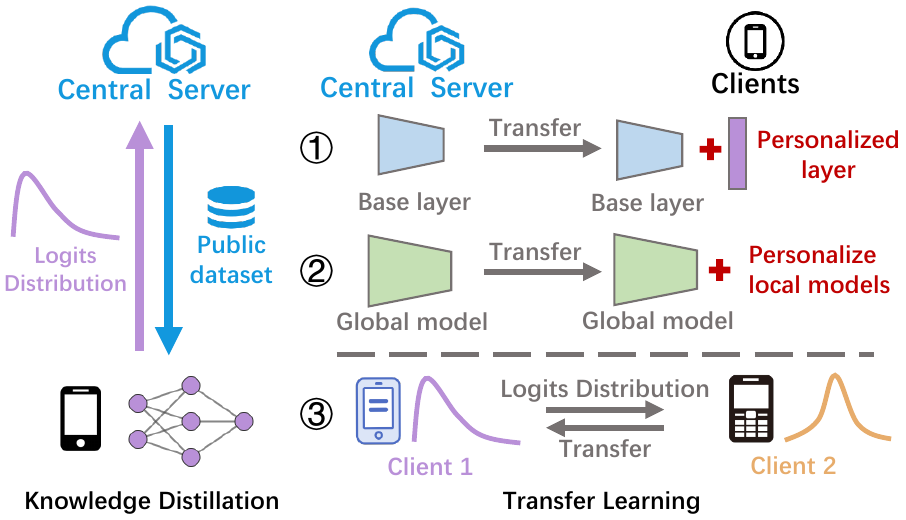}
   \caption{\small{Illustration of  the knowledge transfer approaches across models in heterogeneous federated learning.}}
\label{fig:knowtrans}
\end{figure}

\textbf{Knowledge Distillation across Models.}
In practical federated learning applications, clients may expect to design unique model structures and be reluctant to share the model details. To address the challenges associated with model heterogeneity, knowledge distillation is widely applied in model heterogeneous federated learning. The objective is to refine the knowledge distribution on clients, and then transfer the learned knowledge in a model-agnostic way (Fig.~\ref{fig:knowtrans}). 

FedMD \cite{nips2019fedmd} implements client-side communication by leveraging knowledge distillation and transfer learning. The logits output of the local model $f_k$ on the public dataset $D_0=\{x_i^0\}$ can be considered as the knowledge distribution of the client $k$. The server collects this knowledge to compute the average logits output as a global consensus $\tilde{f}$, which can be expressed as: 
\begin{equation}\label{eq:fedmd}
\begin{split}
\tilde{f}(x_i^0)=\frac{1}{K}\sum\nolimits_{k=1}^{K} f_k(x_i^0),
\end{split}
\end{equation}
Subsequently, each client learns the information from other clients by training its local model to approach the global consensus. The basic ideas of Cronus \cite{chang2019cronus} and FedMD are similar as they both operate on public data via knowledge distillation. However, FedMD calculates a global consensus through a common averaging strategy, whereas Cronus designs an aggregation algorithm that is robust against poisoning attacks. The performance of the aforementioned methods depends heavily on the public data quality. To alleviate the reliance on public data, FedDF \cite{nips2020feddf} utilizes unlabeled or generated data for ensemble distillation. To perform federated learning with heterogeneous clients without relying on a global consensus or shared public models, RHFL~\cite{cvpr2022fccl,cvpr2022rhfl} learns the knowledge distribution of other clients by aligning models feedback on irrelevant public data.
To improve the communication efficiency of federated learning, Sattler \textit{et al.}~\cite{sattler2020cfd} develop a compressed federated distillation method CFD, which employs distilled data curation, soft-label quantization and delta-encoding to reduce the communication from client to server. Unlike these approaches, FedGEN~\cite{icml2021fedgen} does not rely on the server side to possess a proxy dataset. FedGEN performs statistical heterogeneous federated learning through a data-free knowledge distillation approach, in which the server learns a lightweight generator derived only from the prediction rules of client models, thereby integrating client information in a data-free manner. The generator is then broadcast to the clients, using the learned knowledge as an inductive bias to guide local model training.

\textbf{Transfer Learning.}
The goal of transfer learning is to apply the knowledge learned on the source domain to different but related target domains (Fig.~\ref{fig:knowtrans}). In the federated learning scenarios, the clients typically belong to different but related domains and wish to learn knowledge from other domains. Therefore, transfer learning is widely applied in the federated learning field. Knowledge distillation is an effective strategy for transfer learning. Accordingly, federated transfer learning aims to transfer the knowledge learned on clients to the public server for aggregation or to transfer the global consensus to clients for personalization.

In the healthcare domain, FedHealth~\cite{ieee2020fedhealth} is established as a federated transfer learning framework. FedHealth performs data aggregation through federated learning and then builds personalized models through transfer learning. Considering the large distribution differences between the server model and the client models, FedHealth allows clients to train personalized models through transfer learning. In FedMD, transfer learning is a key approach to address the problem of private data scarcity and personalize the local models. In the transfer learning phase, the clients first fully train the local models on the public dataset and then train them on the private datasets until convergence. To solve the problem of slow convergence and degraded learning performance in heterogeneous scenarios, the decentralized federated learning via mutual knowledge transfer (Def-KT) algorithm is designed by Li \textit{et al.}~\cite{iot2020defkt}. Def-KT does not require the participation of the server, and each client directly transfers the knowledge in a point-to-point manner. In the personalized knowledge transfer phase, KT-pFL~\cite{nips2021ktpfl} linearly combines the soft predictions of all clients through the knowledge coefficient matrix to identify the mutual contribution of clients, thereby enhancing the collaboration between clients with similar data distributions. The soft prediction of the local model $\theta_n$ on the public dataset $D_0$ represents the collaborative knowledge $f(\theta_n,D_0)$ from client $n$. Furthermore, KL divergence is utilized to transfer knowledge across clients, and then the local update can be phrased as:
\begin{equation}\label{eq:ktpfl}
\begin{split}
{\theta_n} \leftarrow {\theta_n - \alpha{\nabla_\theta} \mathcal{L}_{kl}(\sum_{m=1}^{N} \mathcal{M}_{m} \cdot f(\theta_m, D_0),f(\theta_n,D_0))},
\end{split}
\end{equation}
where $D_0$ represents a mini-batch of the public dataset, $\alpha$ is the learning rate, and $\mathcal{M}_{m}$ denotes the knowledge coefficient vector of client $m$. Gao \textit{et al.}~\cite{bigdata2019hftl} design a privacy-preserving transfer learning method to remove covariate shifts in homogeneous feature spaces and bridge heterogeneous feature spaces of different clients. Unlike the above methods that transfer the entire global model to the client, FedPer~\cite{aiatits2020fedper} trains the model base layers on the server side based on FedAvg, and then transfers the globally shared base layers to the client. The clients train the model personalization layers on local data with stochastic gradient descent, thereby mitigating the impact of statistical heterogeneity.

\begin{figure}[t]
\centering
   \includegraphics[width=10.6cm]{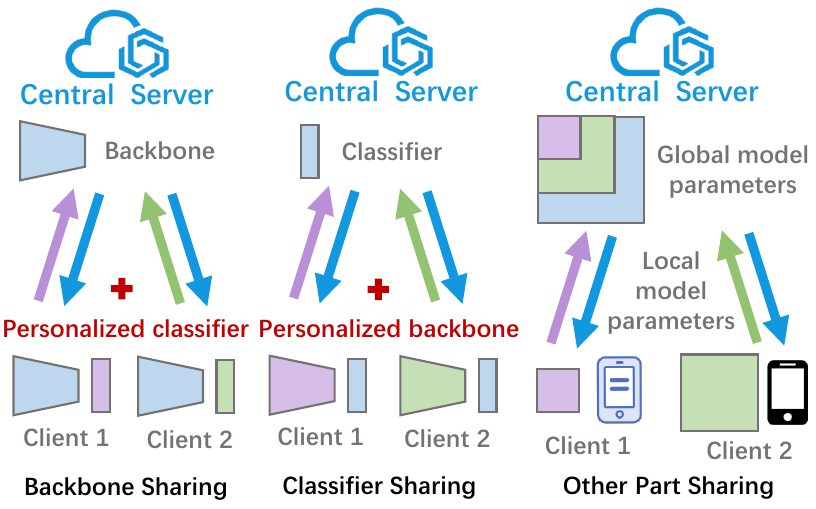}
   \caption{\small{Illustration of the architecture sharing approaches in heterogeneous federated learning.}}
\label{fig:archishare}
\end{figure}

\subsubsection{Architecture Sharing} \label{methods:archishare}
\

\

\textbf{Backbone Sharing.}
In heterogeneous scenarios, the private datasets of clients may be Non-IID. To mitigate the negative effects caused by statistical heterogeneity, clients may share the backbone, but they design personalized layers in the neural network models for personalized demands (Fig.~\ref{fig:archishare}). Furthermore, the clients only need to upload the backbone to the server-side in the aggregation phase, thereby decreasing the communication cost to some extent.

For example, the aforementioned FedPer \cite{aiatits2020fedper}, which combines the base layers and the personalized layers for federated training of deep feedforward neural networks, can effectively capture the personalization aspects of clients. Intuitively, FedPer first uses a FedAvg-based approach to globally train the base layers on the public dataset. Subsequently, each client updates the personalized layers with the private dataset using an SGD-style algorithm. The base layers consist of shallow neural networks for high-level feature extraction, and the personalized layers are deep neural networks for classification. This framework can avoid the problem of retraining in federated transfer learning. In the training process of federated representation learning (FedRep) \cite{icml2021fedrep}, all clients jointly train the global representation learning structure, and then use their private datasets to train their own client-special heads. Here the heads of clients represent personalized, low-dimensional classifiers. Classifier Calibration with Virtual Representations (CCVR) \cite{nips2021ccvr} generates approximate Gaussian mixture model (GMM) based virtual representations in feature space via learned feature extractors. To mitigate the problem that the classifier can be easily biased to the heterogeneous local data, CCVR eliminates the bias of the classifier by regularizing or calibrating the classifier weights. To further consider the problem of long-tail distribution, Classifier Re-training with Federated Features (CReFF) \cite{ijcai2022creff} learns federated features to re-train the classifier, which approximates training the classifier on real data.

\textbf{Classifier Sharing.}
To handle heterogeneous data and tasks, several methods share a classifier instead of a backbone (Fig.~\ref{fig:archishare}). Intuitively, the clients perform feature extraction through their own backbones and share a public classifier for classification. In this way, clients can learn from each other while satisfying personalization and without compromising their data privacy or task specificity.

The recently devised LG-FedAvg \cite{liang2020lgfedavg} jointly learns compact local representations on all clients and a global model across all clients. In contrast to FedPer, LG-FedAvg uses personalized layers to extract high-level, compact features that are important for prediction and uses the base layers shared by the server for classification. In LG-FedAvg, the client $k$ possesses a private dataset $\{(x,y)|x\in X_k,y\in Y_k\}$. Furthermore, the client $k$ extracts the feature $\mathcal{F}_k=l_k(X_k,\theta_k^l)$ of the private data samples through the local model $\theta_k^l$. These features $f\in \mathcal{F}_k$ are predicted by the global model $\theta_k^g$. The overall loss on the client $k$ can be expressed as:
\begin{equation}\label{eq:lgfedavg}
\begin{split}
\mathcal{L}_k(\theta_k^l,\theta_k^g) = \mathbb{E}[-\log \sum_f(p_{\theta_k^g}(y|f),p_{\theta_k^l}(f|x))],
\end{split}
\end{equation}
LG-FedAvg thus effectively decreases the communication cost by extracting useful lower-dimensional representations. Xu~\textit{et al.}~\cite{iclr2023fedpac} propose FedPAC, which reduces inter-client feature variance by constraining each sample feature vector to be close to the global feature centroid of its category. Then, the server performs an optimal weighted aggregation of the personalized classifier headers of the clients.

\textbf{Other Part Sharing.}
\textcolor{black}{Apart from the case of backbone or classifier sharing, several methods employ other part sharing strategies, that is, adaptively share a subset of the local model parameters according to local conditions (\textit{e.g.}, data distribution, computing capability, network bandwidth, privacy preference, etc.) as shown in Fig.~\ref{fig:archishare}. This approach effectively enhances the applicability of federated learning in scenarios involving different resource and network environment constraints across clients, thereby alleviating device heterogeneity and communication heterogeneity.} Furthermore, sharing part of the model for personalization can avoid catastrophic forgetting to some extent~\cite{icml2022fedalt,mccloskey1989cataforget}, which is extremely important for clients with large differences in data distributions.

To alleviate the effect of communication heterogeneity and device heterogeneity, HeteroFL~\cite{iclr2021heterofl} allocates local models of different sizes according to the computing and communication capabilities of each client. The local model parameters are a subset of the global model parameters, which effectively decreases the computation of local clients. Different from the prevailing methods that divide the shared layers and the personalization layers in a layer-wise mechanism, CD$^2$-pFed~\cite{cvpr2022cd2pfed} dynamically decouples the global model parameters for personalization, which is known as channel decoupling. Concretely, a personalization rate is also defined to assign global shared weights and local private weights to each layer of the objective model. \textcolor{black}{FedLA~\cite{cvpr2022pfedla} leverages a hypernetwork on the server to evaluate the importance of each client model layer and generate aggregation weights for each model layer, thereby realizing personalized layer-wised model aggregation.}

\begin{figure}[t]
\centering
   \includegraphics[width=13.6cm]{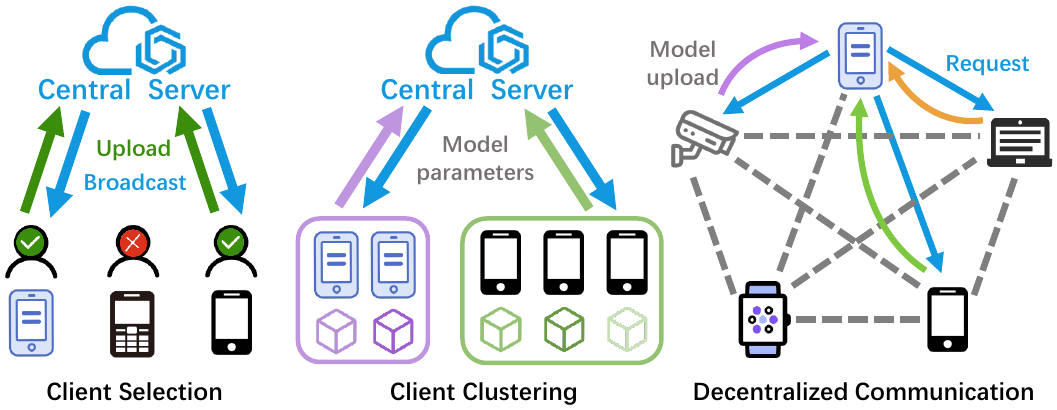}
   \caption{\small{Illustration of the server-level methods in heterogeneous federated learning.}}
\label{fig:Serverlevel}
\end{figure}

\subsection{Server-Level Methods}
\textcolor{black}{In this subsection, we categorize server-level methods and discuss the advantages and disadvantages of existing methods in Tab.~\ref{tab:servermethod}. Server-level methods refer to methods that rely on server-side operations, including client selection, client clustering and decentralized communication. Client selection aims to select the appropriate client to participate in the federated learning process for each iteration, which can solve various heterogeneous challenges. Client clustering improves the federated learning efficiency by aggregating similar clients, thereby alleviating communication and device heterogeneity. Decentralized communication enables peer-to-peer collaboration between devices without relying on a central server.}

\begin{table}[t]\small
\centering
 \caption{\label{tab:servermethod} \textcolor{black}{Server-level methods.}}
\begin{tabular}{|m{2.2cm}|m{2cm}|m{0.6cm}|m{4.8cm}|m{3.6cm}|} 
  \cline{1-5}
  \makecell[c]{Methods} & \makecell[c]{Advantages} & \makecell[c]{Ref.} & \makecell[c]{Key Contributions} & \makecell[c]{Limitations} \\
  \cline{1-5}
  &  & \makecell*[c]{\cite{ieee2020favor}} & Favor is an experience-driven control framework that actively selects the best subset of clients to participate in federated learning iterations. & Training reinforcement learning models may be data-hungry. \\
  \cline{3-5}
  &  & \makecell*[c]{\cite{corr2020cucb}} & CUCB is a client selection algorithm that minimizes the class imbalance and facilitates the global model convergence. & Revealing the class distribution based on updated gradients may be vulnerable to inference attacks. \\
  \cline{3-5}
  &  & \makecell*[c]{\cite{ijcnn2021fedsae}} & FedSAE estimates the reliability of each device and performs client selection based on training losses. & Adjusting workloads based on the training history of clients may be delayed. \\
  \cline{3-5}
  \multirow{-11}{0.15\textwidth}{\makecell*[c]{Client Selection \\ \ref{method:clientselct}}} & \multirow{-11.5}{0.15\textwidth}{These methods accelerate convergence by formulating client selection strategies.} & \cite{icc2019fedcs} & FedCS performs client selection operations based on data resources, computing capabilities, and wireless channel conditions. & Estimating training time for complex models may be difficult. \\
  \cline{1-5}
  &  & \makecell*[c]{\cite{ijcnn2020flhc}} & FL+HC introduces a hierarchical clustering step to separate client clusters based on the similarity of client updates to the global joint model. & The effect of communication heterogeneity and device heterogeneity may be ignored. \\
  \cline{3-5}
  &  & \makecell*[c]{\cite{xie2020fesem}} & FeSEM employs SEM optimization to calculate the distance between local models and cluster centers. & Compared with single-center clustering, multi-center clustering may require higher storage capability. \\
  \cline{3-5}
  &  & \makecell*[c]{\cite{ieee2020cfl}} & CFL clusters similar clients by the cosine similarity between their gradient updates. & It is vulnerable to backdoor attacks in federated learning. \\
  \cline{3-5}
  \multirow{-11}{0.15\textwidth}{\makecell*[c]{Client Clustering \\ \ref{method:clientcluster}}} & \multirow{-11.6}{0.15\textwidth}{These methods enhance HFL efficiency by personalized clustering clients.} & \cite{usenix2022flame} & FLAME detects adversarial model updates through a clustering strategy that limits the noise scale of backdoor noise removal. & Building trusted server in practical settings may be challenging. \\
  \cline{1-5}
  &  & \makecell*[c]{\cite{roy2019braintorrent}} & BrainTorrent randomly selects a client as a temporary server in each round, and then coordinates updates with other clients. & It requires high computational and storage resources for temporary servers.\\
  \cline{3-5}
  &  & \makecell*[c]{\cite{fml2019gossipdfl}} & Combo divides the local model into model segments, and then randomly selects some clients to transfer the model segments. & Transferring model segments alleviate communication delays, but do not reduce overall communication overhead.\\
  \cline{3-5}
  &  & \makecell*[c]{\cite{nature2023decentralflproxy}} & ProxyFL makes each client maintain two models, a private model and a publicly shared proxy model for exchanges. & Proxy models may not capture all the information or complexity of private models.\\
  \cline{3-5}
  \multirow{-10}{0.15\textwidth}{\makecell*[c]{\textcolor{black}{Decentralized}\\\textcolor{black}{Communication} \\ \ref{method:decentralcom}}} & \multirow{-10}{0.14\textwidth}{These methods can effectively reduce the reliance on the secure central server and alleviate the communication bottleneck.} & \cite{network2020blockchaindfl} & BFLC utilizes the blockchain for global model storage and local model update exchange to enhance the security of federated learning. & Maintaining and validating blockchain ledgers can incur high computational and storage costs. \\
  \cline{1-5}
\end{tabular}
\end{table}

\subsubsection{ Client Selection} \label{method:clientselct}
\

\
Client selection is typically performed by the server, so that data can be sampled from clients with uniform data distributions (Fig.~\ref{fig:Serverlevel}). Moreover, constraints such as the network bandwidth, computation capability, and local resources of different clients are considered when formulating selection strategies. Consequently, client selection can accelerate convergence and substantially improve the model accuracy.

The performance of traditional federated learning on Non-IID datasets is inferior to its performance on IID datasets, and the convergence speed on Non-IID datasets is also much slower than that on IID datasets. Several  methods~\cite{access2021csfedavg,iot2022e3cs,camad2020reputationaware,iclr2022divfl} are devised to alleviate the bias introduced by Non-IID data. Favor~\cite{ieee2020favor} is an experience-driven control framework that actively selects the best subset of clients to participate in federated learning iterations. Innovatively, they define device selection for federated learning as a deep reinforcement learning problem that aims to train an agent to learn an appropriate selection policy. In addition, class imbalance may occur when the client data distribution is inconsistent. To address this problem, Yang \textit{et al.}~\cite{corr2020cucb} devise an estimation scheme that clarifies the client class distribution based on the gradient of client service updates without considering the original data. Moreover, they design a client selection algorithm for the minimal class imbalance to improve the global model convergence. \textcolor{black}{Tang \textit{et al.}~\cite{cvpr2022fedcor} propose a correlation-based client selection strategy, using a Gaussian process to model loss changes of clients, and then selecting one client in each iteration to reduce the overall loss expectation. Qin \textit{et al.}~\cite{cvpr2023ripfl} believe that when selecting clients for collaborative training, in addition to considering the independent characteristics of clients, it is necessary to pay attention to the synergy between clients.}

The differences in the hardware and the network connectivity capabilities across clients may lead to high communication costs, low training efficiency, and wasted computing resources, etc. Thus, a large number of client selection strategies~\cite{cho2020powerofchoice,tpds2020cmab,sysml2019sysdesign,iot2020fedmccs,gc2020mab} aim to address these issues. In FedSAE~\cite{ijcnn2021fedsae}, the server estimates the reliability of each device based on the complete information of the client training history tasks, thereby adjusting the training load of each client. Furthermore, the server selects the clients with higher values based on the training losses of the clients to improve communication efficiency. FedCS~\cite{icc2019fedcs} performs the client selection operation based on the differences in data resources, computing capabilities and wireless channel conditions across client models and uses the selected clients $\mathbb{C}$ for aggregation. The core idea of client selection is to solve the following maximization problem:
\begin{equation}\label{eq:fedcs}
\begin{aligned}
 \max_{\mathbb{C}} |\mathbb{C}| s.t.\quad T_{epoch}\ge T_{cs}+T_{\mathbb{C}}^{d}+T_{\mathbb{C}}^{u}+T_{agg},
\end{aligned}
\end{equation}
where $T_{round}$ is the deadline for each round. In addition, $T_{cs}$, $T_{\mathbb{C}}^{d}$, $T_{\mathbb{C}}^{u}$, and $T_{agg}$ represent the time of client selection, distribution, scheduled update and upload, and aggregation, respectively. In this manner, FedCS effectively alleviates the effects of communication heterogeneity and device heterogeneity, while maximizing the number of participants in a round and improving the training efficiency. To better deal with the deviation problem caused by statistical, communication and device heterogeneity, TiFL~\cite{hpdc2020tifl} adopts an adaptive layer selection method that divides clients into different layers according to the training time and then selects clients from the same layer in each training round. \textcolor{black}{Li \textit{et al.}~\cite{iwqos2020mcfl} propose a multi-layer online coordination framework for high-performance energy-efficient federated learning, MCFL, which selects suitable devices by simultaneously considering the training data volume, computing power, and runtime training behavior of each device. Wu \textit{et al.}~\cite{tpds2020hybridfl} propose a multi-layer federated learning protocol, HybridFL, where each edge node randomly selects a subset of clients according to a specific probability distribution depending on the region slack factor. Regulating the proportion of selected clients mitigates the straggler and dropout problems caused by communication and device heterogeneity.}

\subsubsection{ Client Clustering} \label{method:clientcluster}
\

\
The model performance for all clients may not be satisfactory if the entire federated system shares one global model. Therefore, many existing approaches~\cite{ieee2020cfl,aaai2021fedsoft} perform the personalized clustering of all clients by considering the similarities of the data distributions, the local models and the parameter updates of different clients (Fig.~\ref{fig:Serverlevel}).

Briggs \textit{et al.}~\cite{ijcnn2020flhc} aim to decrease the number of communication rounds required to reach convergence while improving the test accuracy. To this end, a hierarchical clustering step is introduced to separate client clusters by the similarity of client updates to the global joint model. In addition, Xie \textit{et al.}~\cite{xie2020fesem} leverage a novel multi-center aggregation mechanism to address the problem of statistical heterogeneity. It utilizes a distance-based multi-center loss function to minimize the distance between a local model and its nearest global model, while constraining the variations in the local model updates. The local models $\{\theta_m\}_{m=1}^{M}$ are divided into $K$ clusters, {\it i.e.} cluster $1$,..., cluster $k$. The cluster $k$ has an aggregated model $\tilde{\theta}^k$ of multiple local models, and the multi-center loss function can be formulated as:
\begin{equation}\label{eq:fesem}
\begin{aligned}
& \min_{\tilde{\theta^k}} \frac{1}{M}\sum_{k=1}^{K}\sum_{m=1}^{M} r_m^k Dist(\theta_m,\tilde{\theta}^k),
\end{aligned}
\end{equation}
where $r_m^k$ represents the client $m$ is assigned to the cluster $k$. Then they propose FeSEM, which employs Stochastic Expectation Maximization (SEM)~\cite{jrstatsocb2009sem} optimization to calculate the distance between the local models and the cluster centers to ensure that an optimal match can be derived. To alleviate the high communication cost incurred by clustering, FedFMC~\cite{kopparapu2020fedfmc} dynamically groups devices of similar prototypes in certain epochs and then merges them into a single model. To improve the clustering efficiency, FedGroup~\cite{ispa2021fedgroup} designs a Euclidean distance of Decomposed Cosine similarity (EDC) to perform clustering based on the similarities between the optimization directions of the clients. A novel method called Clustered Federated Learning (CFL) \cite{ieee2020cfl} measures the similarity between the data distributions of different clients in terms of the cosine similarity between their gradient updates. The above operations can effectively identify the cluster structure, and thus clients with similar data can benefit from one another, while weakening harmful interference between clients with different data. Other methods measure the similarity of the models by comparing their loss values. For example, Iterative Federated Clustering Algorithm (IFCA)~\cite{ghosh2020ifca} minimizes the loss functions by alternately optimizing the cluster model parameters through gradient descent while estimating the client's cluster identity. \textcolor{black}{Lim \textit{et al.}~\cite{tpds2021clusterselect,jsac2021dynamichfl} propose a hierarchical federated learning framework to solve a two-level resource allocation and incentive mechanism design problem. At the edge layer, edge devices choose any cluster to join, and intermediate nodes provide rewards for the participation of edge devices. In the cloud layer, each intermediate node chooses any server, and the servers have to compete with each other for the service of the intermediate node. Feraudo \textit{et al.}~\cite{edgesys2020colearn} propose CoLearn, which leverages manufacturer usage description profiles of IoT devices to cluster devices with similar learning tasks and network requirements. CoLearn also uses a publish/subscribe messaging system to coordinate the learning process between edge devices and cloud servers.}

Several recent methods aim to improve the attack robustness of federated learning systems through client clustering. Sattler \textit{et al.}~\cite{icassp2020robustcfl} apply Clustered Federated Learning (CFL) to the Byzantine scenario, in which a subset of clients interferes with federated training in a detrimental way. A large number of clients are declared benign, and other clients are declared adversarial and excluded from training. This method decreases the computation cost while enhancing the robustness and flexibility of the federated framework. However, it is vulnerable to backdoor attacks (including data poisoning and model poisoning) in federated learning scenarios. To address this problem, Nguyen \textit{et al.}~\cite{usenix2022flame} propose FLAME, which does not rely on the underlying data distributions for benign and adversarial datasets. FLAME detects adversarial model updates through a clustering strategy that limits the noise scale of backdoor noise removal. To improve the performance of the aggregated model, FLAME implements a weighting method to limit the adversary models.

\subsubsection{ \textcolor{black}{Decentralized Communication}} \label{method:decentralcom}
\

\
\textcolor{black}{The general federated learning algorithm relies on a central server, which requires all clients to trust a central institution, and the failure of this institution will destroy the entire federated learning process. Therefore, some algorithms adopt decentralized communication~\cite{icml2023dfedsam}, which conducts peer-to-peer communication between various devices without relying on a central server (Fig.~\ref{fig:Serverlevel}).}

\textcolor{black}{Roy \textit{et al.}~\cite{roy2019braintorrent} propose a peer-to-peer federated learning framework without a central server, BrainTorrent, that is, direct communication between clients. Specifically, in each round, a client is randomly selected as a temporary server, and then cooperates with other clients that have completed the model update for collaborative update. In this way, any client can dynamically start an update process at any time. Lalitha \textit{et al.}~\cite{nips2018fullydecentralfl} propose a fully distributed federated learning algorithm, where clients can only communicate with their one-hop neighbors without relying on a centralized server. Clients update their local models by aggregating information obtained from one-hop neighbors to obtain an optimal resulting model. To sufficiently utilize the bandwidth capacity between clients while maintaining convergence performance, some decentralized federated training algorithms~\cite{fml2019gossipdfl,tpds2022gossipfl} are designed based on the Gossip protocol. For example, in GossipFL~\cite{tpds2022gossipfl}, each client dynamically selects a peer client based on its network bandwidth to fully utilize the global bandwidth resource. Then, the client exchanges a highly compressed model with its peer client, reducing communication traffic. Hu \textit{et al.}~\cite{fml2019gossipdfl} propose a segmented gossip approach, Combo, where the client $k$ divides the local model $W_K$ into $S$ segments, and then randomly selects some clients to transfer the model segments. $K_s$ represents the set of clients providing segment $s$ with $s\in S$ and $D_k$ represents the private dataset of client $k$. The segment $s$ can be aggregated with the private dataset size as the weights to get $\tilde{W}[s]$. Then, merge all aggregated segments to get the complete aggregated model $W$ as follows:}
\begin{equation}\label{eq:seggossip}
\begin{aligned}
\textcolor{black}{W=(\tilde{W}[1], \tilde{W}[2], ... \tilde{W}[s], ... \tilde{W}[S]), \tilde{W}[s]=\frac{\sum_{k\in{K_s}}W_k[s]\cdot|D_k|}{\sum_{k\in{K_s}}|D_k|}}.
\end{aligned}
\end{equation}
\textcolor{black}{Similar to the idea of this model segmentation, Pappas \textit{et al.}~\cite{ifip2021ipls} introduce a fully decentralized federated learning framework, IPLS. Each client retains some model segments. Before model training, it obtains model segments that are not available locally from other clients and combines them into a complete model. Then, the client uses the full model to update locally and then shares the update gradient with other clients. Kalra \textit{et al.}~\cite{nature2023decentralflproxy} design a proxy-based decentralized federated learning scheme, ProxyFL, in which each client maintains two models, a private model and a publicly shared proxy model. During collaborative learning, clients communicate with others by exchanging their proxy models. The way of sharing the proxy model can effectively deal with the challenge of model heterogeneity. To improve the security of decentralized federated learning, Li \textit{et al.}~\cite{network2020blockchaindfl} propose a blockchain-based decentralized federated learning framework, BFLC. The framework leverages blockchain for global model storage and local model update exchange. A committee consisting of honest clients verifies other clients so that the most reliable clients learn from each other, and a small number of malicious client updates are ignored.}

\section{Future Directions}\label{sec:future}
\subsection{Improving Communication Efficiency}
In federated learning, the heterogeneous environment can decrease the training efficiency to some extent. Therefore, improving communication efficiency and effectiveness is the focus of many existing approaches~\cite{icml2020fedboost,aaai2021localsgd,aaai2022feddst,icml2022progfed}. For example, Kone{\v{c}}n{\`y} \textit{et al.}~\cite{konevcny2016comefficiency} study two methods, one to learn an update from a finite space, and the other to update the model and send the compressed model to the server. Communication-Mitigated Federated Learning (CMFL)~\cite{icdcs2019cmfl} avoids the transmission of irrelevant updates to the server by measuring whether local updates are consistent with global updates, which effectively decreases the overhead of communication transmission. Federated Maximum and Mean Discrepancy (FedMMD)~\cite{yao2019fedmmd} decreases the number of communication rounds by introducing the Maximum Mean Discrepancy (MMD) constraint into the loss function. In addition, Caldas \textit{et al.}~\cite{caldas2018feddropout} develop a Fed-Dropout scheme by extending the concept presented in \cite{jmlr2014dropout}. Fed-Dropout derives a small sub-model of the global model for local updates and exchanges the sub-models between the server and the clients to decrease the communication cost. Xiong \textit{et al.}~\cite{cvpr2023feddm} propose FedDM to construct some synthetic data locally on the client so that it has a similar distribution to the original data on the loss function, and then send these synthetic data to the server for global model update. Compared with transmitting model parameters, transmitting a small amount of synthetic data can effectively reduce communication overhead and increase the amount of information. \textcolor{black}{Dai \textit{et al.}~\cite{icml2022dispfl} adopt a decentralized sparse training technique, so that each local model uses a personalized sparse mask to select its own active parameters, and maintains a fixed number of active parameters during local training and peer-to-peer communication. This way, each local model only needs to transmit the index of its active parameters once. In the subsequent communication process, only the values of these active parameters need to be transmitted instead of the entire model, thus greatly reducing the communication overhead. Besides, the multi-layer decentralized federated learning framework can alleviate the communication load and overhead on the server and client to some extent. However, it still faces challenges such as multi-hop communication delay, unbalanced communication load, and asynchronous communication.}

Nevertheless, heterogeneous federated learning encounters the following challenges to communication efficiency~\cite{shahid2021efficiencychallenge,cscwd2021evaluatefficiency}. The existence of a large number of edge nodes can increase computing costs and the required computing power and storage capacity. The differences in the network bandwidth can lead to delays or even losses in sending the local models from clients to the server~\cite{icml2022qsfl,icml2022eden}. In addition, the differences in the sizes of private datasets can also lead to delays in model updates. Thus, in practical application scenarios, a good trade-off between communication efficiency and model accuracy needs to be guaranteed.

\subsection{Federated Fairness}
In real-world scenarios, heterogeneous federated learning encounters the security issues associated with model fairness~\cite{sdm2021agnosticfair,bigdata2020fairfl}. The contribution of participating clients to collaborative learning varies and can be exacerbated by heterogeneity. However, the differences in the contribution of the participating clients in the collaborative process are ignored in most of the existing federated learning frameworks. In a fair collaborative federated system, clients with more contributions should be able to learn more from other clients and obtain superior models through collaborative training. In the case of general machine learning, the training data and the training patterns of models may be biased~\cite{nips2016equalitysl}, and some data sample groups may be discriminated against. In the federated learning scenario, there may exist several free-riding participants \cite{pmlr2021freerider} in the system who want to learn from others in federated communication without providing useful information. Additionally, a global model learned under the constraints common to all clients may be biased towards clients with larger amounts of data or frequent occurrences, and the overall loss function may implicitly advantage or disadvantage certain clients \cite{nips2020backdoor}. Therefore, the emphasis on fairness is expected to continue growing with extended practical deployments of federated learning to more users and enterprises. \textcolor{black}{Furthermore, multi-layer decentralized federated learning scenarios will face more difficult fairness problems, for example, the trade-off between fairness and efficiency in multi-layer communication protocols, and the inconsistency of fairness criteria between different layers.}

Several recently developed federated learning paradigms~\cite{nips2021fjord,icml2019agnosticfl,huang2020fedfa} aim to ensure fairness while maintaining high accuracies.
FPFL~\cite{nipsworkshop2021fpfl} regards the fairness problem in federated learning as an optimization problem subject to the constraints of fairness, and enhances the system fairness by improving the differential multiplier MMDM. q-FedAvg~\cite{iclr2020qfedavg} improves the fairness by narrowing the differences between the accuracies of client models. Besides, CFFL~\cite{flijcai2020fairnessfl} achieves collaborative fairness by assigning models with different performances based on the contributions of each client in federated learning. However, these methods cannot be well applied to heterogeneous federated learning scenarios. Besides, the potential relationship between collaborative fairness and privacy protection~\cite{icml2019privatefair,tpds2020fppdl} needs to be further investigated.

\subsection{Privacy Protection}
Privacy protection is the first and foremost principle of federated learning, and the clients typically safeguard basic privacy constraints by never sharing local data~\cite{icml2022dmesecagg,imwut2020pmf}.  
However, the clients may leak private information to the server, for instance, to infer sensitive private data samples as the client has memorized the previous model and gradient updates \cite{mai2018reconstruction} or information feedback \cite{cvpr2022fccl}. In addition, many methods, such as sharing few samples during data augmentation, and sharing local data distributions during knowledge transfer, inevitably result in privacy leakage to some extent~\cite{icml2022pbm,icml2022fednew}. 

Currently, many researches~\cite{lyu2020privacyrobust,icml2022ppsgd,icml2022cefedavg} dedicate to addressing privacy issues in federated learning. \textcolor{black}{In federated learning, one of the most popular privacy-preserving techniques is DP, which adds noise to local updates and clips the norm of local updates to preserve the original private information.} In FedAvg, larger updates introduce more noise, and a smaller number of iterations decreases the privacy consumption. Therefore, DP-FedAvg~\cite{iclr2018dpfedavg} applies a Gaussian mechanism to add user-level privacy protection in FedAvg, making large updates by user-level data. To ensure user-level privacy protection without compromising model performance, Cheng \textit{et al.}~\cite{cvpr2022dpfedavgbase} improve DP-FedAvg by adding regularization and sparsification processes to the local updates. Similarly, FedMD-NFDP~\cite{ijcai2021fedmdnfdp} claims that random noise perturbations can alleviate privacy concerns, but this process may sacrifice the model performance. Therefore, FedMD-NFDP adds a noise-free DP mechanism to FedMD~\cite{nips2019fedmd}, which protects data privacy without introducing noise. Moreover, in practical heterogeneous scenarios, different clients or data samples may have varying privacy concerns. 
Consequently, it is critical to build stricter and more flexible privacy constraint policies, which can measure and set more fine-grained privacy constraints for each client and sample while providing sufficient privacy guarantees. \textcolor{black}{Compared with the general federated learning scenario, it is more urgent to solve the privacy protection problem in the multi-layer decentralized federated learning scenario. Since edge devices may communicate directly with their neighbors, this may expose their raw data or model updates to other devices.}

\textcolor{black}{Besides, additional privacy concerns regarding personally sensitive information need to be considered when processing and using biometric data in federated learning. The corresponding solutions include raw data anonymization and feature template protection. Raw data anonymization means that when raw biometric data is preprocessed, a method is adopted so that personally identifiable information or other sensitive information such as gender, age, and health status cannot be extracted from the data\cite{zhang2022learnable}. This information may be used by malicious attackers, or correlated with information from other sources, thereby revealing personal privacy. Although existing federated learning frameworks adopt a data-stay-local policy, this cannot completely prevent local clients or other third parties from accessing and analyzing raw data. Therefore, several methods~\cite{csur2021privacyflsurvey,grama2020robustagganony} remove identifiable information through anonymization techniques, thereby protecting local sensitive information from being leaked while maintaining the practicability of published data. However, novel raw data anonymization schemes need to be designed, which can effectively protect the privacy of raw data and retain enough personally identifiable information for model training in federated learning. Meanwhile, the trade-off between privacy and practicality of data anonymization will also be a major challenge. Feature template protection means that after feature templates are extracted from the original biometric data, a method is adopted so that the original biometric data or other sensitive information cannot be reconstructed from these feature templates. A feature template is a representation that encodes and compresses raw biometric data. However, recent studies have indicated that raw biometric data can be reconstructed from feature templates \cite{tifs2020secureface}. Therefore, an irreversible, updateable, and verifiable feature template protection scheme is needed, while maintaining the distinguishability of the protected feature templates under the federated learning framework. In summary, in order to improve the security and efficiency of processing and using biometric data in federated learning systems, the raw data anonymization and the feature template protection in federated learning need further investigation.}



\subsection{Attack Robustness}
Federated systems may be vulnerable to two major types of attacks~\cite{lyu2020attacksurvey}: poisoning attacks and inference attacks. 1) Poisoning attacks attempt to prevent the models from being learned and make the learning directions of the models deviate from the original goal. Such attacks involve data poisoning and model poisoning. Data poisoning~\cite{icml2012poisonattacksvm} means that the adversaries compromise the integrity of the training data through methods such as label flipping~\cite{fung2018sybilspoison} and backdoor insertion~\cite{gu2017badnets,nips2019canyoubackdoor}, thereby deteriorating the model performance. In model poisoning~\cite{aistats2020backdoor}, adversaries cannot directly operate on private data, but they can change the learning direction of the model by destroying client updates. 2) Inference attacks infer information about private user data, thereby compromising user privacy. For example, in the process of parameter transmission, the malicious clients can infer the sensitive data of other clients according to the differences of gradient parameters in each round~\cite{tifs2017privacyinfer}. \textcolor{black}{Moreover, compared with the server-client federated learning scenario, there are more intermediate nodes that can be attacked in the multi-layer decentralized federated learning scenario, so it may face more serious malicious attacks and require stricter defense mechanisms.}

Attack methods~\cite{iot2020poisongan,esorics2020datapoisonattack,icml2020backdoorattckincl,network2022coordinatedbackdoor} in federated learning settings should be studied to improve the attack robustness of federated systems. Xie \textit{et al.}~\cite{iclr2019dba} propose the Distributed Backdoor Attack (DBA) strategy, in which a global trigger is decomposed into local triggers, and they are injected into multiple malicious clients. Such distributed backdoor attacks are more stealthy and effective than centralized backdoor attacks. Moreover, Nguyen \textit{et al.}~\cite{nips2020backdoor} propose edge-case backdoors, which consider poisoning edge-case samples. Edge-case samples typically represent the tail data of the data distributions and are unlikely to be used as training or test data. Fowl \textit{et al.}~\cite{iclr2022robfed} propose an attack method based on the imprint module, so that the server directly obtains a copy of the original data from the gradient uploaded by the clients. The imprint module is a special convolutional layer that can directly copy the input feature map to the output feature map, so that the original input information is contained in the gradient.

\textcolor{black}{These attacks pose a significant threat to federated learning, and thus several defense strategies~\cite{nips2021flwbc,esorics2021romoa,aaai2021defendbackdoor,icml2022neurotoxin} have been developed to enhance system robustness. Xie \textit{et al.}~\cite{icml2021crfl} improve the robustness against backdoor attacks by clipping the model and adding smooth noise. To counteract model poisoning attacks, Li \textit{et al.}~\cite{li2020detectmalicious} learn a detection model on the server side to identify and remove malicious model updates, thereby removing irrelevant features while retaining valid basic features. Wu \textit{et al.}~\cite{ieee2022blockchainmultifl} propose a robust blockchain multi-layer decentralized federated learning framework, RBML-DFL, which can prevent central server failures or malfunctions through blockchain encrypted transactions. To deal with inference attacks, ResSFL~\cite{cvpr2022ressfl} is trained by experts through attacker perception to obtain a resistant feature extractor that can initialize the client models. Besides, Soteria~\cite{cvpr2021soteria} performs attack defense by generating perturbed data representations, thereby decreasing the quality of reconstructed data. Several approaches~\cite{iot2021detedctsecurityattack,ihmmsec2021federatedreverse} implement attack detection to proactively identify the malicious intrusions in federated systems. In BaFFle~\cite{icdcs2021baffle}, the server trains backdoor filters and sends them randomly to clients to detect backdoor instances. The client then removes the backdoor instances from the training data, thereby training a backdoor-free local model. Different from other methods that are studied in the federated setting, Abeshu \textit{et al.}~\cite{ieee2018fogattack} apply federated learning to network attack detection, and propose a network attack detection model based on federated learning-authorized edge network. This enables edge devices to learn from each other without sharing data to improve the accuracy of network detection attacks. Recently, hardware-based Trusted Execution Environment (TEE) that allocates an isolated block of memory for private computing of sensitive data has attracted significant attention from the industry. Unlike general privacy-preserving technologies, TEE is committed to providing a secure platform for federated learning with low computational overhead and high computational efficiency, and protecting models from inference attacks. Therefore, it is suitable for federated learning scenarios with limited computing resources~\cite{grama2020robustagganony}. Mo \textit{et al.}~\cite{mobisys2021ppfl} propose a privacy-preserving federated learning framework for mobile systems, PPFL, which uses TEE for secure training of local models and secure aggregation on the server, thereby hiding gradient updates and preventing adversary attacks. Zhang \textit{et al.}~\cite{cf2021shufflefl} utilize TEE to protect the integrity and privacy of gradients and prevent adversary models from inferring or modifying gradients through side-channel attacks. Specifically, they use TEE for local training, randomly group gradients into gradient segments for encryption, and then send them to the server. The server decrypts and securely aggregates gradient segments from the same group with TEE.}

The inference attacks aim to obtain the information of the clients, and thus inevitably threaten user privacy and security. Therefore, the attack defense strategies can decrease the risk of privacy leakage. Several privacy-preserving mechanisms can not only reduce inference attacks, but also improve robustness against poisoning attacks. Future work should be aimed at exploring the close relationships between attack robustness and privacy protection to enable a better trade-off between these two aspects.

\subsection{Uniform Benchmarks}
The federated learning concept was first proposed by McMahan \textit{et al.}~\cite{mcmahan2017fedavg} in 2016. As this field has been developed for a relatively short period of time, there is a lack of widely recognized benchmark datasets and benchmark testing frameworks for heterogeneous scenarios. The development of unified benchmark datasets and benchmark testing frameworks can facilitate the reproduction of experimental results and widespread application of novel algorithms. Heterogeneous benchmark frameworks provide various possibilities for client-side data distributions and model structures. Moreover, the statistical and model discrepancies of different clients can validate the generalization ability of heterogeneous federated learning algorithms to some extent. Systematic evaluation indicators should be developed to promote the research and development of heterogeneous federated learning. The indicators can fairly and comprehensively evaluate the security, convergence, accuracy, and generalization ability of different algorithms. \textcolor{black}{Therefore, benchmark is very important to promote the development of heterogeneous federated learning field. Several recent works have explored benchmark frameworks and datasets, which we group into the following three categories:}

\textcolor{black}{
\textbf{General Federated Learning Systems.}
FedML~\cite{he2020fedml} is a research library that supports distributed training, mobile on-device training, and stand-alone simulation training. It provides standardized implementations of many existing federated learning algorithms, and provides standardized benchmark settings for a variety of datasets, including Non-IID partition methods, number of devices and baseline models.
FedScale~\cite{icml2022fedscale} is a federated learning benchmark suite that provides real-world datasets covering a wide range of federated learning tasks, including image classification, object detection, language modeling, and speech recognition. Additionally, FedScale includes a scalable and extensible FedScale Runtime to enable and standardize real-world end-point deployments of federated learning.
OARF~\cite{hu2020oarf} leverages public datasets collected from different sources to simulate real-world data distributions. In addition, OARF quantitatively studies the preliminary relationship among various design metrics such as data partitioning and privacy mechanisms in federated learning systems.
FedEval~\cite{chai2020fedeval} is a federated learning evaluation model with five metrics including accuracy, communication, time consumption, privacy and robustness. FedEval is implemented and evaluated on two of the most widely used algorithms, FedSGD and FedAvg.
}

\textcolor{black}{
\textbf{Specific Federated Learning Systems.}
FedReIDBench~\cite{acmmm2020fedreid} is a new benchmark for implementing federated learning to person ReID, which includes nine different datasets and two federated scenarios. Specifically, the two federated scenarios are federated-by-camera scenario and federated-by-dataset scenario, which respectively represent the standard server-client architecture and client-edge-cloud architecture.
pFL-Bench~\cite{nips2022pflbench} is a benchmark for personalized federated learning, which covers twelve different dataset variants, including image, text, graph and recommendation data, with unified data partitioning and realistic heterogeneous settings. And pFL-Bench provides more than 20 competitive personalized federated learning baseline implementations to help them with standardized evaluation.
FedGraphNN~\cite{iclr2021fedgraphnn} is a benchmark system built on a unified formulation of graph federated learning, including extensive datasets from seven different fields, popular Graph Neural Network (GNN) models and federated learning algorithms.
}

\textcolor{black}{
\textbf{Datasets.}
LEAF~\cite{nips2019leaf} contains 6 types of federated datasets covering different fields, including image classification (FEMNIST, Synthetic Dataset), image recognition (Celeba), sentiment analysis (Sentiment140) and next character prediction (Shakespeare, Reddit). In addition, LEAF provides two sampling methods of 'IID' and 'Non-IID' to divide the dataset to different clients.
Luo~\textit{et al.}~\cite{realworldfldataset} introduce a federated dataset for object detection. The dataset contains over 900 images generated from 26 street cameras and 7 object categories annotated with detailed bounding boxes. Besides, the article provides the data division of 5 or 20 clients, in which their data distribution is Non-IID and unbalanced, reflecting the characteristics of real-world federated learning scenarios.
}

Although several federated learning benchmarks are devised, more realistic datasets containing extensive machine learning tasks should be established to facilitate the development of federated learning. In heterogeneous scenarios, the scales and the distributions of local data on different clients may differ significantly, and it is generally difficult to benchmark the Non-IID performance under different algorithms. The Non-IID setting requires a sufficient understanding of the statistical heterogeneity in real-world scenarios, including the four statistical heterogeneity cases we discussed in \textcolor{black}{Subsection} \ref{sec:prodata}. \textcolor{black}{Therefore, developing a realistic benchmark framework that incorporates the four heterogeneities is a challenging task.}

\section{Conclusion}
This survey aims to provide a comprehensive and systematic understanding of heterogeneous federated learning. First, a detailed overview of the research challenges in heterogeneous federated learning is demonstrated in \textcolor{black}{Section} \ref{sec:problems}. Different from existing surveys on federated learning, we focus on the heterogeneity problems in federated learning and categorize them into four categories: statistical heterogeneity, model heterogeneity, communication heterogeneity, and device heterogeneity. Subsequently, we survey the recently published and pre-printed papers on heterogeneous federated learning and provide a reasonable and comprehensive taxonomy of existing techniques in \textcolor{black}{Section} \ref{sec:methods}. This taxonomy divides the state-of-the-art methods at three different levels: data level, model level, and server level. Finally, we provide an outlook analysis on the directions worthy of further exploration and open problems for future development in heterogeneous federated learning in \textcolor{black}{Section} \ref{sec:future}. We believe that these valuable discussions can promote the high-quality development of the heterogeneous federated learning community.

\bibliographystyle{ACM-Reference-Format}
\bibliography{survey}


\begin{thebibliography}{256}


\ifx \showCODEN    \undefined \def \showCODEN     #1{\unskip}     \fi
\ifx \showDOI      \undefined \def \showDOI       #1{#1}\fi
\ifx \showISBNx    \undefined \def \showISBNx     #1{\unskip}     \fi
\ifx \showISBNxiii \undefined \def \showISBNxiii  #1{\unskip}     \fi
\ifx \showISSN     \undefined \def \showISSN      #1{\unskip}     \fi
\ifx \showLCCN     \undefined \def \showLCCN      #1{\unskip}     \fi
\ifx \shownote     \undefined \def \shownote      #1{#1}          \fi
\ifx \showarticletitle \undefined \def \showarticletitle #1{#1}   \fi
\ifx \showURL      \undefined \def \showURL       {\relax}        \fi
\providecommand\bibfield[2]{#2}
\providecommand\bibinfo[2]{#2}
\providecommand\natexlab[1]{#1}
\providecommand\showeprint[2][]{arXiv:#2}

\bibitem[AbdulRahman et~al\mbox{.}(2020)]%
        {iot2020fedmccs}
\bibfield{author}{\bibinfo{person}{Sawsan AbdulRahman}, \bibinfo{person}{Hanine
  Tout}, \bibinfo{person}{Azzam Mourad}, {and} \bibinfo{person}{Chamseddine
  Talhi}.} \bibinfo{year}{2020}\natexlab{}.
\newblock \showarticletitle{FedMCCS: Multicriteria client selection model for
  optimal IoT federated learning}.
\newblock \bibinfo{journal}{\emph{IEEE IoT}} (\bibinfo{year}{2020}).
\newblock


\bibitem[Abeshu and Chilamkurti(2018)]%
        {ieee2018fogattack}
\bibfield{author}{\bibinfo{person}{Abebe Abeshu} {and} \bibinfo{person}{Naveen
  Chilamkurti}.} \bibinfo{year}{2018}\natexlab{}.
\newblock \showarticletitle{Deep learning: The frontier for distributed attack
  detection in fog-to-things computing}.
\newblock \bibinfo{journal}{\emph{IEEE Communications Magazine}}
  (\bibinfo{year}{2018}).
\newblock


\bibitem[Alex et~al\mbox{.}(2018)]%
        {nichol2018reptile}
\bibfield{author}{\bibinfo{person}{Nichol Alex}, \bibinfo{person}{Achiam
  Joshua}, {and} \bibinfo{person}{Schulman John}.}
  \bibinfo{year}{2018}\natexlab{}.
\newblock \showarticletitle{On First-Order Meta-Learning Algorithms}.
\newblock \bibinfo{journal}{\emph{CoRR}} (\bibinfo{year}{2018}).
\newblock


\bibitem[Andreina et~al\mbox{.}(2021)]%
        {icdcs2021baffle}
\bibfield{author}{\bibinfo{person}{Sebastien Andreina},
  \bibinfo{person}{Giorgia~Azzurra Marson}, \bibinfo{person}{Helen
  M{\"o}llering}, {and} \bibinfo{person}{Ghassan Karame}.}
  \bibinfo{year}{2021}\natexlab{}.
\newblock \showarticletitle{Baffle: Backdoor detection via feedback-based
  federated learning}. In \bibinfo{booktitle}{\emph{ICDCS}}.
\newblock


\bibitem[Aono et~al\mbox{.}(2017)]%
        {tifs2017privacyinfer}
\bibfield{author}{\bibinfo{person}{Yoshinori Aono}, \bibinfo{person}{Takuya
  Hayashi}, \bibinfo{person}{Lihua Wang}, \bibinfo{person}{Shiho Moriai},
  {et~al\mbox{.}}} \bibinfo{year}{2017}\natexlab{}.
\newblock \showarticletitle{Privacy-preserving deep learning via additively
  homomorphic encryption}.
\newblock \bibinfo{journal}{\emph{IEEE TIFS}} (\bibinfo{year}{2017}).
\newblock


\bibitem[Arivazhagan et~al\mbox{.}(2019)]%
        {aiatits2020fedper}
\bibfield{author}{\bibinfo{person}{Manoj~Ghuhan Arivazhagan},
  \bibinfo{person}{Vinay Aggarwal}, \bibinfo{person}{Aaditya~Kumar Singh},
  {and} \bibinfo{person}{Sunav Choudhary}.} \bibinfo{year}{2019}\natexlab{}.
\newblock \showarticletitle{Federated learning with personalization layers}.
\newblock \bibinfo{journal}{\emph{arXiv preprint arXiv:1912.00818}}
  (\bibinfo{year}{2019}).
\newblock


\bibitem[Asad et~al\mbox{.}(2020)]%
        {appsci2020fedopt}
\bibfield{author}{\bibinfo{person}{Muhammad Asad}, \bibinfo{person}{Ahmed
  Moustafa}, {and} \bibinfo{person}{Takayuki Ito}.}
  \bibinfo{year}{2020}\natexlab{}.
\newblock \showarticletitle{FedOpt: Towards communication efficiency and
  privacy preservation in federated learning}.
\newblock \bibinfo{journal}{\emph{Applied Sciences}} (\bibinfo{year}{2020}).
\newblock


\bibitem[Asad et~al\mbox{.}(2021)]%
        {cscwd2021evaluatefficiency}
\bibfield{author}{\bibinfo{person}{Muhammad Asad}, \bibinfo{person}{Ahmed
  Moustafa}, \bibinfo{person}{Takayuki Ito}, {and} \bibinfo{person}{Muhammad
  Aslam}.} \bibinfo{year}{2021}\natexlab{}.
\newblock \showarticletitle{Evaluating the communication efficiency in
  federated learning algorithms}. In \bibinfo{booktitle}{\emph{IEEE CSCWD}}.
\newblock


\bibitem[Bagdasaryan et~al\mbox{.}(2020)]%
        {aistats2020backdoor}
\bibfield{author}{\bibinfo{person}{Eugene Bagdasaryan},
  \bibinfo{person}{Andreas Veit}, \bibinfo{person}{Yiqing Hua},
  \bibinfo{person}{Deborah Estrin}, {and} \bibinfo{person}{Vitaly Shmatikov}.}
  \bibinfo{year}{2020}\natexlab{}.
\newblock \showarticletitle{How to backdoor federated learning}. In
  \bibinfo{booktitle}{\emph{AISTATS}}.
\newblock


\bibitem[Balakrishnan et~al\mbox{.}(2022)]%
        {iclr2022divfl}
\bibfield{author}{\bibinfo{person}{Ravikumar Balakrishnan},
  \bibinfo{person}{Tian Li}, \bibinfo{person}{Tianyi Zhou},
  \bibinfo{person}{Nageen Himayat}, \bibinfo{person}{Virginia Smith}, {and}
  \bibinfo{person}{Jeff Bilmes}.} \bibinfo{year}{2022}\natexlab{}.
\newblock \showarticletitle{Diverse client selection for federated learning via
  submodular maximization}. In \bibinfo{booktitle}{\emph{ICLR}}.
\newblock


\bibitem[Bibikar et~al\mbox{.}(2022)]%
        {aaai2022feddst}
\bibfield{author}{\bibinfo{person}{Sameer Bibikar}, \bibinfo{person}{Haris
  Vikalo}, \bibinfo{person}{Zhangyang Wang}, {and} \bibinfo{person}{Xiaohan
  Chen}.} \bibinfo{year}{2022}\natexlab{}.
\newblock \showarticletitle{Federated dynamic sparse training: Computing less,
  communicating less, yet learning better}. In
  \bibinfo{booktitle}{\emph{AAAI}}.
\newblock


\bibitem[Bietti et~al\mbox{.}(2022)]%
        {icml2022ppsgd}
\bibfield{author}{\bibinfo{person}{Alberto Bietti}, \bibinfo{person}{Chen-Yu
  Wei}, \bibinfo{person}{Miroslav Dudik}, \bibinfo{person}{John Langford},
  {and} \bibinfo{person}{Steven Wu}.} \bibinfo{year}{2022}\natexlab{}.
\newblock \showarticletitle{Personalization Improves Privacy-Accuracy Tradeoffs
  in Federated Learning}. In \bibinfo{booktitle}{\emph{ICML}}.
\newblock


\bibitem[Biggio et~al\mbox{.}(2012)]%
        {icml2012poisonattacksvm}
\bibfield{author}{\bibinfo{person}{Battista Biggio}, \bibinfo{person}{Blaine
  Nelson}, {and} \bibinfo{person}{Pavel Laskov}.}
  \bibinfo{year}{2012}\natexlab{}.
\newblock \showarticletitle{Poisoning attacks against support vector machines}.
  In \bibinfo{booktitle}{\emph{ICML}}.
\newblock


\bibitem[Bonawitz et~al\mbox{.}(2019)]%
        {sysml2019sysdesign}
\bibfield{author}{\bibinfo{person}{Keith Bonawitz}, \bibinfo{person}{Hubert
  Eichner}, \bibinfo{person}{Wolfgang Grieskamp}, \bibinfo{person}{Dzmitry
  Huba}, \bibinfo{person}{Alex Ingerman}, \bibinfo{person}{Vladimir Ivanov},
  \bibinfo{person}{Chloe Kiddon}, \bibinfo{person}{Jakub Kone{\v{c}}n{\`y}},
  \bibinfo{person}{Stefano Mazzocchi}, \bibinfo{person}{Brendan McMahan},
  {et~al\mbox{.}}} \bibinfo{year}{2019}\natexlab{}.
\newblock \showarticletitle{Towards federated learning at scale: System
  design}.
\newblock \bibinfo{journal}{\emph{SysML}} (\bibinfo{year}{2019}).
\newblock


\bibitem[Briggs et~al\mbox{.}(2020)]%
        {ijcnn2020flhc}
\bibfield{author}{\bibinfo{person}{Christopher Briggs}, \bibinfo{person}{Zhong
  Fan}, {and} \bibinfo{person}{Peter Andras}.} \bibinfo{year}{2020}\natexlab{}.
\newblock \showarticletitle{Federated learning with hierarchical clustering of
  local updates to improve training on non-IID data}. In
  \bibinfo{booktitle}{\emph{IJCNN}}.
\newblock


\bibitem[Caldas et~al\mbox{.}(2019)]%
        {nips2019leaf}
\bibfield{author}{\bibinfo{person}{Sebastian Caldas}, \bibinfo{person}{Sai
  Meher~Karthik Duddu}, \bibinfo{person}{Peter Wu}, \bibinfo{person}{Tian Li},
  \bibinfo{person}{Jakub Kone{\v{c}}n{\`y}}, \bibinfo{person}{H~Brendan
  McMahan}, \bibinfo{person}{Virginia Smith}, {and} \bibinfo{person}{Ameet
  Talwalkar}.} \bibinfo{year}{2019}\natexlab{}.
\newblock \showarticletitle{Leaf: A benchmark for federated settings}. In
  \bibinfo{booktitle}{\emph{NeurIPS Workshop}}.
\newblock


\bibitem[Caldas et~al\mbox{.}(2018)]%
        {caldas2018feddropout}
\bibfield{author}{\bibinfo{person}{Sebastian Caldas}, \bibinfo{person}{Jakub
  Kone{\v{c}}ny}, \bibinfo{person}{H~Brendan McMahan}, {and}
  \bibinfo{person}{Ameet Talwalkar}.} \bibinfo{year}{2018}\natexlab{}.
\newblock \showarticletitle{Expanding the reach of federated learning by
  reducing client resource requirements}.
\newblock \bibinfo{journal}{\emph{arXiv preprint arXiv:1812.07210}}
  (\bibinfo{year}{2018}).
\newblock


\bibitem[Capp{\'e} and Moulines(2009)]%
        {jrstatsocb2009sem}
\bibfield{author}{\bibinfo{person}{Olivier Capp{\'e}} {and}
  \bibinfo{person}{Eric Moulines}.} \bibinfo{year}{2009}\natexlab{}.
\newblock \showarticletitle{On-line expectation--maximization algorithm for
  latent data models}.
\newblock \bibinfo{journal}{\emph{J R STAT SOC B}} (\bibinfo{year}{2009}).
\newblock


\bibitem[Chai et~al\mbox{.}(2020b)]%
        {chai2020fedeval}
\bibfield{author}{\bibinfo{person}{Di Chai}, \bibinfo{person}{Leye Wang},
  \bibinfo{person}{Kai Chen}, {and} \bibinfo{person}{Qiang Yang}.}
  \bibinfo{year}{2020}\natexlab{b}.
\newblock \showarticletitle{FedEval: A Holistic Evaluation Framework for
  Federated Learning}.
\newblock \bibinfo{journal}{\emph{arXiv preprint arXiv:2011.09655}}
  (\bibinfo{year}{2020}).
\newblock


\bibitem[Chai et~al\mbox{.}(2020a)]%
        {hpdc2020tifl}
\bibfield{author}{\bibinfo{person}{Zheng Chai}, \bibinfo{person}{Ahsan Ali},
  \bibinfo{person}{Syed Zawad}, \bibinfo{person}{Stacey Truex},
  \bibinfo{person}{Ali Anwar}, \bibinfo{person}{Nathalie Baracaldo},
  \bibinfo{person}{Yi Zhou}, \bibinfo{person}{Heiko Ludwig},
  \bibinfo{person}{Feng Yan}, {and} \bibinfo{person}{Yue Cheng}.}
  \bibinfo{year}{2020}\natexlab{a}.
\newblock \showarticletitle{Tifl: A tier-based federated learning system}. In
  \bibinfo{booktitle}{\emph{ACM HPDC}}.
\newblock


\bibitem[Chang et~al\mbox{.}(2019)]%
        {chang2019cronus}
\bibfield{author}{\bibinfo{person}{Hongyan Chang}, \bibinfo{person}{Virat
  Shejwalkar}, \bibinfo{person}{Reza Shokri}, {and} \bibinfo{person}{Amir
  Houmansadr}.} \bibinfo{year}{2019}\natexlab{}.
\newblock \showarticletitle{Cronus: Robust and heterogeneous collaborative
  learning with black-box knowledge transfer}.
\newblock \bibinfo{journal}{\emph{arXiv preprint arXiv:1912.11279}}
  (\bibinfo{year}{2019}).
\newblock


\bibitem[Chen et~al\mbox{.}(2022b)]%
        {nips2022pflbench}
\bibfield{author}{\bibinfo{person}{Daoyuan Chen}, \bibinfo{person}{Dawei Gao},
  \bibinfo{person}{Weirui Kuang}, \bibinfo{person}{Yaliang Li}, {and}
  \bibinfo{person}{Bolin Ding}.} \bibinfo{year}{2022}\natexlab{b}.
\newblock \showarticletitle{pFL-Bench: A Comprehensive Benchmark for
  Personalized Federated Learning}. In \bibinfo{booktitle}{\emph{NeurIPS Track
  on Datasets and Benchmarks}}.
\newblock


\bibitem[Chen et~al\mbox{.}(2023a)]%
        {cvpr2023elasticagg}
\bibfield{author}{\bibinfo{person}{Dengsheng Chen}, \bibinfo{person}{Jie Hu},
  \bibinfo{person}{Vince~Junkai Tan}, \bibinfo{person}{Xiaoming Wei}, {and}
  \bibinfo{person}{Enhua Wu}.} \bibinfo{year}{2023}\natexlab{a}.
\newblock \showarticletitle{Elastic Aggregation for Federated Optimization}. In
  \bibinfo{booktitle}{\emph{CVPR}}.
\newblock


\bibitem[Chen et~al\mbox{.}(2023b)]%
        {icml2023pfedgate}
\bibfield{author}{\bibinfo{person}{Daoyuan Chen}, \bibinfo{person}{Liuyi Yao},
  \bibinfo{person}{Dawei Gao}, \bibinfo{person}{Bolin Ding}, {and}
  \bibinfo{person}{Yaliang Li}.} \bibinfo{year}{2023}\natexlab{b}.
\newblock \showarticletitle{Efficient Personalized Federated Learning via
  Sparse Model-Adaptation}.
\newblock


\bibitem[Chen et~al\mbox{.}(2018)]%
        {chen2018fedmeta}
\bibfield{author}{\bibinfo{person}{Fei Chen}, \bibinfo{person}{Mi Luo},
  \bibinfo{person}{Zhenhua Dong}, \bibinfo{person}{Zhenguo Li}, {and}
  \bibinfo{person}{Xiuqiang He}.} \bibinfo{year}{2018}\natexlab{}.
\newblock \showarticletitle{Federated meta-learning with fast convergence and
  efficient communication}.
\newblock \bibinfo{journal}{\emph{arXiv preprint arXiv:1802.07876}}
  (\bibinfo{year}{2018}).
\newblock


\bibitem[Chen et~al\mbox{.}(2021)]%
        {pnas2021comeff}
\bibfield{author}{\bibinfo{person}{Mingzhe Chen}, \bibinfo{person}{Nir
  Shlezinger}, \bibinfo{person}{H~Vincent Poor}, \bibinfo{person}{Yonina~C
  Eldar}, {and} \bibinfo{person}{Shuguang Cui}.}
  \bibinfo{year}{2021}\natexlab{}.
\newblock \showarticletitle{Communication-efficient federated learning}.
\newblock \bibinfo{journal}{\emph{PNAS}} (\bibinfo{year}{2021}).
\newblock


\bibitem[Chen et~al\mbox{.}(2020a)]%
        {chen2020vafl}
\bibfield{author}{\bibinfo{person}{Tianyi Chen}, \bibinfo{person}{Xiao Jin},
  \bibinfo{person}{Yuejiao Sun}, {and} \bibinfo{person}{Wotao Yin}.}
  \bibinfo{year}{2020}\natexlab{a}.
\newblock \showarticletitle{Vafl: a method of vertical asynchronous federated
  learning}. In \bibinfo{booktitle}{\emph{ICML Workshop}}.
\newblock


\bibitem[Chen et~al\mbox{.}(2022a)]%
        {icml2022dmesecagg}
\bibfield{author}{\bibinfo{person}{Wei-Ning Chen}, \bibinfo{person}{Christopher
  A~Choquette Choo}, \bibinfo{person}{Peter Kairouz}, {and}
  \bibinfo{person}{Ananda~Theertha Suresh}.} \bibinfo{year}{2022}\natexlab{a}.
\newblock \showarticletitle{The fundamental price of secure aggregation in
  differentially private federated learning}. In
  \bibinfo{booktitle}{\emph{ICML}}.
\newblock


\bibitem[Chen et~al\mbox{.}(2022c)]%
        {icml2022pbm}
\bibfield{author}{\bibinfo{person}{Wei-Ning Chen}, \bibinfo{person}{Ayfer
  Ozgur}, {and} \bibinfo{person}{Peter Kairouz}.}
  \bibinfo{year}{2022}\natexlab{c}.
\newblock \showarticletitle{The Poisson Binomial Mechanism for Unbiased
  Federated Learning with Secure Aggregation}. In
  \bibinfo{booktitle}{\emph{ICML}}.
\newblock


\bibitem[Chen et~al\mbox{.}(2020b)]%
        {ieee2020fedhealth}
\bibfield{author}{\bibinfo{person}{Yiqiang Chen}, \bibinfo{person}{Xin Qin},
  \bibinfo{person}{Jindong Wang}, \bibinfo{person}{Chaohui Yu}, {and}
  \bibinfo{person}{Wen Gao}.} \bibinfo{year}{2020}\natexlab{b}.
\newblock \showarticletitle{Fedhealth: A federated transfer learning framework
  for wearable healthcare}.
\newblock \bibinfo{journal}{\emph{IEEE Intelligent Systems}}
  (\bibinfo{year}{2020}).
\newblock


\bibitem[Cheng et~al\mbox{.}(2022)]%
        {cvpr2022dpfedavgbase}
\bibfield{author}{\bibinfo{person}{Anda Cheng}, \bibinfo{person}{Peisong Wang},
  \bibinfo{person}{Xi~Sheryl Zhang}, {and} \bibinfo{person}{Jian Cheng}.}
  \bibinfo{year}{2022}\natexlab{}.
\newblock \showarticletitle{Differentially Private Federated Learning with
  Local Regularization and Sparsification}. In
  \bibinfo{booktitle}{\emph{CVPR}}.
\newblock


\bibitem[Cho et~al\mbox{.}(2020)]%
        {cho2020powerofchoice}
\bibfield{author}{\bibinfo{person}{Yae~Jee Cho}, \bibinfo{person}{Jianyu Wang},
  {and} \bibinfo{person}{Gauri Joshi}.} \bibinfo{year}{2020}\natexlab{}.
\newblock \showarticletitle{Client selection in federated learning: Convergence
  analysis and power-of-choice selection strategies}.
\newblock \bibinfo{journal}{\emph{arXiv preprint arXiv:2010.01243}}
  (\bibinfo{year}{2020}).
\newblock


\bibitem[Choudhury et~al\mbox{.}(2020)]%
        {choudhury2020anonymizing}
\bibfield{author}{\bibinfo{person}{Olivia Choudhury}, \bibinfo{person}{Aris
  Gkoulalas-Divanis}, \bibinfo{person}{Theodoros Salonidis},
  \bibinfo{person}{Issa Sylla}, \bibinfo{person}{Yoonyoung Park},
  \bibinfo{person}{Grace Hsu}, {and} \bibinfo{person}{Amar Das}.}
  \bibinfo{year}{2020}\natexlab{}.
\newblock \showarticletitle{Anonymizing data for privacy-preserving federated
  learning}.
\newblock \bibinfo{journal}{\emph{arXiv preprint arXiv:2002.09096}}
  (\bibinfo{year}{2020}).
\newblock


\bibitem[Chu et~al\mbox{.}(2022)]%
        {chu2022multilayerperfl}
\bibfield{author}{\bibinfo{person}{Yun-Wei Chu}, \bibinfo{person}{Seyyedali
  Hosseinalipour}, \bibinfo{person}{Elizabeth Tenorio}, \bibinfo{person}{Laura
  Cruz}, \bibinfo{person}{Kerrie Douglas}, \bibinfo{person}{Andrew Lan}, {and}
  \bibinfo{person}{Christopher Brinton}.} \bibinfo{year}{2022}\natexlab{}.
\newblock \showarticletitle{Multi-Layer Personalized Federated Learning for
  Mitigating Biases in Student Predictive Analytics}.
\newblock \bibinfo{journal}{\emph{arXiv preprint arXiv:2212.02985}}
  (\bibinfo{year}{2022}).
\newblock


\bibitem[Collins et~al\mbox{.}(2021)]%
        {icml2021fedrep}
\bibfield{author}{\bibinfo{person}{Liam Collins}, \bibinfo{person}{Hamed
  Hassani}, \bibinfo{person}{Aryan Mokhtari}, {and} \bibinfo{person}{Sanjay
  Shakkottai}.} \bibinfo{year}{2021}\natexlab{}.
\newblock \showarticletitle{Exploiting shared representations for personalized
  federated learning}. In \bibinfo{booktitle}{\emph{ICML}}.
\newblock


\bibitem[Corinzia et~al\mbox{.}(2019)]%
        {corinzia2019virtual}
\bibfield{author}{\bibinfo{person}{Luca Corinzia}, \bibinfo{person}{Ami
  Beuret}, {and} \bibinfo{person}{Joachim~M Buhmann}.}
  \bibinfo{year}{2019}\natexlab{}.
\newblock \showarticletitle{Variational federated multi-task learning}.
\newblock \bibinfo{journal}{\emph{arXiv preprint arXiv:1906.06268}}
  (\bibinfo{year}{2019}).
\newblock


\bibitem[Dai et~al\mbox{.}(2022)]%
        {icml2022dispfl}
\bibfield{author}{\bibinfo{person}{Rong Dai}, \bibinfo{person}{Li Shen},
  \bibinfo{person}{Fengxiang He}, \bibinfo{person}{Xinmei Tian}, {and}
  \bibinfo{person}{Dacheng Tao}.} \bibinfo{year}{2022}\natexlab{}.
\newblock \showarticletitle{DisPFL: Towards Communication-Efficient
  Personalized Federated Learning via Decentralized Sparse Training}. In
  \bibinfo{booktitle}{\emph{ICML}}.
\newblock


\bibitem[Dai et~al\mbox{.}(2015)]%
        {aaai2015hpml}
\bibfield{author}{\bibinfo{person}{Wei Dai}, \bibinfo{person}{Abhimanu Kumar},
  \bibinfo{person}{Jinliang Wei}, \bibinfo{person}{Qirong Ho},
  \bibinfo{person}{Garth Gibson}, {and} \bibinfo{person}{Eric Xing}.}
  \bibinfo{year}{2015}\natexlab{}.
\newblock \showarticletitle{High-performance distributed ML at scale through
  parameter server consistency models}. In \bibinfo{booktitle}{\emph{AAAI}}.
\newblock


\bibitem[Dayan et~al\mbox{.}(2021)]%
        {nature2021exam}
\bibfield{author}{\bibinfo{person}{Ittai Dayan}, \bibinfo{person}{Holger~R
  Roth}, \bibinfo{person}{Aoxiao Zhong}, \bibinfo{person}{Ahmed Harouni},
  \bibinfo{person}{Amilcare Gentili}, \bibinfo{person}{Anas~Z Abidin},
  \bibinfo{person}{Andrew Liu}, \bibinfo{person}{Anthony~Beardsworth Costa},
  \bibinfo{person}{Bradford~J Wood}, \bibinfo{person}{Chien-Sung Tsai},
  {et~al\mbox{.}}} \bibinfo{year}{2021}\natexlab{}.
\newblock \showarticletitle{Federated learning for predicting clinical outcomes
  in patients with COVID-19}.
\newblock \bibinfo{journal}{\emph{Nature medicine}} (\bibinfo{year}{2021}).
\newblock


\bibitem[Diao et~al\mbox{.}(2021)]%
        {iclr2021heterofl}
\bibfield{author}{\bibinfo{person}{Enmao Diao}, \bibinfo{person}{Jie Ding},
  {and} \bibinfo{person}{Vahid Tarokh}.} \bibinfo{year}{2021}\natexlab{}.
\newblock \showarticletitle{Heterofl: Computation and communication efficient
  federated learning for heterogeneous clients}. In
  \bibinfo{booktitle}{\emph{ICLR}}.
\newblock


\bibitem[Dinh et~al\mbox{.}(2021)]%
        {dinh2021fedu}
\bibfield{author}{\bibinfo{person}{Canh~T Dinh}, \bibinfo{person}{Tung~T Vu},
  \bibinfo{person}{Nguyen~H Tran}, \bibinfo{person}{Minh~N Dao}, {and}
  \bibinfo{person}{Hongyu Zhang}.} \bibinfo{year}{2021}\natexlab{}.
\newblock \showarticletitle{FedU: A Unified Framework for Federated Multi-Task
  Learning with Laplacian Regularization}.
\newblock \bibinfo{journal}{\emph{arXiv preprint arXiv:2102.07148}}
  (\bibinfo{year}{2021}).
\newblock


\bibitem[Du et~al\mbox{.}(2021)]%
        {sdm2021agnosticfair}
\bibfield{author}{\bibinfo{person}{Wei Du}, \bibinfo{person}{Depeng Xu},
  \bibinfo{person}{Xintao Wu}, {and} \bibinfo{person}{Hanghang Tong}.}
  \bibinfo{year}{2021}\natexlab{}.
\newblock \showarticletitle{Fairness-aware agnostic federated learning}. In
  \bibinfo{booktitle}{\emph{SDM}}.
\newblock


\bibitem[Duan et~al\mbox{.}(2019)]%
        {iccd2019astraea}
\bibfield{author}{\bibinfo{person}{Moming Duan}, \bibinfo{person}{Duo Liu},
  \bibinfo{person}{Xianzhang Chen}, \bibinfo{person}{Yujuan Tan},
  \bibinfo{person}{Jinting Ren}, \bibinfo{person}{Lei Qiao}, {and}
  \bibinfo{person}{Liang Liang}.} \bibinfo{year}{2019}\natexlab{}.
\newblock \showarticletitle{Astraea: Self-balancing federated learning for
  improving classification accuracy of mobile deep learning applications}. In
  \bibinfo{booktitle}{\emph{ICCD}}.
\newblock


\bibitem[Duan et~al\mbox{.}(2021)]%
        {ispa2021fedgroup}
\bibfield{author}{\bibinfo{person}{Moming Duan}, \bibinfo{person}{Duo Liu},
  \bibinfo{person}{Xinyuan Ji}, \bibinfo{person}{Renping Liu},
  \bibinfo{person}{Liang Liang}, \bibinfo{person}{Xianzhang Chen}, {and}
  \bibinfo{person}{Yujuan Tan}.} \bibinfo{year}{2021}\natexlab{}.
\newblock \showarticletitle{FedGroup: Efficient Federated Learning via
  Decomposed Similarity-Based Clustering}. In \bibinfo{booktitle}{\emph{IEEE
  ISPA}}.
\newblock


\bibitem[Elgabli et~al\mbox{.}(2022)]%
        {icml2022fednew}
\bibfield{author}{\bibinfo{person}{Anis Elgabli}, \bibinfo{person}{Chaouki~Ben
  Issaid}, \bibinfo{person}{Amrit~Singh Bedi}, \bibinfo{person}{Ketan Rajawat},
  \bibinfo{person}{Mehdi Bennis}, {and} \bibinfo{person}{Vaneet Aggarwal}.}
  \bibinfo{year}{2022}\natexlab{}.
\newblock \showarticletitle{FedNew: A Communication-Efficient and
  Privacy-Preserving Newton-Type Method for Federated Learning}. In
  \bibinfo{booktitle}{\emph{ICML}}.
\newblock


\bibitem[Fallah et~al\mbox{.}(2020)]%
        {nips2020perfedavg}
\bibfield{author}{\bibinfo{person}{Alireza Fallah}, \bibinfo{person}{Aryan
  Mokhtari}, {and} \bibinfo{person}{Asuman Ozdaglar}.}
  \bibinfo{year}{2020}\natexlab{}.
\newblock \showarticletitle{Personalized federated learning with theoretical
  guarantees: A model-agnostic meta-learning approach}. In
  \bibinfo{booktitle}{\emph{NeurIPS}}.
\newblock


\bibitem[Fang et~al\mbox{.}(2020)]%
        {comsec2020homoencryfl}
\bibfield{author}{\bibinfo{person}{Chen Fang}, \bibinfo{person}{Yuanbo Guo},
  \bibinfo{person}{Na Wang}, {and} \bibinfo{person}{Ankang Ju}.}
  \bibinfo{year}{2020}\natexlab{}.
\newblock \showarticletitle{Highly efficient federated learning with strong
  privacy preservation in cloud computing}.
\newblock \bibinfo{journal}{\emph{Computers \& Security}}
  (\bibinfo{year}{2020}).
\newblock


\bibitem[Fang and Ye(2022)]%
        {cvpr2022rhfl}
\bibfield{author}{\bibinfo{person}{Xiuwen Fang} {and} \bibinfo{person}{Mang
  Ye}.} \bibinfo{year}{2022}\natexlab{}.
\newblock \showarticletitle{Robust Federated Learning with Noisy and
  Heterogeneous Clients}. In \bibinfo{booktitle}{\emph{CVPR}}.
\newblock


\bibitem[Feng et~al\mbox{.}(2020)]%
        {imwut2020pmf}
\bibfield{author}{\bibinfo{person}{Jie Feng}, \bibinfo{person}{Can Rong},
  \bibinfo{person}{Funing Sun}, \bibinfo{person}{Diansheng Guo}, {and}
  \bibinfo{person}{Yong Li}.} \bibinfo{year}{2020}\natexlab{}.
\newblock \showarticletitle{PMF: A privacy-preserving human mobility prediction
  framework via federated learning}.
\newblock \bibinfo{journal}{\emph{ACM IMWUT}} (\bibinfo{year}{2020}).
\newblock


\bibitem[Feraudo et~al\mbox{.}(2020)]%
        {edgesys2020colearn}
\bibfield{author}{\bibinfo{person}{Angelo Feraudo}, \bibinfo{person}{Poonam
  Yadav}, \bibinfo{person}{Vadim Safronov}, \bibinfo{person}{Diana~Andreea
  Popescu}, \bibinfo{person}{Richard Mortier}, \bibinfo{person}{Shiqiang Wang},
  \bibinfo{person}{Paolo Bellavista}, {and} \bibinfo{person}{Jon Crowcroft}.}
  \bibinfo{year}{2020}\natexlab{}.
\newblock \showarticletitle{CoLearn: Enabling federated learning in
  MUD-compliant IoT edge networks}. In \bibinfo{booktitle}{\emph{ACM EdgeSys}}.
\newblock


\bibitem[Finn et~al\mbox{.}(2017)]%
        {icml2017maml}
\bibfield{author}{\bibinfo{person}{Chelsea Finn}, \bibinfo{person}{Pieter
  Abbeel}, {and} \bibinfo{person}{Sergey Levine}.}
  \bibinfo{year}{2017}\natexlab{}.
\newblock \showarticletitle{Model-agnostic meta-learning for fast adaptation of
  deep networks}. In \bibinfo{booktitle}{\emph{ICML}}.
\newblock


\bibitem[Fowl et~al\mbox{.}(2022)]%
        {iclr2022robfed}
\bibfield{author}{\bibinfo{person}{Liam Fowl}, \bibinfo{person}{Jonas Geiping},
  \bibinfo{person}{Wojtek Czaja}, \bibinfo{person}{Micah Goldblum}, {and}
  \bibinfo{person}{Tom Goldstein}.} \bibinfo{year}{2022}\natexlab{}.
\newblock \showarticletitle{Robbing the fed: Directly obtaining private data in
  federated learning with modified models}. In
  \bibinfo{booktitle}{\emph{ICLR}}.
\newblock


\bibitem[Fraboni et~al\mbox{.}(2021)]%
        {pmlr2021freerider}
\bibfield{author}{\bibinfo{person}{Yann Fraboni}, \bibinfo{person}{Richard
  Vidal}, {and} \bibinfo{person}{Marco Lorenzi}.}
  \bibinfo{year}{2021}\natexlab{}.
\newblock \showarticletitle{Free-rider attacks on model aggregation in
  federated learning}. In \bibinfo{booktitle}{\emph{AISTATS}}.
\newblock


\bibitem[Froelicher et~al\mbox{.}(2021)]%
        {nature2021famhe}
\bibfield{author}{\bibinfo{person}{David Froelicher}, \bibinfo{person}{Juan~R
  Troncoso-Pastoriza}, \bibinfo{person}{Jean~Louis Raisaro},
  \bibinfo{person}{Michel~A Cuendet}, \bibinfo{person}{Joao~Sa Sousa},
  \bibinfo{person}{Hyunghoon Cho}, \bibinfo{person}{Bonnie Berger},
  \bibinfo{person}{Jacques Fellay}, {and} \bibinfo{person}{Jean-Pierre
  Hubaux}.} \bibinfo{year}{2021}\natexlab{}.
\newblock \showarticletitle{Truly privacy-preserving federated analytics for
  precision medicine with multiparty homomorphic encryption}.
\newblock \bibinfo{journal}{\emph{Nature communications}}
  (\bibinfo{year}{2021}).
\newblock


\bibitem[Fung et~al\mbox{.}(2018)]%
        {fung2018sybilspoison}
\bibfield{author}{\bibinfo{person}{Clement Fung}, \bibinfo{person}{Chris~JM
  Yoon}, {and} \bibinfo{person}{Ivan Beschastnikh}.}
  \bibinfo{year}{2018}\natexlab{}.
\newblock \showarticletitle{Mitigating sybils in federated learning poisoning}.
\newblock \bibinfo{journal}{\emph{arXiv preprint arXiv:1808.04866}}
  (\bibinfo{year}{2018}).
\newblock


\bibitem[G{\'a}lvez et~al\mbox{.}(2021)]%
        {nipsworkshop2021fpfl}
\bibfield{author}{\bibinfo{person}{Borja~Rodr{\'\i}guez G{\'a}lvez},
  \bibinfo{person}{Filip Granqvist}, \bibinfo{person}{Rogier van Dalen}, {and}
  \bibinfo{person}{Matt Seigel}.} \bibinfo{year}{2021}\natexlab{}.
\newblock \showarticletitle{Enforcing fairness in private federated learning
  via the modified method of differential multipliers}. In
  \bibinfo{booktitle}{\emph{NeurIPS Workshop}}.
\newblock


\bibitem[Gao et~al\mbox{.}(2019)]%
        {bigdata2019hftl}
\bibfield{author}{\bibinfo{person}{Dashan Gao}, \bibinfo{person}{Yang Liu},
  \bibinfo{person}{Anbu Huang}, \bibinfo{person}{Ce Ju}, \bibinfo{person}{Han
  Yu}, {and} \bibinfo{person}{Qiang Yang}.} \bibinfo{year}{2019}\natexlab{}.
\newblock \showarticletitle{Privacy-preserving heterogeneous federated transfer
  learning}. In \bibinfo{booktitle}{\emph{IEEE Big Data}}.
\newblock


\bibitem[Gao et~al\mbox{.}(2022)]%
        {gao2022hflsurvey}
\bibfield{author}{\bibinfo{person}{Dashan Gao}, \bibinfo{person}{Xin Yao},
  {and} \bibinfo{person}{Qiang Yang}.} \bibinfo{year}{2022}\natexlab{}.
\newblock \showarticletitle{A survey on heterogeneous federated learning}.
\newblock \bibinfo{journal}{\emph{arXiv preprint arXiv:2210.04505}}
  (\bibinfo{year}{2022}).
\newblock


\bibitem[Gao et~al\mbox{.}(2021)]%
        {aaai2021localsgd}
\bibfield{author}{\bibinfo{person}{Hongchang Gao}, \bibinfo{person}{An Xu},
  {and} \bibinfo{person}{Heng Huang}.} \bibinfo{year}{2021}\natexlab{}.
\newblock \showarticletitle{On the convergence of communication-efficient local
  SGD for federated learning}. In \bibinfo{booktitle}{\emph{AAAI}}.
\newblock


\bibitem[Ghosh et~al\mbox{.}(2020)]%
        {ghosh2020ifca}
\bibfield{author}{\bibinfo{person}{Avishek Ghosh}, \bibinfo{person}{Jichan
  Chung}, \bibinfo{person}{Dong Yin}, {and} \bibinfo{person}{Kannan
  Ramchandran}.} \bibinfo{year}{2020}\natexlab{}.
\newblock \showarticletitle{An efficient framework for clustered federated
  learning}. In \bibinfo{booktitle}{\emph{NeurIPS}}.
\newblock


\bibitem[Girgis et~al\mbox{.}(2021)]%
        {aistats2021cldpsgd}
\bibfield{author}{\bibinfo{person}{Antonious Girgis}, \bibinfo{person}{Deepesh
  Data}, \bibinfo{person}{Suhas Diggavi}, \bibinfo{person}{Peter Kairouz},
  {and} \bibinfo{person}{Ananda~Theertha Suresh}.}
  \bibinfo{year}{2021}\natexlab{}.
\newblock \showarticletitle{Shuffled model of differential privacy in federated
  learning}. In \bibinfo{booktitle}{\emph{AISTATS}}.
\newblock


\bibitem[Gong et~al\mbox{.}(2022)]%
        {network2022coordinatedbackdoor}
\bibfield{author}{\bibinfo{person}{Xueluan Gong}, \bibinfo{person}{Yanjiao
  Chen}, \bibinfo{person}{Huayang Huang}, \bibinfo{person}{Yuqing Liao},
  \bibinfo{person}{Shuai Wang}, {and} \bibinfo{person}{Qian Wang}.}
  \bibinfo{year}{2022}\natexlab{}.
\newblock \showarticletitle{Coordinated Backdoor Attacks against Federated
  Learning with Model-Dependent Triggers}.
\newblock \bibinfo{journal}{\emph{IEEE Network}} (\bibinfo{year}{2022}).
\newblock


\bibitem[Goodfellow et~al\mbox{.}(2014)]%
        {nips2014gan}
\bibfield{author}{\bibinfo{person}{Ian Goodfellow}, \bibinfo{person}{Jean
  Pouget-Abadie}, \bibinfo{person}{Mehdi Mirza}, \bibinfo{person}{Bing Xu},
  \bibinfo{person}{David Warde-Farley}, \bibinfo{person}{Sherjil Ozair},
  \bibinfo{person}{Aaron Courville}, {and} \bibinfo{person}{Yoshua Bengio}.}
  \bibinfo{year}{2014}\natexlab{}.
\newblock \showarticletitle{Generative adversarial nets}. In
  \bibinfo{booktitle}{\emph{NeurIPS}}.
\newblock


\bibitem[Grama et~al\mbox{.}(2020)]%
        {grama2020robustagganony}
\bibfield{author}{\bibinfo{person}{Matei Grama}, \bibinfo{person}{Maria Musat},
  \bibinfo{person}{Luis Mu{\~n}oz-Gonz{\'a}lez}, \bibinfo{person}{Jonathan
  Passerat-Palmbach}, \bibinfo{person}{Daniel Rueckert}, {and}
  \bibinfo{person}{Amir Alansary}.} \bibinfo{year}{2020}\natexlab{}.
\newblock \showarticletitle{Robust aggregation for adaptive privacy preserving
  federated learning in healthcare}.
\newblock \bibinfo{journal}{\emph{arXiv preprint arXiv:2009.08294}}
  (\bibinfo{year}{2020}).
\newblock


\bibitem[Gu et~al\mbox{.}(2017)]%
        {gu2017badnets}
\bibfield{author}{\bibinfo{person}{Tianyu Gu}, \bibinfo{person}{Brendan
  Dolan-Gavitt}, {and} \bibinfo{person}{Siddharth Garg}.}
  \bibinfo{year}{2017}\natexlab{}.
\newblock \showarticletitle{Badnets: Identifying vulnerabilities in the machine
  learning model supply chain}.
\newblock \bibinfo{journal}{\emph{arXiv preprint arXiv:1708.06733}}
  (\bibinfo{year}{2017}).
\newblock


\bibitem[Guo et~al\mbox{.}(2022)]%
        {tist2022fedhumor}
\bibfield{author}{\bibinfo{person}{Xu Guo}, \bibinfo{person}{Han Yu},
  \bibinfo{person}{Boyang Li}, \bibinfo{person}{Hao Wang},
  \bibinfo{person}{Pengwei Xing}, \bibinfo{person}{Siwei Feng},
  \bibinfo{person}{Zaiqing Nie}, {and} \bibinfo{person}{Chunyan Miao}.}
  \bibinfo{year}{2022}\natexlab{}.
\newblock \showarticletitle{Federated Learning for Personalized Humor
  Recognition}. In \bibinfo{booktitle}{\emph{ACM TIST}}.
\newblock


\bibitem[G{\"u}rsoy et~al\mbox{.}(2022)]%
        {nature2022genomics}
\bibfield{author}{\bibinfo{person}{Gamze G{\"u}rsoy}, \bibinfo{person}{Tianxiao
  Li}, \bibinfo{person}{Susanna Liu}, \bibinfo{person}{Eric Ni},
  \bibinfo{person}{Charlotte~M Brannon}, {and} \bibinfo{person}{Mark~B
  Gerstein}.} \bibinfo{year}{2022}\natexlab{}.
\newblock \showarticletitle{Functional genomics data: privacy risk assessment
  and technological mitigation}.
\newblock \bibinfo{journal}{\emph{Nature Reviews Genetics}}
  (\bibinfo{year}{2022}).
\newblock


\bibitem[Hamer et~al\mbox{.}(2020)]%
        {icml2020fedboost}
\bibfield{author}{\bibinfo{person}{Jenny Hamer}, \bibinfo{person}{Mehryar
  Mohri}, {and} \bibinfo{person}{Ananda~Theertha Suresh}.}
  \bibinfo{year}{2020}\natexlab{}.
\newblock \showarticletitle{Fedboost: A communication-efficient algorithm for
  federated learning}. In \bibinfo{booktitle}{\emph{ICML}}.
\newblock


\bibitem[Hanzely and Richt{\'a}rik(2020)]%
        {hanzely2020l2gd}
\bibfield{author}{\bibinfo{person}{Filip Hanzely} {and} \bibinfo{person}{Peter
  Richt{\'a}rik}.} \bibinfo{year}{2020}\natexlab{}.
\newblock \showarticletitle{Federated learning of a mixture of global and local
  models}.
\newblock \bibinfo{journal}{\emph{arXiv preprint arXiv:2002.05516}}
  (\bibinfo{year}{2020}).
\newblock


\bibitem[Hardt et~al\mbox{.}(2016)]%
        {nips2016equalitysl}
\bibfield{author}{\bibinfo{person}{Moritz Hardt}, \bibinfo{person}{Eric Price},
  {and} \bibinfo{person}{Nati Srebro}.} \bibinfo{year}{2016}\natexlab{}.
\newblock \showarticletitle{Equality of opportunity in supervised learning}. In
  \bibinfo{booktitle}{\emph{NeurIPS}}.
\newblock


\bibitem[He et~al\mbox{.}(2020a)]%
        {fedgkt2020nips}
\bibfield{author}{\bibinfo{person}{Chaoyang He}, \bibinfo{person}{Murali
  Annavaram}, {and} \bibinfo{person}{Salman Avestimehr}.}
  \bibinfo{year}{2020}\natexlab{a}.
\newblock \showarticletitle{Group knowledge transfer: Federated learning of
  large cnns at the edge}. In \bibinfo{booktitle}{\emph{NeurIPS}}.
\newblock


\bibitem[He et~al\mbox{.}(2021)]%
        {iclr2021fedgraphnn}
\bibfield{author}{\bibinfo{person}{Chaoyang He}, \bibinfo{person}{Keshav
  Balasubramanian}, \bibinfo{person}{Emir Ceyani}, \bibinfo{person}{Carl Yang},
  \bibinfo{person}{Han Xie}, \bibinfo{person}{Lichao Sun},
  \bibinfo{person}{Lifang He}, \bibinfo{person}{Liangwei Yang},
  \bibinfo{person}{Philip~S Yu}, \bibinfo{person}{Yu Rong}, {et~al\mbox{.}}}
  \bibinfo{year}{2021}\natexlab{}.
\newblock \showarticletitle{Fedgraphnn: A federated learning system and
  benchmark for graph neural networks}. In \bibinfo{booktitle}{\emph{ICLR
  Workshop}}.
\newblock


\bibitem[He et~al\mbox{.}(2020b)]%
        {he2020fedml}
\bibfield{author}{\bibinfo{person}{Chaoyang He}, \bibinfo{person}{Songze Li},
  \bibinfo{person}{Jinhyun So}, \bibinfo{person}{Xiao Zeng},
  \bibinfo{person}{Mi Zhang}, \bibinfo{person}{Hongyi Wang},
  \bibinfo{person}{Xiaoyang Wang}, \bibinfo{person}{Praneeth Vepakomma},
  \bibinfo{person}{Abhishek Singh}, \bibinfo{person}{Hang Qiu},
  {et~al\mbox{.}}} \bibinfo{year}{2020}\natexlab{b}.
\newblock \showarticletitle{Fedml: A research library and benchmark for
  federated machine learning}.
\newblock \bibinfo{journal}{\emph{arXiv preprint arXiv:2007.13518}}
  (\bibinfo{year}{2020}).
\newblock


\bibitem[Hong et~al\mbox{.}(2022)]%
        {iclr2022splitmix}
\bibfield{author}{\bibinfo{person}{Junyuan Hong}, \bibinfo{person}{Haotao
  Wang}, \bibinfo{person}{Zhangyang Wang}, {and} \bibinfo{person}{Jiayu Zhou}.}
  \bibinfo{year}{2022}\natexlab{}.
\newblock \showarticletitle{Efficient Split-Mix Federated Learning for
  On-Demand and In-Situ Customization}. In \bibinfo{booktitle}{\emph{ICLR}}.
\newblock


\bibitem[H{\"o}nig et~al\mbox{.}(2022)]%
        {icml2022dadaquant}
\bibfield{author}{\bibinfo{person}{Robert H{\"o}nig}, \bibinfo{person}{Yiren
  Zhao}, {and} \bibinfo{person}{Robert Mullins}.}
  \bibinfo{year}{2022}\natexlab{}.
\newblock \showarticletitle{DAdaQuant: Doubly-adaptive quantization for
  communication-efficient Federated Learning}. In
  \bibinfo{booktitle}{\emph{ICML}}.
\newblock


\bibitem[Horvath et~al\mbox{.}(2021)]%
        {nips2021fjord}
\bibfield{author}{\bibinfo{person}{Samuel Horvath}, \bibinfo{person}{Stefanos
  Laskaridis}, \bibinfo{person}{Mario Almeida}, \bibinfo{person}{Ilias
  Leontiadis}, \bibinfo{person}{Stylianos Venieris}, {and}
  \bibinfo{person}{Nicholas Lane}.} \bibinfo{year}{2021}\natexlab{}.
\newblock \showarticletitle{Fjord: Fair and accurate federated learning under
  heterogeneous targets with ordered dropout}. In
  \bibinfo{booktitle}{\emph{NeurIPS}}.
\newblock


\bibitem[Hosseinalipour et~al\mbox{.}(2020)]%
        {ieee2020fogfl}
\bibfield{author}{\bibinfo{person}{Seyyedali Hosseinalipour},
  \bibinfo{person}{Christopher~G Brinton}, \bibinfo{person}{Vaneet Aggarwal},
  \bibinfo{person}{Huaiyu Dai}, {and} \bibinfo{person}{Mung Chiang}.}
  \bibinfo{year}{2020}\natexlab{}.
\newblock \showarticletitle{From federated to fog learning: Distributed machine
  learning over heterogeneous wireless networks}.
\newblock \bibinfo{journal}{\emph{IEEE Communications Magazine}}
  (\bibinfo{year}{2020}).
\newblock


\bibitem[Hou et~al\mbox{.}(2022)]%
        {iclr2022fedchain}
\bibfield{author}{\bibinfo{person}{Charlie Hou}, \bibinfo{person}{Kiran~Koshy
  Thekumparampil}, \bibinfo{person}{Giulia Fanti}, {and}
  \bibinfo{person}{Sewoong Oh}.} \bibinfo{year}{2022}\natexlab{}.
\newblock \showarticletitle{FedChain: Chained Algorithms for Near-optimal
  Communication Cost in Federated Learning}. In
  \bibinfo{booktitle}{\emph{ICLR}}.
\newblock


\bibitem[Hu et~al\mbox{.}(2019)]%
        {fml2019gossipdfl}
\bibfield{author}{\bibinfo{person}{Chenghao Hu}, \bibinfo{person}{Jingyan
  Jiang}, {and} \bibinfo{person}{Zhi Wang}.} \bibinfo{year}{2019}\natexlab{}.
\newblock \showarticletitle{Decentralized federated learning: A segmented
  gossip approach}. In \bibinfo{booktitle}{\emph{FML Workshop}}.
\newblock


\bibitem[Hu et~al\mbox{.}(2020a)]%
        {iot2020perfldprivacy}
\bibfield{author}{\bibinfo{person}{Rui Hu}, \bibinfo{person}{Yuanxiong Guo},
  \bibinfo{person}{Hongning Li}, \bibinfo{person}{Qingqi Pei}, {and}
  \bibinfo{person}{Yanmin Gong}.} \bibinfo{year}{2020}\natexlab{a}.
\newblock \showarticletitle{Personalized federated learning with differential
  privacy}.
\newblock \bibinfo{journal}{\emph{IEEE IoT}} (\bibinfo{year}{2020}).
\newblock


\bibitem[Hu et~al\mbox{.}(2020b)]%
        {hu2020oarf}
\bibfield{author}{\bibinfo{person}{Sixu Hu}, \bibinfo{person}{Yuan Li},
  \bibinfo{person}{Xu Liu}, \bibinfo{person}{Qinbin Li},
  \bibinfo{person}{Zhaomin Wu}, {and} \bibinfo{person}{Bingsheng He}.}
  \bibinfo{year}{2020}\natexlab{b}.
\newblock \showarticletitle{The oarf benchmark suite: Characterization and
  implications for federated learning systems}.
\newblock \bibinfo{journal}{\emph{Intelligent Systems and Technology}}
  (\bibinfo{year}{2020}).
\newblock


\bibitem[Huang et~al\mbox{.}(2022a)]%
        {iot2022e3cs}
\bibfield{author}{\bibinfo{person}{Tiansheng Huang}, \bibinfo{person}{Weiwei
  Lin}, \bibinfo{person}{Li Shen}, \bibinfo{person}{Keqin Li}, {and}
  \bibinfo{person}{Albert~Y Zomaya}.} \bibinfo{year}{2022}\natexlab{a}.
\newblock \showarticletitle{Stochastic client selection for federated learning
  with volatile clients}.
\newblock \bibinfo{journal}{\emph{IEEE IoT}} (\bibinfo{year}{2022}).
\newblock


\bibitem[Huang et~al\mbox{.}(2020b)]%
        {tpds2020cmab}
\bibfield{author}{\bibinfo{person}{Tiansheng Huang}, \bibinfo{person}{Weiwei
  Lin}, \bibinfo{person}{Wentai Wu}, \bibinfo{person}{Ligang He},
  \bibinfo{person}{Keqin Li}, {and} \bibinfo{person}{Albert~Y Zomaya}.}
  \bibinfo{year}{2020}\natexlab{b}.
\newblock \showarticletitle{An efficiency-boosting client selection scheme for
  federated learning with fairness guarantee}.
\newblock \bibinfo{journal}{\emph{IEEE TPDS}} (\bibinfo{year}{2020}).
\newblock


\bibitem[Huang et~al\mbox{.}(2020a)]%
        {huang2020fedfa}
\bibfield{author}{\bibinfo{person}{Wei Huang}, \bibinfo{person}{Tianrui Li},
  \bibinfo{person}{Dexian Wang}, \bibinfo{person}{Shengdong Du}, {and}
  \bibinfo{person}{Junbo Zhang}.} \bibinfo{year}{2020}\natexlab{a}.
\newblock \showarticletitle{Fairness and accuracy in federated learning}.
\newblock \bibinfo{journal}{\emph{arXiv preprint arXiv:2012.10069}}
  (\bibinfo{year}{2020}).
\newblock


\bibitem[Huang et~al\mbox{.}(2022b)]%
        {cvpr2022fccl}
\bibfield{author}{\bibinfo{person}{Wenke Huang}, \bibinfo{person}{Mang Ye},
  {and} \bibinfo{person}{Bo Du}.} \bibinfo{year}{2022}\natexlab{b}.
\newblock \showarticletitle{Learn from Others and Be Yourself in Heterogeneous
  Federated Learning}. In \bibinfo{booktitle}{\emph{CVPR}}.
\newblock


\bibitem[Huang et~al\mbox{.}(2022c)]%
        {acmmm2022fewshot}
\bibfield{author}{\bibinfo{person}{Wenke Huang}, \bibinfo{person}{Mang Ye},
  \bibinfo{person}{Xiang Gao}, {and} \bibinfo{person}{Bo Du}.}
  \bibinfo{year}{2022}\natexlab{c}.
\newblock \showarticletitle{Few-Shot Model Agnostic Federated Learning}. In
  \bibinfo{booktitle}{\emph{ACM Multimedia}}.
\newblock


\bibitem[Huang et~al\mbox{.}(2023)]%
        {cvpr2023pfl}
\bibfield{author}{\bibinfo{person}{Wenke Huang}, \bibinfo{person}{Mang Ye},
  \bibinfo{person}{Zekun Shi}, \bibinfo{person}{He Li}, {and}
  \bibinfo{person}{Bo Du}.} \bibinfo{year}{2023}\natexlab{}.
\newblock \showarticletitle{Rethinking Federated Learning With Domain Shift: A
  Prototype View}. In \bibinfo{booktitle}{\emph{CVPR}}.
\newblock


\bibitem[Huang et~al\mbox{.}(2021)]%
        {aaai2021fedamp}
\bibfield{author}{\bibinfo{person}{Yutao Huang}, \bibinfo{person}{Lingyang
  Chu}, \bibinfo{person}{Zirui Zhou}, \bibinfo{person}{Lanjun Wang},
  \bibinfo{person}{Jiangchuan Liu}, \bibinfo{person}{Jian Pei}, {and}
  \bibinfo{person}{Yong Zhang}.} \bibinfo{year}{2021}\natexlab{}.
\newblock \showarticletitle{Personalized cross-silo federated learning on
  non-iid data}. In \bibinfo{booktitle}{\emph{AAAI}}.
\newblock


\bibitem[Hyeon-Woo et~al\mbox{.}(2022)]%
        {iclr2022fedpara}
\bibfield{author}{\bibinfo{person}{Nam Hyeon-Woo}, \bibinfo{person}{Moon
  Ye-Bin}, {and} \bibinfo{person}{Tae-Hyun Oh}.}
  \bibinfo{year}{2022}\natexlab{}.
\newblock \showarticletitle{FedPara: Low-rank Hadamard Product for
  Communication-Efficient Federated Learning}. In
  \bibinfo{booktitle}{\emph{ICLR}}.
\newblock


\bibitem[Jaggi et~al\mbox{.}(2014)]%
        {nips2014cocoa1}
\bibfield{author}{\bibinfo{person}{Martin Jaggi}, \bibinfo{person}{Virginia
  Smith}, \bibinfo{person}{Martin Tak{\'a}c}, \bibinfo{person}{Jonathan
  Terhorst}, \bibinfo{person}{Sanjay Krishnan}, \bibinfo{person}{Thomas
  Hofmann}, {and} \bibinfo{person}{Michael~I Jordan}.}
  \bibinfo{year}{2014}\natexlab{}.
\newblock \showarticletitle{Communication-efficient distributed dual coordinate
  ascent}. In \bibinfo{booktitle}{\emph{NeurIPS}}.
\newblock


\bibitem[Jagielski et~al\mbox{.}(2019)]%
        {icml2019privatefair}
\bibfield{author}{\bibinfo{person}{Matthew Jagielski}, \bibinfo{person}{Michael
  Kearns}, \bibinfo{person}{Jieming Mao}, \bibinfo{person}{Alina Oprea},
  \bibinfo{person}{Aaron Roth}, \bibinfo{person}{Saeed Sharifi-Malvajerdi},
  {and} \bibinfo{person}{Jonathan Ullman}.} \bibinfo{year}{2019}\natexlab{}.
\newblock \showarticletitle{Differentially private fair learning}. In
  \bibinfo{booktitle}{\emph{ICML}}.
\newblock


\bibitem[Jeong et~al\mbox{.}(2018)]%
        {corr2018faug}
\bibfield{author}{\bibinfo{person}{Eunjeong Jeong}, \bibinfo{person}{Seungeun
  Oh}, \bibinfo{person}{Hyesung Kim}, \bibinfo{person}{Jihong Park},
  \bibinfo{person}{Mehdi Bennis}, {and} \bibinfo{person}{Seong-Lyun Kim}.}
  \bibinfo{year}{2018}\natexlab{}.
\newblock \showarticletitle{Communication-efficient on-device machine learning:
  Federated distillation and augmentation under non-iid private data}.
\newblock \bibinfo{journal}{\emph{arXiv preprint arXiv:1811.11479}}
  (\bibinfo{year}{2018}).
\newblock


\bibitem[Jia et~al\mbox{.}(2021)]%
        {cvpr2021intentonomy}
\bibfield{author}{\bibinfo{person}{Menglin Jia}, \bibinfo{person}{Zuxuan Wu},
  \bibinfo{person}{Austin Reiter}, \bibinfo{person}{Claire Cardie},
  \bibinfo{person}{Serge Belongie}, {and} \bibinfo{person}{Ser-Nam Lim}.}
  \bibinfo{year}{2021}\natexlab{}.
\newblock \showarticletitle{Intentonomy: a dataset and study towards human
  intent understanding}. In \bibinfo{booktitle}{\emph{CVPR}}.
\newblock


\bibitem[Jiang et~al\mbox{.}(2019)]%
        {Jiang2019fedmaml}
\bibfield{author}{\bibinfo{person}{Yihan Jiang}, \bibinfo{person}{Jakub
  Kone{\v{c}}n{\`y}}, \bibinfo{person}{Keith Rush}, {and}
  \bibinfo{person}{Sreeram Kannan}.} \bibinfo{year}{2019}\natexlab{}.
\newblock \showarticletitle{Improving federated learning personalization via
  model agnostic meta learning}.
\newblock \bibinfo{journal}{\emph{arXiv preprint arXiv:1909.12488}}
  (\bibinfo{year}{2019}).
\newblock


\bibitem[Kairouz et~al\mbox{.}(2021)]%
        {icml2021gaussian}
\bibfield{author}{\bibinfo{person}{Peter Kairouz}, \bibinfo{person}{Ziyu Liu},
  {and} \bibinfo{person}{Thomas Steinke}.} \bibinfo{year}{2021}\natexlab{}.
\newblock \showarticletitle{The distributed discrete gaussian mechanism for
  federated learning with secure aggregation}. In
  \bibinfo{booktitle}{\emph{ICML}}.
\newblock


\bibitem[Kairouz et~al\mbox{.}(2019)]%
        {hal2019advances}
\bibfield{author}{\bibinfo{person}{Peter Kairouz}, \bibinfo{person}{H~Brendan
  McMahan}, \bibinfo{person}{Brendan Avent}, \bibinfo{person}{Aur{\'e}lien
  Bellet}, \bibinfo{person}{Mehdi Bennis}, \bibinfo{person}{Arjun~Nitin
  Bhagoji}, \bibinfo{person}{Kallista Bonawitz}, \bibinfo{person}{Zachary
  Charles}, \bibinfo{person}{Graham Cormode}, \bibinfo{person}{Rachel
  Cummings}, {et~al\mbox{.}}} \bibinfo{year}{2019}\natexlab{}.
\newblock \showarticletitle{Advances and open problems in federated learning}.
\newblock \bibinfo{journal}{\emph{arXiv preprint arXiv:1912.04977}}
  (\bibinfo{year}{2019}).
\newblock


\bibitem[Kaissis et~al\mbox{.}(2021)]%
        {nature2021primia}
\bibfield{author}{\bibinfo{person}{Georgios Kaissis},
  \bibinfo{person}{Alexander Ziller}, \bibinfo{person}{Jonathan
  Passerat-Palmbach}, \bibinfo{person}{Th{\'e}o Ryffel},
  \bibinfo{person}{Dmitrii Usynin}, \bibinfo{person}{Andrew Trask},
  \bibinfo{person}{Ion{\'e}sio Lima}, \bibinfo{person}{Jason Mancuso},
  \bibinfo{person}{Friederike Jungmann}, \bibinfo{person}{Marc-Matthias
  Steinborn}, {et~al\mbox{.}}} \bibinfo{year}{2021}\natexlab{}.
\newblock \showarticletitle{End-to-end privacy preserving deep learning on
  multi-institutional medical imaging}.
\newblock \bibinfo{journal}{\emph{Nature Machine Intelligence}}
  (\bibinfo{year}{2021}).
\newblock


\bibitem[Kalra et~al\mbox{.}(2023)]%
        {nature2023decentralflproxy}
\bibfield{author}{\bibinfo{person}{Shivam Kalra}, \bibinfo{person}{Junfeng
  Wen}, \bibinfo{person}{Jesse~C Cresswell}, \bibinfo{person}{Maksims Volkovs},
  {and} \bibinfo{person}{HR Tizhoosh}.} \bibinfo{year}{2023}\natexlab{}.
\newblock \showarticletitle{Decentralized federated learning through proxy
  model sharing}.
\newblock \bibinfo{journal}{\emph{Nature communications}}
  (\bibinfo{year}{2023}).
\newblock


\bibitem[Karimireddy et~al\mbox{.}(2020)]%
        {icml2020scaffold}
\bibfield{author}{\bibinfo{person}{Sai~Praneeth Karimireddy},
  \bibinfo{person}{Satyen Kale}, \bibinfo{person}{Mehryar Mohri},
  \bibinfo{person}{Sashank Reddi}, \bibinfo{person}{Sebastian Stich}, {and}
  \bibinfo{person}{Ananda~Theertha Suresh}.} \bibinfo{year}{2020}\natexlab{}.
\newblock \showarticletitle{Scaffold: Stochastic controlled averaging for
  federated learning}. In \bibinfo{booktitle}{\emph{ICML}}.
\newblock


\bibitem[Khan et~al\mbox{.}(2021)]%
        {ieee2021iotfladvance}
\bibfield{author}{\bibinfo{person}{Latif~U Khan}, \bibinfo{person}{Walid Saad},
  \bibinfo{person}{Zhu Han}, \bibinfo{person}{Ekram Hossain}, {and}
  \bibinfo{person}{Choong~Seon Hong}.} \bibinfo{year}{2021}\natexlab{}.
\newblock \showarticletitle{Federated learning for internet of things: Recent
  advances, taxonomy, and open challenges}.
\newblock \bibinfo{journal}{\emph{IEEE Communications Surveys \& Tutorials}}
  (\bibinfo{year}{2021}).
\newblock


\bibitem[Khodak et~al\mbox{.}(2019)]%
        {nips2019aruba}
\bibfield{author}{\bibinfo{person}{Mikhail Khodak},
  \bibinfo{person}{Maria-Florina~F Balcan}, {and} \bibinfo{person}{Ameet~S
  Talwalkar}.} \bibinfo{year}{2019}\natexlab{}.
\newblock \showarticletitle{Adaptive gradient-based meta-learning methods}. In
  \bibinfo{booktitle}{\emph{NeurIPS}}.
\newblock


\bibitem[Kim et~al\mbox{.}(2022)]%
        {icml2022fedmlb}
\bibfield{author}{\bibinfo{person}{Jinkyu Kim}, \bibinfo{person}{Geeho Kim},
  {and} \bibinfo{person}{Bohyung Han}.} \bibinfo{year}{2022}\natexlab{}.
\newblock \showarticletitle{Multi-Level Branched Regularization for Federated
  Learning}. In \bibinfo{booktitle}{\emph{ICML}}.
\newblock


\bibitem[Kirkpatrick et~al\mbox{.}(2017)]%
        {pnas2017ewc}
\bibfield{author}{\bibinfo{person}{James Kirkpatrick}, \bibinfo{person}{Razvan
  Pascanu}, \bibinfo{person}{Neil Rabinowitz}, \bibinfo{person}{Joel Veness},
  \bibinfo{person}{Guillaume Desjardins}, \bibinfo{person}{Andrei~A Rusu},
  \bibinfo{person}{Kieran Milan}, \bibinfo{person}{John Quan},
  \bibinfo{person}{Tiago Ramalho}, \bibinfo{person}{Agnieszka
  Grabska-Barwinska}, {et~al\mbox{.}}} \bibinfo{year}{2017}\natexlab{}.
\newblock \showarticletitle{Overcoming catastrophic forgetting in neural
  networks}.
\newblock \bibinfo{journal}{\emph{PNAS}} (\bibinfo{year}{2017}).
\newblock


\bibitem[Kone{\v{c}}n{\`y} et~al\mbox{.}(2016)]%
        {konevcny2016comefficiency}
\bibfield{author}{\bibinfo{person}{Jakub Kone{\v{c}}n{\`y}},
  \bibinfo{person}{H~Brendan McMahan}, \bibinfo{person}{Felix~X Yu},
  \bibinfo{person}{Peter Richt{\'a}rik}, \bibinfo{person}{Ananda~Theertha
  Suresh}, {and} \bibinfo{person}{Dave Bacon}.}
  \bibinfo{year}{2016}\natexlab{}.
\newblock \showarticletitle{Federated learning: Strategies for improving
  communication efficiency}.
\newblock \bibinfo{journal}{\emph{arXiv preprint arXiv:1610.05492}}
  (\bibinfo{year}{2016}).
\newblock


\bibitem[Kopparapu and Lin(2020)]%
        {kopparapu2020fedfmc}
\bibfield{author}{\bibinfo{person}{Kavya Kopparapu} {and} \bibinfo{person}{Eric
  Lin}.} \bibinfo{year}{2020}\natexlab{}.
\newblock \showarticletitle{Fedfmc: Sequential efficient federated learning on
  non-iid data}.
\newblock \bibinfo{journal}{\emph{arXiv preprint arXiv:2006.10937}}
  (\bibinfo{year}{2020}).
\newblock


\bibitem[Kulkarni et~al\mbox{.}(2020)]%
        {kulkarni2020worlds4}
\bibfield{author}{\bibinfo{person}{Viraj Kulkarni}, \bibinfo{person}{Milind
  Kulkarni}, {and} \bibinfo{person}{Aniruddha Pant}.}
  \bibinfo{year}{2020}\natexlab{}.
\newblock \showarticletitle{Survey of personalization techniques for federated
  learning}. In \bibinfo{booktitle}{\emph{IEEE WorldS4}}.
\newblock


\bibitem[Laguel et~al\mbox{.}(2020)]%
        {laguel2020superquantile}
\bibfield{author}{\bibinfo{person}{Yassine Laguel}, \bibinfo{person}{Krishna
  Pillutla}, \bibinfo{person}{J{\'e}r{\^o}me Malick}, {and}
  \bibinfo{person}{Zaid Harchaoui}.} \bibinfo{year}{2020}\natexlab{}.
\newblock \showarticletitle{Device heterogeneity in federated learning: A
  superquantile approach}.
\newblock \bibinfo{journal}{\emph{arXiv preprint arXiv:2002.11223}}
  (\bibinfo{year}{2020}).
\newblock


\bibitem[Laguel et~al\mbox{.}(2021)]%
        {ciss2021superquantile}
\bibfield{author}{\bibinfo{person}{Yassine Laguel}, \bibinfo{person}{Krishna
  Pillutla}, \bibinfo{person}{J{\'e}r{\^o}me Malick}, {and}
  \bibinfo{person}{Zaid Harchaoui}.} \bibinfo{year}{2021}\natexlab{}.
\newblock \showarticletitle{A superquantile approach to federated learning with
  heterogeneous devices}. In \bibinfo{booktitle}{\emph{IEEE CISS}}.
\newblock


\bibitem[Lai et~al\mbox{.}(2022)]%
        {icml2022fedscale}
\bibfield{author}{\bibinfo{person}{Fan Lai}, \bibinfo{person}{Yinwei Dai},
  \bibinfo{person}{Xiangfeng Zhu}, \bibinfo{person}{Harsha~V Madhyastha}, {and}
  \bibinfo{person}{Mosharaf Chowdhury}.} \bibinfo{year}{2022}\natexlab{}.
\newblock \showarticletitle{FedScale: Benchmarking Model and System Performance
  of Federated Learning at Scale}. In \bibinfo{booktitle}{\emph{ICML}}.
\newblock


\bibitem[Lalitha et~al\mbox{.}(2018)]%
        {nips2018fullydecentralfl}
\bibfield{author}{\bibinfo{person}{Anusha Lalitha}, \bibinfo{person}{Shubhanshu
  Shekhar}, \bibinfo{person}{Tara Javidi}, {and} \bibinfo{person}{Farinaz
  Koushanfar}.} \bibinfo{year}{2018}\natexlab{}.
\newblock \showarticletitle{Fully decentralized federated learning}. In
  \bibinfo{booktitle}{\emph{NeurIPS Workshop}}.
\newblock


\bibitem[Li et~al\mbox{.}(2021h)]%
        {infocom2021sampleselect}
\bibfield{author}{\bibinfo{person}{Anran Li}, \bibinfo{person}{Lan Zhang},
  \bibinfo{person}{Juntao Tan}, \bibinfo{person}{Yaxuan Qin},
  \bibinfo{person}{Junhao Wang}, {and} \bibinfo{person}{Xiang-Yang Li}.}
  \bibinfo{year}{2021}\natexlab{h}.
\newblock \showarticletitle{Sample-level data selection for federated
  learning}. In \bibinfo{booktitle}{\emph{IEEE INFOCOM}}.
\newblock


\bibitem[Li et~al\mbox{.}(2021f)]%
        {iot2020defkt}
\bibfield{author}{\bibinfo{person}{Chengxi Li}, \bibinfo{person}{Gang Li},
  {and} \bibinfo{person}{Pramod~K Varshney}.} \bibinfo{year}{2021}\natexlab{f}.
\newblock \showarticletitle{Decentralized federated learning via mutual
  knowledge transfer}.
\newblock \bibinfo{journal}{\emph{IEEE IoT}} (\bibinfo{year}{2021}).
\newblock


\bibitem[Li and Wang(2019)]%
        {nips2019fedmd}
\bibfield{author}{\bibinfo{person}{Daliang Li} {and} \bibinfo{person}{Junpu
  Wang}.} \bibinfo{year}{2019}\natexlab{}.
\newblock \showarticletitle{Fedmd: Heterogenous federated learning via model
  distillation}. In \bibinfo{booktitle}{\emph{NeurIPS Workshop}}.
\newblock


\bibitem[Li et~al\mbox{.}(2022)]%
        {cvpr2022ressfl}
\bibfield{author}{\bibinfo{person}{Jingtao Li}, \bibinfo{person}{Adnan~Siraj
  Rakin}, \bibinfo{person}{Xing Chen}, \bibinfo{person}{Zhezhi He},
  \bibinfo{person}{Deliang Fan}, {and} \bibinfo{person}{Chaitali Chakrabarti}.}
  \bibinfo{year}{2022}\natexlab{}.
\newblock \showarticletitle{ResSFL: A Resistance Transfer Framework for
  Defending Model Inversion Attack in Split Federated Learning}. In
  \bibinfo{booktitle}{\emph{CVPR}}.
\newblock


\bibitem[Li et~al\mbox{.}(2021b)]%
        {ijcnn2021fedsae}
\bibfield{author}{\bibinfo{person}{Li Li}, \bibinfo{person}{Moming Duan},
  \bibinfo{person}{Duo Liu}, \bibinfo{person}{Yu Zhang}, \bibinfo{person}{Ao
  Ren}, \bibinfo{person}{Xianzhang Chen}, \bibinfo{person}{Yujuan Tan}, {and}
  \bibinfo{person}{Chengliang Wang}.} \bibinfo{year}{2021}\natexlab{b}.
\newblock \showarticletitle{FedSAE: A novel self-adaptive federated learning
  framework in heterogeneous systems}. In \bibinfo{booktitle}{\emph{IJCNN}}.
\newblock


\bibitem[Li et~al\mbox{.}(2020f)]%
        {iwqos2020mcfl}
\bibfield{author}{\bibinfo{person}{Li Li}, \bibinfo{person}{Jun Wang},
  \bibinfo{person}{Xu Chen}, {and} \bibinfo{person}{Cheng-Zhong Xu}.}
  \bibinfo{year}{2020}\natexlab{f}.
\newblock \showarticletitle{Multi-layer coordination for high-performance
  energy-efficient federated learning}. In \bibinfo{booktitle}{\emph{IWQoS}}.
\newblock


\bibitem[Li et~al\mbox{.}(2021a)]%
        {li2021noniidsilos}
\bibfield{author}{\bibinfo{person}{Qinbin Li}, \bibinfo{person}{Yiqun Diao},
  \bibinfo{person}{Quan Chen}, {and} \bibinfo{person}{Bingsheng He}.}
  \bibinfo{year}{2021}\natexlab{a}.
\newblock \showarticletitle{Federated learning on non-iid data silos: An
  experimental study}.
\newblock \bibinfo{journal}{\emph{arXiv preprint arXiv:2102.02079}}
  (\bibinfo{year}{2021}).
\newblock


\bibitem[Li et~al\mbox{.}(2021c)]%
        {cvpr2021moon}
\bibfield{author}{\bibinfo{person}{Qinbin Li}, \bibinfo{person}{Bingsheng He},
  {and} \bibinfo{person}{Dawn Song}.} \bibinfo{year}{2021}\natexlab{c}.
\newblock \showarticletitle{Model-Contrastive Federated Learning}. In
  \bibinfo{booktitle}{\emph{CVPR}}.
\newblock


\bibitem[Li et~al\mbox{.}(2021g)]%
        {tkde2021surveyfl}
\bibfield{author}{\bibinfo{person}{Qinbin Li}, \bibinfo{person}{Zeyi Wen},
  \bibinfo{person}{Zhaomin Wu}, \bibinfo{person}{Sixu Hu},
  \bibinfo{person}{Naibo Wang}, \bibinfo{person}{Yuan Li}, \bibinfo{person}{Xu
  Liu}, {and} \bibinfo{person}{Bingsheng He}.}
  \bibinfo{year}{2021}\natexlab{g}.
\newblock \showarticletitle{A survey on federated learning systems: vision,
  hype and reality for data privacy and protection}.
\newblock \bibinfo{journal}{\emph{IEEE TKDE}} (\bibinfo{year}{2021}).
\newblock


\bibitem[Li et~al\mbox{.}(2019)]%
        {bigdata2019ofmtl}
\bibfield{author}{\bibinfo{person}{Rui Li}, \bibinfo{person}{Fenglong Ma},
  \bibinfo{person}{Wenjun Jiang}, {and} \bibinfo{person}{Jing Gao}.}
  \bibinfo{year}{2019}\natexlab{}.
\newblock \showarticletitle{Online federated multitask learning}. In
  \bibinfo{booktitle}{\emph{IEEE Big Data}}.
\newblock


\bibitem[Li et~al\mbox{.}(2020b)]%
        {li2020detectmalicious}
\bibfield{author}{\bibinfo{person}{Suyi Li}, \bibinfo{person}{Yong Cheng},
  \bibinfo{person}{Wei Wang}, \bibinfo{person}{Yang Liu}, {and}
  \bibinfo{person}{Tianjian Chen}.} \bibinfo{year}{2020}\natexlab{b}.
\newblock \showarticletitle{Learning to detect malicious clients for robust
  federated learning}.
\newblock \bibinfo{journal}{\emph{arXiv preprint arXiv:2002.00211}}
  (\bibinfo{year}{2020}).
\newblock


\bibitem[Li et~al\mbox{.}(2021d)]%
        {icml2021ditto}
\bibfield{author}{\bibinfo{person}{Tian Li}, \bibinfo{person}{Shengyuan Hu},
  \bibinfo{person}{Ahmad Beirami}, {and} \bibinfo{person}{Virginia Smith}.}
  \bibinfo{year}{2021}\natexlab{d}.
\newblock \showarticletitle{Ditto: Fair and robust federated learning through
  personalization}. In \bibinfo{booktitle}{\emph{ICML}}.
\newblock


\bibitem[Li et~al\mbox{.}(2020c)]%
        {tli2020flchallenges}
\bibfield{author}{\bibinfo{person}{Tian Li}, \bibinfo{person}{Anit~Kumar Sahu},
  \bibinfo{person}{Ameet Talwalkar}, {and} \bibinfo{person}{Virginia Smith}.}
  \bibinfo{year}{2020}\natexlab{c}.
\newblock \showarticletitle{Federated Learning: Challenges, Methods, and Future
  Directions}.
\newblock \bibinfo{journal}{\emph{IEEE Signal Process Mag}}
  (\bibinfo{year}{2020}).
\newblock


\bibitem[Li et~al\mbox{.}(2020d)]%
        {li2020fedprox}
\bibfield{author}{\bibinfo{person}{Tian Li}, \bibinfo{person}{Anit~Kumar Sahu},
  \bibinfo{person}{Manzil Zaheer}, \bibinfo{person}{Maziar Sanjabi},
  \bibinfo{person}{Ameet Talwalkar}, {and} \bibinfo{person}{Virginia Smith}.}
  \bibinfo{year}{2020}\natexlab{d}.
\newblock \showarticletitle{Federated optimization in heterogeneous networks}.
  In \bibinfo{booktitle}{\emph{MLSys}}.
\newblock


\bibitem[Li et~al\mbox{.}(2020e)]%
        {iclr2020qfedavg}
\bibfield{author}{\bibinfo{person}{Tian Li}, \bibinfo{person}{Maziar Sanjabi},
  \bibinfo{person}{Ahmad Beirami}, {and} \bibinfo{person}{Virginia Smith}.}
  \bibinfo{year}{2020}\natexlab{e}.
\newblock \showarticletitle{Fair Resource Allocation in Federated Learning}. In
  \bibinfo{booktitle}{\emph{ICLR}}.
\newblock


\bibitem[Li et~al\mbox{.}(2021e)]%
        {iclr2021fedbn}
\bibfield{author}{\bibinfo{person}{Xiaoxiao Li}, \bibinfo{person}{Meirui
  Jiang}, \bibinfo{person}{Xiaofei Zhang}, \bibinfo{person}{Michael Kamp},
  {and} \bibinfo{person}{Qi Dou}.} \bibinfo{year}{2021}\natexlab{e}.
\newblock \showarticletitle{Fedbn: Federated learning on non-iid features via
  local batch normalization}. In \bibinfo{booktitle}{\emph{ICLR}}.
\newblock


\bibitem[Li et~al\mbox{.}(2020a)]%
        {network2020blockchaindfl}
\bibfield{author}{\bibinfo{person}{Yuzheng Li}, \bibinfo{person}{Chuan Chen},
  \bibinfo{person}{Nan Liu}, \bibinfo{person}{Huawei Huang},
  \bibinfo{person}{Zibin Zheng}, {and} \bibinfo{person}{Qiang Yan}.}
  \bibinfo{year}{2020}\natexlab{a}.
\newblock \showarticletitle{A blockchain-based decentralized federated learning
  framework with committee consensus}.
\newblock \bibinfo{journal}{\emph{IEEE Network}} (\bibinfo{year}{2020}).
\newblock


\bibitem[Liang et~al\mbox{.}(2020)]%
        {liang2020lgfedavg}
\bibfield{author}{\bibinfo{person}{Paul~Pu Liang}, \bibinfo{person}{Terrance
  Liu}, \bibinfo{person}{Liu Ziyin}, \bibinfo{person}{Nicholas~B Allen},
  \bibinfo{person}{Randy~P Auerbach}, \bibinfo{person}{David Brent},
  \bibinfo{person}{Ruslan Salakhutdinov}, {and} \bibinfo{person}{Louis-Philippe
  Morency}.} \bibinfo{year}{2020}\natexlab{}.
\newblock \showarticletitle{Think locally, act globally: Federated learning
  with local and global representations}.
\newblock \bibinfo{journal}{\emph{arXiv preprint arXiv:2001.01523}}
  (\bibinfo{year}{2020}).
\newblock


\bibitem[Lim et~al\mbox{.}(2020)]%
        {lim2020flmobileedge}
\bibfield{author}{\bibinfo{person}{Wei Yang~Bryan Lim},
  \bibinfo{person}{Nguyen~Cong Luong}, \bibinfo{person}{Dinh~Thai Hoang},
  \bibinfo{person}{Yutao Jiao}, \bibinfo{person}{Ying-Chang Liang},
  \bibinfo{person}{Qiang Yang}, \bibinfo{person}{Dusit Niyato}, {and}
  \bibinfo{person}{Chunyan Miao}.} \bibinfo{year}{2020}\natexlab{}.
\newblock \showarticletitle{Federated learning in mobile edge networks: A
  comprehensive survey}.
\newblock \bibinfo{journal}{\emph{IEEE Communications Surveys \& Tutorials}}
  (\bibinfo{year}{2020}).
\newblock


\bibitem[Lim et~al\mbox{.}(2021a)]%
        {tpds2021clusterselect}
\bibfield{author}{\bibinfo{person}{Wei Yang~Bryan Lim},
  \bibinfo{person}{Jer~Shyuan Ng}, \bibinfo{person}{Zehui Xiong},
  \bibinfo{person}{Jiangming Jin}, \bibinfo{person}{Yang Zhang},
  \bibinfo{person}{Dusit Niyato}, \bibinfo{person}{Cyril Leung}, {and}
  \bibinfo{person}{Chunyan Miao}.} \bibinfo{year}{2021}\natexlab{a}.
\newblock \showarticletitle{Decentralized edge intelligence: A dynamic resource
  allocation framework for hierarchical federated learning}.
\newblock \bibinfo{journal}{\emph{IEEE TPDS}} (\bibinfo{year}{2021}).
\newblock


\bibitem[Lim et~al\mbox{.}(2021b)]%
        {jsac2021dynamichfl}
\bibfield{author}{\bibinfo{person}{Wei Yang~Bryan Lim},
  \bibinfo{person}{Jer~Shyuan Ng}, \bibinfo{person}{Zehui Xiong},
  \bibinfo{person}{Dusit Niyato}, \bibinfo{person}{Chunyan Miao}, {and}
  \bibinfo{person}{Dong~In Kim}.} \bibinfo{year}{2021}\natexlab{b}.
\newblock \showarticletitle{Dynamic edge association and resource allocation in
  self-organizing hierarchical federated learning networks}.
\newblock \bibinfo{journal}{\emph{IEEE JSAC}} (\bibinfo{year}{2021}).
\newblock


\bibitem[Lin et~al\mbox{.}(2020b)]%
        {icdcs2020fedml}
\bibfield{author}{\bibinfo{person}{Sen Lin}, \bibinfo{person}{Guang Yang},
  {and} \bibinfo{person}{Junshan Zhang}.} \bibinfo{year}{2020}\natexlab{b}.
\newblock \showarticletitle{A collaborative learning framework via federated
  meta-learning}. In \bibinfo{booktitle}{\emph{IEEE ICDCS}}.
\newblock


\bibitem[Lin et~al\mbox{.}(2020a)]%
        {nips2020feddf}
\bibfield{author}{\bibinfo{person}{Tao Lin}, \bibinfo{person}{Lingjing Kong},
  \bibinfo{person}{Sebastian~U Stich}, {and} \bibinfo{person}{Martin Jaggi}.}
  \bibinfo{year}{2020}\natexlab{a}.
\newblock \showarticletitle{Ensemble distillation for robust model fusion in
  federated learning}. In \bibinfo{booktitle}{\emph{NeurIPS}}.
\newblock


\bibitem[Liu et~al\mbox{.}(2020b)]%
        {icc2020hierfavg}
\bibfield{author}{\bibinfo{person}{Lumin Liu}, \bibinfo{person}{Jun Zhang},
  \bibinfo{person}{SH Song}, {and} \bibinfo{person}{Khaled~B Letaief}.}
  \bibinfo{year}{2020}\natexlab{b}.
\newblock \showarticletitle{Client-edge-cloud hierarchical federated learning}.
  In \bibinfo{booktitle}{\emph{IEEE ICC}}.
\newblock


\bibitem[Liu et~al\mbox{.}(2022)]%
        {jsa2022hybrid}
\bibfield{author}{\bibinfo{person}{Wenyan Liu}, \bibinfo{person}{Junhong
  Cheng}, \bibinfo{person}{Xiaoling Wang}, \bibinfo{person}{Xingjian Lu}, {and}
  \bibinfo{person}{Jianwei Yin}.} \bibinfo{year}{2022}\natexlab{}.
\newblock \showarticletitle{Hybrid differential privacy based federated
  learning for Internet of Things}.
\newblock \bibinfo{journal}{\emph{JSA}} (\bibinfo{year}{2022}).
\newblock


\bibitem[Liu et~al\mbox{.}(2020a)]%
        {icml2020backdoorattckincl}
\bibfield{author}{\bibinfo{person}{Yang Liu}, \bibinfo{person}{Zhihao Yi},
  {and} \bibinfo{person}{Tianjian Chen}.} \bibinfo{year}{2020}\natexlab{a}.
\newblock \showarticletitle{Backdoor attacks and defenses in
  feature-partitioned collaborative learning}. In
  \bibinfo{booktitle}{\emph{FL-ICML workshop}}.
\newblock


\bibitem[Lubana et~al\mbox{.}(2022)]%
        {icml2022orchestra}
\bibfield{author}{\bibinfo{person}{Ekdeep~Singh Lubana},
  \bibinfo{person}{Chi~Ian Tang}, \bibinfo{person}{Fahim Kawsar},
  \bibinfo{person}{Robert~P Dick}, {and} \bibinfo{person}{Akhil Mathur}.}
  \bibinfo{year}{2022}\natexlab{}.
\newblock \showarticletitle{Orchestra: Unsupervised Federated Learning via
  Globally Consistent Clustering}.
\newblock  (\bibinfo{year}{2022}).
\newblock


\bibitem[Luo et~al\mbox{.}(2019)]%
        {realworldfldataset}
\bibfield{author}{\bibinfo{person}{Jiahuan Luo}, \bibinfo{person}{Xueyang Wu},
  \bibinfo{person}{Yun Luo}, \bibinfo{person}{Anbu Huang},
  \bibinfo{person}{Yunfeng Huang}, \bibinfo{person}{Yang Liu}, {and}
  \bibinfo{person}{Qiang Yang}.} \bibinfo{year}{2019}\natexlab{}.
\newblock \showarticletitle{Real-world image datasets for federated learning}.
\newblock  (\bibinfo{year}{2019}).
\newblock


\bibitem[Luo et~al\mbox{.}(2021)]%
        {nips2021ccvr}
\bibfield{author}{\bibinfo{person}{Mi Luo}, \bibinfo{person}{Fei Chen},
  \bibinfo{person}{Dapeng Hu}, \bibinfo{person}{Yifan Zhang},
  \bibinfo{person}{Jian Liang}, {and} \bibinfo{person}{Jiashi Feng}.}
  \bibinfo{year}{2021}\natexlab{}.
\newblock \showarticletitle{No fear of heterogeneity: Classifier calibration
  for federated learning with non-iid data}. In
  \bibinfo{booktitle}{\emph{NeurIPS}}.
\newblock


\bibitem[Luo et~al\mbox{.}(2022)]%
        {icml2022dfl}
\bibfield{author}{\bibinfo{person}{Zhengquan Luo}, \bibinfo{person}{Yunlong
  Wang}, \bibinfo{person}{Zilei Wang}, \bibinfo{person}{Zhenan Sun}, {and}
  \bibinfo{person}{Tieniu Tan}.} \bibinfo{year}{2022}\natexlab{}.
\newblock \showarticletitle{Disentangled Federated Learning for Tackling
  Attributes Skew via Invariant Aggregation and Diversity Transferring}. In
  \bibinfo{booktitle}{\emph{ICML}}.
\newblock


\bibitem[Luping et~al\mbox{.}(2019)]%
        {icdcs2019cmfl}
\bibfield{author}{\bibinfo{person}{Wang Luping}, \bibinfo{person}{Wang Wei},
  {and} \bibinfo{person}{LI Bo}.} \bibinfo{year}{2019}\natexlab{}.
\newblock \showarticletitle{CMFL: Mitigating communication overhead for
  federated learning}. In \bibinfo{booktitle}{\emph{IEEE ICDCS}}.
\newblock


\bibitem[Lyu et~al\mbox{.}(2020a)]%
        {flijcai2020fairnessfl}
\bibfield{author}{\bibinfo{person}{Lingjuan Lyu}, \bibinfo{person}{Xinyi Xu},
  {and} \bibinfo{person}{Qian Wang}.} \bibinfo{year}{2020}\natexlab{a}.
\newblock \showarticletitle{Collaborative Fairness in Federated Learning}. In
  \bibinfo{booktitle}{\emph{FL-IJCAI'20 workshop}}.
\newblock


\bibitem[Lyu et~al\mbox{.}(2020c)]%
        {lyu2020privacyrobust}
\bibfield{author}{\bibinfo{person}{Lingjuan Lyu}, \bibinfo{person}{Han Yu},
  \bibinfo{person}{Xingjun Ma}, \bibinfo{person}{Lichao Sun},
  \bibinfo{person}{Jun Zhao}, \bibinfo{person}{Qiang Yang}, {and}
  \bibinfo{person}{Philip~S Yu}.} \bibinfo{year}{2020}\natexlab{c}.
\newblock \showarticletitle{Privacy and robustness in federated learning:
  Attacks and defenses}.
\newblock \bibinfo{journal}{\emph{arXiv preprint arXiv:2012.06337}}
  (\bibinfo{year}{2020}).
\newblock


\bibitem[Lyu et~al\mbox{.}(2020b)]%
        {lyu2020attacksurvey}
\bibfield{author}{\bibinfo{person}{Lingjuan Lyu}, \bibinfo{person}{Han Yu},
  {and} \bibinfo{person}{Qiang Yang}.} \bibinfo{year}{2020}\natexlab{b}.
\newblock \showarticletitle{Threats to federated learning: A survey}.
\newblock \bibinfo{journal}{\emph{arXiv preprint arXiv:2003.02133}}
  (\bibinfo{year}{2020}).
\newblock


\bibitem[Lyu et~al\mbox{.}(2020d)]%
        {tpds2020fppdl}
\bibfield{author}{\bibinfo{person}{Lingjuan Lyu}, \bibinfo{person}{Jiangshan
  Yu}, \bibinfo{person}{Karthik Nandakumar}, \bibinfo{person}{Yitong Li},
  \bibinfo{person}{Xingjun Ma}, \bibinfo{person}{Jiong Jin},
  \bibinfo{person}{Han Yu}, {and} \bibinfo{person}{Kee~Siong Ng}.}
  \bibinfo{year}{2020}\natexlab{d}.
\newblock \showarticletitle{Towards fair and privacy-preserving federated deep
  models}. In \bibinfo{booktitle}{\emph{TPDS}}.
\newblock


\bibitem[Ma et~al\mbox{.}(2015)]%
        {icml2015cocoa2}
\bibfield{author}{\bibinfo{person}{Chenxin Ma}, \bibinfo{person}{Virginia
  Smith}, \bibinfo{person}{Martin Jaggi}, \bibinfo{person}{Michael Jordan},
  \bibinfo{person}{Peter Richt{\'a}rik}, {and} \bibinfo{person}{Martin
  Tak{\'a}c}.} \bibinfo{year}{2015}\natexlab{}.
\newblock \showarticletitle{Adding vs. averaging in distributed primal-dual
  optimization}. In \bibinfo{booktitle}{\emph{ICML}}.
\newblock


\bibitem[Ma et~al\mbox{.}(2022)]%
        {cvpr2022pfedla}
\bibfield{author}{\bibinfo{person}{Xiaosong Ma}, \bibinfo{person}{Jie Zhang},
  \bibinfo{person}{Song Guo}, {and} \bibinfo{person}{Wenchao Xu}.}
  \bibinfo{year}{2022}\natexlab{}.
\newblock \showarticletitle{Layer-wised Model Aggregation for Personalized
  Federated Learning}. In \bibinfo{booktitle}{\emph{CVPR}}.
\newblock


\bibitem[Mai et~al\mbox{.}(2020)]%
        {tifs2020secureface}
\bibfield{author}{\bibinfo{person}{Guangcan Mai}, \bibinfo{person}{Kai Cao},
  \bibinfo{person}{Xiangyuan Lan}, {and} \bibinfo{person}{Pong~C Yuen}.}
  \bibinfo{year}{2020}\natexlab{}.
\newblock \showarticletitle{Secureface: Face template protection}.
\newblock \bibinfo{journal}{\emph{IEEE TIFS}}  \bibinfo{volume}{16}
  (\bibinfo{year}{2020}), \bibinfo{pages}{262--277}.
\newblock


\bibitem[Mai et~al\mbox{.}(2018)]%
        {mai2018reconstruction}
\bibfield{author}{\bibinfo{person}{Guangcan Mai}, \bibinfo{person}{Kai Cao},
  \bibinfo{person}{Pong~C Yuen}, {and} \bibinfo{person}{Anil~K Jain}.}
  \bibinfo{year}{2018}\natexlab{}.
\newblock \showarticletitle{On the reconstruction of face images from deep face
  templates}.
\newblock \bibinfo{journal}{\emph{IEEE TPAMI}} \bibinfo{volume}{41},
  \bibinfo{number}{5} (\bibinfo{year}{2018}), \bibinfo{pages}{1188--1202}.
\newblock


\bibitem[Makhija et~al\mbox{.}(2022)]%
        {icml2022fedhenn}
\bibfield{author}{\bibinfo{person}{Disha Makhija}, \bibinfo{person}{Xing Han},
  \bibinfo{person}{Nhat Ho}, {and} \bibinfo{person}{Joydeep Ghosh}.}
  \bibinfo{year}{2022}\natexlab{}.
\newblock \showarticletitle{Architecture Agnostic Federated Learning for Neural
  Networks}. In \bibinfo{booktitle}{\emph{ICML}}.
\newblock


\bibitem[Mao et~al\mbox{.}(2021)]%
        {esorics2021romoa}
\bibfield{author}{\bibinfo{person}{Yunlong Mao}, \bibinfo{person}{Xinyu Yuan},
  \bibinfo{person}{Xinyang Zhao}, {and} \bibinfo{person}{Sheng Zhong}.}
  \bibinfo{year}{2021}\natexlab{}.
\newblock \showarticletitle{Romoa: Robust model aggregation for the resistance
  of federated learning to model poisoning attacks}. In
  \bibinfo{booktitle}{\emph{ESORICS}}.
\newblock


\bibitem[Marfoq et~al\mbox{.}(2021)]%
        {nips2021fedem}
\bibfield{author}{\bibinfo{person}{Othmane Marfoq}, \bibinfo{person}{Giovanni
  Neglia}, \bibinfo{person}{Aur{\'e}lien Bellet}, \bibinfo{person}{Laetitia
  Kameni}, {and} \bibinfo{person}{Richard Vidal}.}
  \bibinfo{year}{2021}\natexlab{}.
\newblock \showarticletitle{Federated multi-task learning under a mixture of
  distributions}. In \bibinfo{booktitle}{\emph{NeurIPS}}.
\newblock


\bibitem[Marfoq et~al\mbox{.}(2022)]%
        {icml2022knnper}
\bibfield{author}{\bibinfo{person}{Othmane Marfoq}, \bibinfo{person}{Giovanni
  Neglia}, \bibinfo{person}{Richard Vidal}, {and} \bibinfo{person}{Laetitia
  Kameni}.} \bibinfo{year}{2022}\natexlab{}.
\newblock \showarticletitle{Personalized Federated Learning through Local
  Memorization}. In \bibinfo{booktitle}{\emph{ICML}}.
\newblock


\bibitem[McCloskey and Cohen(1989)]%
        {mccloskey1989cataforget}
\bibfield{author}{\bibinfo{person}{Michael McCloskey} {and}
  \bibinfo{person}{Neal~J Cohen}.} \bibinfo{year}{1989}\natexlab{}.
\newblock \showarticletitle{Catastrophic interference in connectionist
  networks: The sequential learning problem}.
\newblock  (\bibinfo{year}{1989}).
\newblock


\bibitem[McMahan et~al\mbox{.}(2017)]%
        {mcmahan2017fedavg}
\bibfield{author}{\bibinfo{person}{Brendan McMahan}, \bibinfo{person}{Eider
  Moore}, \bibinfo{person}{Daniel Ramage}, \bibinfo{person}{Seth Hampson},
  {and} \bibinfo{person}{Blaise~Aguera y Arcas}.}
  \bibinfo{year}{2017}\natexlab{}.
\newblock \showarticletitle{Communication-efficient learning of deep networks
  from decentralized data}. In \bibinfo{booktitle}{\emph{AISTATS}}.
\newblock


\bibitem[McMahan et~al\mbox{.}(2018)]%
        {iclr2018dpfedavg}
\bibfield{author}{\bibinfo{person}{H~Brendan McMahan}, \bibinfo{person}{Daniel
  Ramage}, \bibinfo{person}{Kunal Talwar}, {and} \bibinfo{person}{Li Zhang}.}
  \bibinfo{year}{2018}\natexlab{}.
\newblock \showarticletitle{Learning Differentially Private Recurrent Language
  Models}. In \bibinfo{booktitle}{\emph{ICLR}}.
\newblock


\bibitem[Mo et~al\mbox{.}(2021)]%
        {mobisys2021ppfl}
\bibfield{author}{\bibinfo{person}{Fan Mo}, \bibinfo{person}{Hamed Haddadi},
  \bibinfo{person}{Kleomenis Katevas}, \bibinfo{person}{Eduard Marin},
  \bibinfo{person}{Diego Perino}, {and} \bibinfo{person}{Nicolas Kourtellis}.}
  \bibinfo{year}{2021}\natexlab{}.
\newblock \showarticletitle{PPFL: privacy-preserving federated learning with
  trusted execution environments}. In \bibinfo{booktitle}{\emph{MobiSys}}.
\newblock


\bibitem[Mohri et~al\mbox{.}(2019)]%
        {icml2019agnosticfl}
\bibfield{author}{\bibinfo{person}{Mehryar Mohri}, \bibinfo{person}{Gary
  Sivek}, {and} \bibinfo{person}{Ananda~Theertha Suresh}.}
  \bibinfo{year}{2019}\natexlab{}.
\newblock \showarticletitle{Agnostic federated learning}. In
  \bibinfo{booktitle}{\emph{ICML}}.
\newblock


\bibitem[Mothukuri et~al\mbox{.}(2021a)]%
        {iot2021detedctsecurityattack}
\bibfield{author}{\bibinfo{person}{Viraaji Mothukuri}, \bibinfo{person}{Prachi
  Khare}, \bibinfo{person}{Reza~M Parizi}, \bibinfo{person}{Seyedamin
  Pouriyeh}, \bibinfo{person}{Ali Dehghantanha}, {and} \bibinfo{person}{Gautam
  Srivastava}.} \bibinfo{year}{2021}\natexlab{a}.
\newblock \showarticletitle{Federated-Learning-Based Anomaly Detection for IoT
  Security Attacks}.
\newblock \bibinfo{journal}{\emph{IEEE IoT}} (\bibinfo{year}{2021}).
\newblock


\bibitem[Mothukuri et~al\mbox{.}(2021b)]%
        {fgcs2021privacysurvey}
\bibfield{author}{\bibinfo{person}{Viraaji Mothukuri}, \bibinfo{person}{Reza~M
  Parizi}, \bibinfo{person}{Seyedamin Pouriyeh}, \bibinfo{person}{Yan Huang},
  \bibinfo{person}{Ali Dehghantanha}, {and} \bibinfo{person}{Gautam
  Srivastava}.} \bibinfo{year}{2021}\natexlab{b}.
\newblock \showarticletitle{A survey on security and privacy of federated
  learning}.
\newblock \bibinfo{journal}{\emph{FGCS}} (\bibinfo{year}{2021}).
\newblock


\bibitem[Mu et~al\mbox{.}(2021)]%
        {mu2021fedproc}
\bibfield{author}{\bibinfo{person}{Xutong Mu}, \bibinfo{person}{Yulong Shen},
  \bibinfo{person}{Ke Cheng}, \bibinfo{person}{Xueli Geng},
  \bibinfo{person}{Jiaxuan Fu}, \bibinfo{person}{Tao Zhang}, {and}
  \bibinfo{person}{Zhiwei Zhang}.} \bibinfo{year}{2021}\natexlab{}.
\newblock \showarticletitle{FedProc: Prototypical Contrastive Federated
  Learning on Non-IID data}.
\newblock \bibinfo{journal}{\emph{arXiv preprint arXiv:2109.12273}}
  (\bibinfo{year}{2021}).
\newblock


\bibitem[Nguyen et~al\mbox{.}(2021)]%
        {ieee2021iotflsurvey}
\bibfield{author}{\bibinfo{person}{Dinh~C Nguyen}, \bibinfo{person}{Ming Ding},
  \bibinfo{person}{Pubudu~N Pathirana}, \bibinfo{person}{Aruna Seneviratne},
  \bibinfo{person}{Jun Li}, {and} \bibinfo{person}{H~Vincent Poor}.}
  \bibinfo{year}{2021}\natexlab{}.
\newblock \showarticletitle{Federated learning for internet of things: A
  comprehensive survey}.
\newblock \bibinfo{journal}{\emph{IEEE Communications Surveys \& Tutorials}}
  (\bibinfo{year}{2021}).
\newblock


\bibitem[Nguyen et~al\mbox{.}(2022b)]%
        {usenix2022flame}
\bibfield{author}{\bibinfo{person}{Thien~Duc Nguyen}, \bibinfo{person}{Phillip
  Rieger}, \bibinfo{person}{Huili Chen}, \bibinfo{person}{Hossein Yalame},
  \bibinfo{person}{Helen M{\"o}llering}, \bibinfo{person}{Hossein Fereidooni},
  \bibinfo{person}{Samuel Marchal}, \bibinfo{person}{Markus Miettinen},
  \bibinfo{person}{Azalia Mirhoseini}, \bibinfo{person}{Shaza Zeitouni},
  {et~al\mbox{.}}} \bibinfo{year}{2022}\natexlab{b}.
\newblock \showarticletitle{FLAME: Taming Backdoors in Federated Learning}. In
  \bibinfo{booktitle}{\emph{USENIX Security}}.
\newblock


\bibitem[Nguyen et~al\mbox{.}(2022a)]%
        {ieee2022fedfog}
\bibfield{author}{\bibinfo{person}{Van-Dinh Nguyen}, \bibinfo{person}{Symeon
  Chatzinotas}, \bibinfo{person}{Bj{\"o}rn Ottersten}, {and}
  \bibinfo{person}{Trung~Q Duong}.} \bibinfo{year}{2022}\natexlab{a}.
\newblock \showarticletitle{FedFog: Network-aware optimization of federated
  learning over wireless fog-cloud systems}.
\newblock \bibinfo{journal}{\emph{IEEE Transactions on Wireless
  Communications}} (\bibinfo{year}{2022}).
\newblock


\bibitem[Niknam et~al\mbox{.}(2020)]%
        {commag2020flwireless}
\bibfield{author}{\bibinfo{person}{Solmaz Niknam}, \bibinfo{person}{Harpreet~S
  Dhillon}, {and} \bibinfo{person}{Jeffrey~H Reed}.}
  \bibinfo{year}{2020}\natexlab{}.
\newblock \showarticletitle{Federated learning for wireless communications:
  Motivation, opportunities, and challenges}.
\newblock \bibinfo{journal}{\emph{IEEE Communications Magazine}}
  (\bibinfo{year}{2020}).
\newblock


\bibitem[Nishio and Yonetani(2019)]%
        {icc2019fedcs}
\bibfield{author}{\bibinfo{person}{Takayuki Nishio} {and} \bibinfo{person}{Ryo
  Yonetani}.} \bibinfo{year}{2019}\natexlab{}.
\newblock \showarticletitle{Client selection for federated learning with
  heterogeneous resources in mobile edge}. In \bibinfo{booktitle}{\emph{ICC}}.
\newblock


\bibitem[Ozdayi et~al\mbox{.}(2021)]%
        {aaai2021defendbackdoor}
\bibfield{author}{\bibinfo{person}{Mustafa~Safa Ozdayi}, \bibinfo{person}{Murat
  Kantarcioglu}, {and} \bibinfo{person}{Yulia~R Gel}.}
  \bibinfo{year}{2021}\natexlab{}.
\newblock \showarticletitle{Defending against backdoors in federated learning
  with robust learning rate}. In \bibinfo{booktitle}{\emph{AAAI}}.
\newblock


\bibitem[Pappas et~al\mbox{.}(2021)]%
        {ifip2021ipls}
\bibfield{author}{\bibinfo{person}{Christodoulos Pappas},
  \bibinfo{person}{Dimitris Chatzopoulos}, \bibinfo{person}{Spyros Lalis},
  {and} \bibinfo{person}{Manolis Vavalis}.} \bibinfo{year}{2021}\natexlab{}.
\newblock \showarticletitle{Ipls: A framework for decentralized federated
  learning}. In \bibinfo{booktitle}{\emph{IFIP Networking}}.
\newblock


\bibitem[Perera et~al\mbox{.}(2017)]%
        {csur2017fogcompute}
\bibfield{author}{\bibinfo{person}{Charith Perera}, \bibinfo{person}{Yongrui
  Qin}, \bibinfo{person}{Julio~C Estrella}, \bibinfo{person}{Stephan
  Reiff-Marganiec}, {and} \bibinfo{person}{Athanasios~V Vasilakos}.}
  \bibinfo{year}{2017}\natexlab{}.
\newblock \showarticletitle{Fog computing for sustainable smart cities: A
  survey}.
\newblock \bibinfo{journal}{\emph{ACM Computing Surveys (CSUR)}}
  (\bibinfo{year}{2017}).
\newblock


\bibitem[Pillutla et~al\mbox{.}(2023)]%
        {springerml2023superquantile}
\bibfield{author}{\bibinfo{person}{Krishna Pillutla}, \bibinfo{person}{Yassine
  Laguel}, \bibinfo{person}{J{\'e}r{\^o}me Malick}, {and} \bibinfo{person}{Zaid
  Harchaoui}.} \bibinfo{year}{2023}\natexlab{}.
\newblock \showarticletitle{Federated learning with superquantile aggregation
  for heterogeneous data}.
\newblock \bibinfo{journal}{\emph{Springer Machine Learning}}
  (\bibinfo{year}{2023}).
\newblock


\bibitem[Pillutla et~al\mbox{.}(2022)]%
        {icml2022fedalt}
\bibfield{author}{\bibinfo{person}{Krishna Pillutla}, \bibinfo{person}{Kshitiz
  Malik}, \bibinfo{person}{Abdel-Rahman Mohamed}, \bibinfo{person}{Mike
  Rabbat}, \bibinfo{person}{Maziar Sanjabi}, {and} \bibinfo{person}{Lin Xiao}.}
  \bibinfo{year}{2022}\natexlab{}.
\newblock \showarticletitle{Federated Learning with Partial Model
  Personalization}. In \bibinfo{booktitle}{\emph{ICML}}.
\newblock


\bibitem[Qin et~al\mbox{.}(2023)]%
        {cvpr2023ripfl}
\bibfield{author}{\bibinfo{person}{Zixuan Qin}, \bibinfo{person}{Liu Yang},
  \bibinfo{person}{Qilong Wang}, \bibinfo{person}{Yahong Han}, {and}
  \bibinfo{person}{Qinghua Hu}.} \bibinfo{year}{2023}\natexlab{}.
\newblock \showarticletitle{Reliable and Interpretable Personalized Federated
  Learning}. In \bibinfo{booktitle}{\emph{CVPR}}.
\newblock


\bibitem[Qu et~al\mbox{.}(2022b)]%
        {cvpr2022vitfl}
\bibfield{author}{\bibinfo{person}{Liangqiong Qu}, \bibinfo{person}{Yuyin
  Zhou}, \bibinfo{person}{Paul~Pu Liang}, \bibinfo{person}{Yingda Xia},
  \bibinfo{person}{Feifei Wang}, \bibinfo{person}{Ehsan Adeli},
  \bibinfo{person}{Li Fei-Fei}, {and} \bibinfo{person}{Daniel Rubin}.}
  \bibinfo{year}{2022}\natexlab{b}.
\newblock \showarticletitle{Rethinking architecture design for tackling data
  heterogeneity in federated learning}. In \bibinfo{booktitle}{\emph{CVPR}}.
\newblock


\bibitem[Qu et~al\mbox{.}(2020)]%
        {ieeeiot2020flfogcompute}
\bibfield{author}{\bibinfo{person}{Youyang Qu}, \bibinfo{person}{Longxiang
  Gao}, \bibinfo{person}{Tom~H Luan}, \bibinfo{person}{Yong Xiang},
  \bibinfo{person}{Shui Yu}, \bibinfo{person}{Bai Li}, {and}
  \bibinfo{person}{Gavin Zheng}.} \bibinfo{year}{2020}\natexlab{}.
\newblock \showarticletitle{Decentralized privacy using blockchain-enabled
  federated learning in fog computing}.
\newblock \bibinfo{journal}{\emph{IEEE Internet of Things Journal}}
  (\bibinfo{year}{2020}).
\newblock


\bibitem[Qu et~al\mbox{.}(2022a)]%
        {icml2022fedsam}
\bibfield{author}{\bibinfo{person}{Zhe Qu}, \bibinfo{person}{Xingyu Li},
  \bibinfo{person}{Rui Duan}, \bibinfo{person}{Yao Liu}, \bibinfo{person}{Bo
  Tang}, {and} \bibinfo{person}{Zhuo Lu}.} \bibinfo{year}{2022}\natexlab{a}.
\newblock \showarticletitle{Generalized Federated Learning via Sharpness Aware
  Minimization}. In \bibinfo{booktitle}{\emph{ICML}}.
\newblock


\bibitem[Roy et~al\mbox{.}(2019)]%
        {roy2019braintorrent}
\bibfield{author}{\bibinfo{person}{Abhijit~Guha Roy}, \bibinfo{person}{Shayan
  Siddiqui}, \bibinfo{person}{Sebastian P{\"o}lsterl}, \bibinfo{person}{Nassir
  Navab}, {and} \bibinfo{person}{Christian Wachinger}.}
  \bibinfo{year}{2019}\natexlab{}.
\newblock \showarticletitle{Braintorrent: A peer-to-peer environment for
  decentralized federated learning}.
\newblock \bibinfo{journal}{\emph{arXiv preprint arXiv:1905.06731}}
  (\bibinfo{year}{2019}).
\newblock


\bibitem[Ruan and Joe-Wong(2022)]%
        {aaai2021fedsoft}
\bibfield{author}{\bibinfo{person}{Yichen Ruan} {and} \bibinfo{person}{Carlee
  Joe-Wong}.} \bibinfo{year}{2022}\natexlab{}.
\newblock \showarticletitle{FedSoft: Soft Clustered Federated Learning with
  Proximal Local Updating}. In \bibinfo{booktitle}{\emph{AAAI}}.
\newblock


\bibitem[Sadilek et~al\mbox{.}(2021)]%
        {npj2021privacy}
\bibfield{author}{\bibinfo{person}{Adam Sadilek}, \bibinfo{person}{Luyang Liu},
  \bibinfo{person}{Dung Nguyen}, \bibinfo{person}{Methun Kamruzzaman},
  \bibinfo{person}{Stylianos Serghiou}, \bibinfo{person}{Benjamin Rader},
  \bibinfo{person}{Alex Ingerman}, \bibinfo{person}{Stefan Mellem},
  \bibinfo{person}{Peter Kairouz}, \bibinfo{person}{Elaine~O Nsoesie},
  {et~al\mbox{.}}} \bibinfo{year}{2021}\natexlab{}.
\newblock \showarticletitle{Privacy-first health research with federated
  learning}.
\newblock \bibinfo{journal}{\emph{NPJ digital medicine}}
  (\bibinfo{year}{2021}).
\newblock


\bibitem[Sattler et~al\mbox{.}(2020a)]%
        {sattler2020cfd}
\bibfield{author}{\bibinfo{person}{Felix Sattler}, \bibinfo{person}{Arturo
  Marban}, \bibinfo{person}{Roman Rischke}, {and} \bibinfo{person}{Wojciech
  Samek}.} \bibinfo{year}{2020}\natexlab{a}.
\newblock \showarticletitle{Communication-efficient federated distillation}.
\newblock \bibinfo{journal}{\emph{arXiv preprint arXiv:2012.00632}}
  (\bibinfo{year}{2020}).
\newblock


\bibitem[Sattler et~al\mbox{.}(2020b)]%
        {ieee2020cfl}
\bibfield{author}{\bibinfo{person}{Felix Sattler},
  \bibinfo{person}{Klaus-Robert M{\"u}ller}, {and} \bibinfo{person}{Wojciech
  Samek}.} \bibinfo{year}{2020}\natexlab{b}.
\newblock \showarticletitle{Clustered federated learning: Model-agnostic
  distributed multitask optimization under privacy constraints}. In
  \bibinfo{booktitle}{\emph{IEEE TNNLS}}.
\newblock


\bibitem[Sattler et~al\mbox{.}(2020c)]%
        {icassp2020robustcfl}
\bibfield{author}{\bibinfo{person}{Felix Sattler},
  \bibinfo{person}{Klaus-Robert M{\"u}ller}, \bibinfo{person}{Thomas Wiegand},
  {and} \bibinfo{person}{Wojciech Samek}.} \bibinfo{year}{2020}\natexlab{c}.
\newblock \showarticletitle{On the byzantine robustness of clustered federated
  learning}. In \bibinfo{booktitle}{\emph{IEEE ICASSP}}.
\newblock


\bibitem[Sattler et~al\mbox{.}(2019)]%
        {tnnls2019robustcom}
\bibfield{author}{\bibinfo{person}{Felix Sattler}, \bibinfo{person}{Simon
  Wiedemann}, \bibinfo{person}{Klaus-Robert M{\"u}ller}, {and}
  \bibinfo{person}{Wojciech Samek}.} \bibinfo{year}{2019}\natexlab{}.
\newblock \showarticletitle{Robust and communication-efficient federated
  learning from non-iid data}.
\newblock \bibinfo{journal}{\emph{IEEE TNNLS}} (\bibinfo{year}{2019}).
\newblock


\bibitem[Shahid et~al\mbox{.}(2021)]%
        {shahid2021efficiencychallenge}
\bibfield{author}{\bibinfo{person}{Osama Shahid}, \bibinfo{person}{Seyedamin
  Pouriyeh}, \bibinfo{person}{Reza~M Parizi}, \bibinfo{person}{Quan~Z Sheng},
  \bibinfo{person}{Gautam Srivastava}, {and} \bibinfo{person}{Liang Zhao}.}
  \bibinfo{year}{2021}\natexlab{}.
\newblock \showarticletitle{Communication Efficiency in Federated Learning:
  Achievements and Challenges}.
\newblock \bibinfo{journal}{\emph{arXiv preprint arXiv:2107.10996}}
  (\bibinfo{year}{2021}).
\newblock


\bibitem[Shang et~al\mbox{.}(2022)]%
        {ijcai2022creff}
\bibfield{author}{\bibinfo{person}{Xinyi Shang}, \bibinfo{person}{Yang Lu},
  \bibinfo{person}{Gang Huang}, {and} \bibinfo{person}{Hanzi Wang}.}
  \bibinfo{year}{2022}\natexlab{}.
\newblock \showarticletitle{Federated Learning on Heterogeneous and Long-Tailed
  Data via Classifier Re-Training with Federated Features}. In
  \bibinfo{booktitle}{\emph{IJCAI}}.
\newblock


\bibitem[Shao et~al\mbox{.}(2022)]%
        {tnnls2022fedpad}
\bibfield{author}{\bibinfo{person}{Rui Shao}, \bibinfo{person}{Pramuditha
  Perera}, \bibinfo{person}{Pong~C Yuen}, {and} \bibinfo{person}{Vishal~M
  Patel}.} \bibinfo{year}{2022}\natexlab{}.
\newblock \showarticletitle{Federated Generalized Face Presentation Attack
  Detection}. In \bibinfo{booktitle}{\emph{IEEE TNNLS}}.
\newblock


\bibitem[Shen et~al\mbox{.}(2022)]%
        {cvpr2022cd2pfed}
\bibfield{author}{\bibinfo{person}{Yiqing Shen}, \bibinfo{person}{Yuyin Zhou},
  {and} \bibinfo{person}{Lequan Yu}.} \bibinfo{year}{2022}\natexlab{}.
\newblock \showarticletitle{CD2-pFed: Cyclic Distillation-guided Channel
  Decoupling for Model Personalization in Federated Learning}. In
  \bibinfo{booktitle}{\emph{CVPR}}.
\newblock


\bibitem[Shi et~al\mbox{.}(2023a)]%
        {cvpr2023dpfedsam}
\bibfield{author}{\bibinfo{person}{Yifan Shi}, \bibinfo{person}{Yingqi Liu},
  \bibinfo{person}{Kang Wei}, \bibinfo{person}{Li Shen},
  \bibinfo{person}{Xueqian Wang}, {and} \bibinfo{person}{Dacheng Tao}.}
  \bibinfo{year}{2023}\natexlab{a}.
\newblock \showarticletitle{Make landscape flatter in differentially private
  federated learning}. In \bibinfo{booktitle}{\emph{CVPR}}.
\newblock


\bibitem[Shi et~al\mbox{.}(2023b)]%
        {icml2023dfedsam}
\bibfield{author}{\bibinfo{person}{Yifan Shi}, \bibinfo{person}{Li Shen},
  \bibinfo{person}{Kang Wei}, \bibinfo{person}{Yan Sun}, \bibinfo{person}{Bo
  Yuan}, \bibinfo{person}{Xueqian Wang}, {and} \bibinfo{person}{Dacheng Tao}.}
  \bibinfo{year}{2023}\natexlab{b}.
\newblock \showarticletitle{Improving the Model Consistency of Decentralized
  Federated Learning}. In \bibinfo{booktitle}{\emph{ICML}}.
\newblock


\bibitem[Shin et~al\mbox{.}(2020)]%
        {shin2020xormixup}
\bibfield{author}{\bibinfo{person}{MyungJae Shin}, \bibinfo{person}{Chihoon
  Hwang}, \bibinfo{person}{Joongheon Kim}, \bibinfo{person}{Jihong Park},
  \bibinfo{person}{Mehdi Bennis}, {and} \bibinfo{person}{Seong-Lyun Kim}.}
  \bibinfo{year}{2020}\natexlab{}.
\newblock \showarticletitle{Xor mixup: Privacy-preserving data augmentation for
  one-shot federated learning}.
\newblock \bibinfo{journal}{\emph{arXiv preprint arXiv:2006.05148}}
  (\bibinfo{year}{2020}).
\newblock


\bibitem[Shoham et~al\mbox{.}(2019)]%
        {nips2019fedcurv}
\bibfield{author}{\bibinfo{person}{Neta Shoham}, \bibinfo{person}{Tomer
  Avidor}, \bibinfo{person}{Aviv Keren}, \bibinfo{person}{Nadav Israel},
  \bibinfo{person}{Daniel Benditkis}, \bibinfo{person}{Liron Mor-Yosef}, {and}
  \bibinfo{person}{Itai Zeitak}.} \bibinfo{year}{2019}\natexlab{}.
\newblock \showarticletitle{Overcoming forgetting in federated learning on
  non-iid data}. In \bibinfo{booktitle}{\emph{NeurIPS Workshop}}.
\newblock


\bibitem[Smith et~al\mbox{.}(2017)]%
        {nips2017mocha}
\bibfield{author}{\bibinfo{person}{Virginia Smith}, \bibinfo{person}{Chao-Kai
  Chiang}, \bibinfo{person}{Maziar Sanjabi}, {and} \bibinfo{person}{Ameet~S
  Talwalkar}.} \bibinfo{year}{2017}\natexlab{}.
\newblock \showarticletitle{Federated multi-task learning}. In
  \bibinfo{booktitle}{\emph{NeurIPS}}.
\newblock


\bibitem[Srivastava et~al\mbox{.}(2014)]%
        {jmlr2014dropout}
\bibfield{author}{\bibinfo{person}{Nitish Srivastava},
  \bibinfo{person}{Geoffrey Hinton}, \bibinfo{person}{Alex Krizhevsky},
  \bibinfo{person}{Ilya Sutskever}, {and} \bibinfo{person}{Ruslan
  Salakhutdinov}.} \bibinfo{year}{2014}\natexlab{}.
\newblock \showarticletitle{Dropout: a simple way to prevent neural networks
  from overfitting}.
\newblock \bibinfo{journal}{\emph{JMLR}} (\bibinfo{year}{2014}).
\newblock


\bibitem[Sun et~al\mbox{.}(2021a)]%
        {nips2021flwbc}
\bibfield{author}{\bibinfo{person}{Jingwei Sun}, \bibinfo{person}{Ang Li},
  \bibinfo{person}{Louis DiValentin}, \bibinfo{person}{Amin Hassanzadeh},
  \bibinfo{person}{Yiran Chen}, {and} \bibinfo{person}{Hai Li}.}
  \bibinfo{year}{2021}\natexlab{a}.
\newblock \showarticletitle{Fl-wbc: Enhancing robustness against model
  poisoning attacks in federated learning from a client perspective}. In
  \bibinfo{booktitle}{\emph{NeurIPS}}.
\newblock


\bibitem[Sun et~al\mbox{.}(2021b)]%
        {cvpr2021soteria}
\bibfield{author}{\bibinfo{person}{Jingwei Sun}, \bibinfo{person}{Ang Li},
  \bibinfo{person}{Binghui Wang}, \bibinfo{person}{Huanrui Yang},
  \bibinfo{person}{Hai Li}, {and} \bibinfo{person}{Yiran Chen}.}
  \bibinfo{year}{2021}\natexlab{b}.
\newblock \showarticletitle{Soteria: Provable defense against privacy leakage
  in federated learning from representation perspective}. In
  \bibinfo{booktitle}{\emph{CVPR}}.
\newblock


\bibitem[Sun and Lyu(2021)]%
        {ijcai2021fedmdnfdp}
\bibfield{author}{\bibinfo{person}{Lichao Sun} {and} \bibinfo{person}{Lingjuan
  Lyu}.} \bibinfo{year}{2021}\natexlab{}.
\newblock \showarticletitle{Federated model distillation with noise-free
  differential privacy}. In \bibinfo{booktitle}{\emph{IJCAI}}.
\newblock


\bibitem[Sun et~al\mbox{.}(2019)]%
        {nips2019canyoubackdoor}
\bibfield{author}{\bibinfo{person}{Ziteng Sun}, \bibinfo{person}{Peter
  Kairouz}, \bibinfo{person}{Ananda~Theertha Suresh}, {and}
  \bibinfo{person}{H~Brendan McMahan}.} \bibinfo{year}{2019}\natexlab{}.
\newblock \showarticletitle{Can you really backdoor federated learning?}. In
  \bibinfo{booktitle}{\emph{NeurIPS Workshop}}.
\newblock


\bibitem[T~Dinh et~al\mbox{.}(2020)]%
        {nips2020pfedme}
\bibfield{author}{\bibinfo{person}{Canh T~Dinh}, \bibinfo{person}{Nguyen Tran},
  {and} \bibinfo{person}{Josh Nguyen}.} \bibinfo{year}{2020}\natexlab{}.
\newblock \showarticletitle{Personalized federated learning with moreau
  envelopes}. In \bibinfo{booktitle}{\emph{NeurIPS}}.
\newblock


\bibitem[Tan et~al\mbox{.}(2022b)]%
        {tan2022toward}
\bibfield{author}{\bibinfo{person}{Alysa~Ziying Tan}, \bibinfo{person}{Han Yu},
  \bibinfo{person}{Lizhen Cui}, {and} \bibinfo{person}{Qiang Yang}.}
  \bibinfo{year}{2022}\natexlab{b}.
\newblock \showarticletitle{Towards Personalized Federated Learning}.
\newblock \bibinfo{journal}{\emph{IEEE TNNLS}} (\bibinfo{year}{2022}).
\newblock


\bibitem[Tan et~al\mbox{.}(2022a)]%
        {aaai2022fedproto}
\bibfield{author}{\bibinfo{person}{Yue Tan}, \bibinfo{person}{Guodong Long},
  \bibinfo{person}{Lu Liu}, \bibinfo{person}{Tianyi Zhou},
  \bibinfo{person}{Qinghua Lu}, \bibinfo{person}{Jing Jiang}, {and}
  \bibinfo{person}{Chengqi Zhang}.} \bibinfo{year}{2022}\natexlab{a}.
\newblock \showarticletitle{Fedproto: Federated prototype learning across
  heterogeneous clients}. In \bibinfo{booktitle}{\emph{AAAI}}.
\newblock


\bibitem[Tang et~al\mbox{.}(2022a)]%
        {cvpr2022fedcor}
\bibfield{author}{\bibinfo{person}{Minxue Tang}, \bibinfo{person}{Xuefei Ning},
  \bibinfo{person}{Yitu Wang}, \bibinfo{person}{Jingwei Sun},
  \bibinfo{person}{Yu Wang}, \bibinfo{person}{Hai Li}, {and}
  \bibinfo{person}{Yiran Chen}.} \bibinfo{year}{2022}\natexlab{a}.
\newblock \showarticletitle{FedCor: Correlation-Based Active Client Selection
  Strategy for Heterogeneous Federated Learning}. In
  \bibinfo{booktitle}{\emph{CVPR}}.
\newblock


\bibitem[Tang et~al\mbox{.}(2022b)]%
        {tpds2022gossipfl}
\bibfield{author}{\bibinfo{person}{Zhenheng Tang}, \bibinfo{person}{Shaohuai
  Shi}, \bibinfo{person}{Bo Li}, {and} \bibinfo{person}{Xiaowen Chu}.}
  \bibinfo{year}{2022}\natexlab{b}.
\newblock \showarticletitle{GossipFL: A decentralized federated learning
  framework with sparsified and adaptive communication}.
\newblock \bibinfo{journal}{\emph{IEEE TPDS}} (\bibinfo{year}{2022}).
\newblock


\bibitem[Tang et~al\mbox{.}(2022c)]%
        {icml2022vhl}
\bibfield{author}{\bibinfo{person}{Zhenheng Tang}, \bibinfo{person}{Yonggang
  Zhang}, \bibinfo{person}{Shaohuai Shi}, \bibinfo{person}{Xin He},
  \bibinfo{person}{Bo Han}, {and} \bibinfo{person}{Xiaowen Chu}.}
  \bibinfo{year}{2022}\natexlab{c}.
\newblock \showarticletitle{Virtual Homogeneity Learning: Defending against
  Data Heterogeneity in Federated Learning}. In
  \bibinfo{booktitle}{\emph{ICML}}.
\newblock


\bibitem[Tolpegin et~al\mbox{.}(2020)]%
        {esorics2020datapoisonattack}
\bibfield{author}{\bibinfo{person}{Vale Tolpegin}, \bibinfo{person}{Stacey
  Truex}, \bibinfo{person}{Mehmet~Emre Gursoy}, {and} \bibinfo{person}{Ling
  Liu}.} \bibinfo{year}{2020}\natexlab{}.
\newblock \showarticletitle{Data poisoning attacks against federated learning
  systems}. In \bibinfo{booktitle}{\emph{ESORICS}}.
\newblock


\bibitem[Truex et~al\mbox{.}(2020)]%
        {acm2020ldpfed}
\bibfield{author}{\bibinfo{person}{Stacey Truex}, \bibinfo{person}{Ling Liu},
  \bibinfo{person}{Ka-Ho Chow}, \bibinfo{person}{Mehmet~Emre Gursoy}, {and}
  \bibinfo{person}{Wenqi Wei}.} \bibinfo{year}{2020}\natexlab{}.
\newblock \showarticletitle{LDP-Fed: Federated learning with local differential
  privacy}. In \bibinfo{booktitle}{\emph{Proceedings of the Third ACM
  International Workshop on Edge Systems, Analytics and Networking}}.
\newblock


\bibitem[Usynin et~al\mbox{.}(2021)]%
        {nature2021adversarial}
\bibfield{author}{\bibinfo{person}{Dmitrii Usynin}, \bibinfo{person}{Alexander
  Ziller}, \bibinfo{person}{Marcus Makowski}, \bibinfo{person}{Rickmer Braren},
  \bibinfo{person}{Daniel Rueckert}, \bibinfo{person}{Ben Glocker},
  \bibinfo{person}{Georgios Kaissis}, {and} \bibinfo{person}{Jonathan
  Passerat-Palmbach}.} \bibinfo{year}{2021}\natexlab{}.
\newblock \showarticletitle{Adversarial interference and its mitigations in
  privacy-preserving collaborative machine learning}.
\newblock \bibinfo{journal}{\emph{Nature Machine Intelligence}}
  (\bibinfo{year}{2021}).
\newblock


\bibitem[van Berlo et~al\mbox{.}(2020)]%
        {acm2020furl}
\bibfield{author}{\bibinfo{person}{Bram van Berlo}, \bibinfo{person}{Aaqib
  Saeed}, {and} \bibinfo{person}{Tanir Ozcelebi}.}
  \bibinfo{year}{2020}\natexlab{}.
\newblock \showarticletitle{Towards federated unsupervised representation
  learning}. In \bibinfo{booktitle}{\emph{ACM EdgeSys}}.
\newblock


\bibitem[Vargaftik et~al\mbox{.}(2022)]%
        {icml2022eden}
\bibfield{author}{\bibinfo{person}{Shay Vargaftik}, \bibinfo{person}{Ran~Ben
  Basat}, \bibinfo{person}{Amit Portnoy}, \bibinfo{person}{Gal Mendelson},
  \bibinfo{person}{Yaniv~Ben Itzhak}, {and} \bibinfo{person}{Michael
  Mitzenmacher}.} \bibinfo{year}{2022}\natexlab{}.
\newblock \showarticletitle{Eden: Communication-efficient and robust
  distributed mean estimation for federated learning}. In
  \bibinfo{booktitle}{\emph{ICML}}.
\newblock


\bibitem[Wahab et~al\mbox{.}(2021)]%
        {wahab2021fml}
\bibfield{author}{\bibinfo{person}{Omar~Abdel Wahab}, \bibinfo{person}{Azzam
  Mourad}, \bibinfo{person}{Hadi Otrok}, {and} \bibinfo{person}{Tarik Taleb}.}
  \bibinfo{year}{2021}\natexlab{}.
\newblock \showarticletitle{Federated machine learning: Survey, multi-level
  classification, desirable criteria and future directions in communication and
  networking systems}.
\newblock \bibinfo{journal}{\emph{IEEE Communications Surveys \& Tutorials}}
  (\bibinfo{year}{2021}).
\newblock


\bibitem[Wang et~al\mbox{.}(2020a)]%
        {ieee2020favor}
\bibfield{author}{\bibinfo{person}{Hao Wang}, \bibinfo{person}{Zakhary Kaplan},
  \bibinfo{person}{Di Niu}, {and} \bibinfo{person}{Baochun Li}.}
  \bibinfo{year}{2020}\natexlab{a}.
\newblock \showarticletitle{Optimizing federated learning on non-iid data with
  reinforcement learning}. In \bibinfo{booktitle}{\emph{IEEE INFOCOM}}.
\newblock


\bibitem[Wang et~al\mbox{.}(2020b)]%
        {nips2020backdoor}
\bibfield{author}{\bibinfo{person}{Hongyi Wang}, \bibinfo{person}{Kartik
  Sreenivasan}, \bibinfo{person}{Shashank Rajput}, {and}
  \bibinfo{person}{et~al. Vishwakarma}.} \bibinfo{year}{2020}\natexlab{b}.
\newblock \showarticletitle{Attack of the tails: Yes, you really can backdoor
  federated learning}. In \bibinfo{booktitle}{\emph{NeurIPS}}.
\newblock


\bibitem[Wang et~al\mbox{.}(2022b)]%
        {icml2022progfed}
\bibfield{author}{\bibinfo{person}{Hui-Po Wang}, \bibinfo{person}{Sebastian~U
  Stich}, \bibinfo{person}{Yang He}, {and} \bibinfo{person}{Mario Fritz}.}
  \bibinfo{year}{2022}\natexlab{b}.
\newblock \showarticletitle{ProgFed: Effective, Communication, and Computation
  Efficient Federated Learning by Progressive Training}. In
  \bibinfo{booktitle}{\emph{ICML}}.
\newblock


\bibitem[Wang and Kantarci(2020)]%
        {camad2020reputationaware}
\bibfield{author}{\bibinfo{person}{Yuwei Wang} {and} \bibinfo{person}{Burak
  Kantarci}.} \bibinfo{year}{2020}\natexlab{}.
\newblock \showarticletitle{A novel reputation-aware client selection scheme
  for federated learning within mobile environments}. In
  \bibinfo{booktitle}{\emph{IEEE CAMAD}}.
\newblock


\bibitem[Wang et~al\mbox{.}(2022a)]%
        {icml2022fedcams}
\bibfield{author}{\bibinfo{person}{Yujia Wang}, \bibinfo{person}{Lu Lin}, {and}
  \bibinfo{person}{Jinghui Chen}.} \bibinfo{year}{2022}\natexlab{a}.
\newblock \showarticletitle{Communication-Efficient Adaptive Federated
  Learning}. In \bibinfo{booktitle}{\emph{ICML}}.
\newblock


\bibitem[Wu et~al\mbox{.}(2022b)]%
        {nature2022fedkd}
\bibfield{author}{\bibinfo{person}{Chuhan Wu}, \bibinfo{person}{Fangzhao Wu},
  \bibinfo{person}{Lingjuan Lyu}, \bibinfo{person}{Yongfeng Huang}, {and}
  \bibinfo{person}{Xing Xie}.} \bibinfo{year}{2022}\natexlab{b}.
\newblock \showarticletitle{Communication-efficient federated learning via
  knowledge distillation}.
\newblock \bibinfo{journal}{\emph{Nature communications}}
  (\bibinfo{year}{2022}).
\newblock


\bibitem[Wu et~al\mbox{.}(2022a)]%
        {ieee2022blockchainmultifl}
\bibfield{author}{\bibinfo{person}{Di Wu}, \bibinfo{person}{Nai Wang},
  \bibinfo{person}{Jiale Zhang}, \bibinfo{person}{Yuan Zhang},
  \bibinfo{person}{Yong Xiang}, {and} \bibinfo{person}{Longxiang Gao}.}
  \bibinfo{year}{2022}\natexlab{a}.
\newblock \showarticletitle{A Blockchain-based Multi-layer Decentralized
  Framework for Robust Federated Learning}. In \bibinfo{booktitle}{\emph{2022
  International Joint Conference on Neural Networks (IJCNN)}}.
\newblock


\bibitem[Wu et~al\mbox{.}(2020a)]%
        {wu2020ojcs}
\bibfield{author}{\bibinfo{person}{Qiong Wu}, \bibinfo{person}{Kaiwen He},
  {and} \bibinfo{person}{Xu Chen}.} \bibinfo{year}{2020}\natexlab{a}.
\newblock \showarticletitle{Personalized federated learning for intelligent IoT
  applications: A cloud-edge based framework}.
\newblock \bibinfo{journal}{\emph{IEEE OJ-CS}} (\bibinfo{year}{2020}).
\newblock


\bibitem[Wu et~al\mbox{.}(2020b)]%
        {tpds2020hybridfl}
\bibfield{author}{\bibinfo{person}{Wentai Wu}, \bibinfo{person}{Ligang He},
  \bibinfo{person}{Weiwei Lin}, {and} \bibinfo{person}{Rui Mao}.}
  \bibinfo{year}{2020}\natexlab{b}.
\newblock \showarticletitle{Accelerating federated learning over
  reliability-agnostic clients in mobile edge computing systems}.
\newblock \bibinfo{journal}{\emph{IEEE TPDS}} (\bibinfo{year}{2020}).
\newblock


\bibitem[Xie et~al\mbox{.}(2021)]%
        {icml2021crfl}
\bibfield{author}{\bibinfo{person}{Chulin Xie}, \bibinfo{person}{Minghao Chen},
  \bibinfo{person}{Pin-Yu Chen}, {and} \bibinfo{person}{Bo Li}.}
  \bibinfo{year}{2021}\natexlab{}.
\newblock \showarticletitle{Crfl: Certifiably robust federated learning against
  backdoor attacks}. In \bibinfo{booktitle}{\emph{ICML}}.
\newblock


\bibitem[Xie et~al\mbox{.}(2019)]%
        {iclr2019dba}
\bibfield{author}{\bibinfo{person}{Chulin Xie}, \bibinfo{person}{Keli Huang},
  \bibinfo{person}{Pin-Yu Chen}, {and} \bibinfo{person}{Bo Li}.}
  \bibinfo{year}{2019}\natexlab{}.
\newblock \showarticletitle{Dba: Distributed backdoor attacks against federated
  learning}. In \bibinfo{booktitle}{\emph{ICLR}}.
\newblock


\bibitem[Xie et~al\mbox{.}(2020)]%
        {xie2020fesem}
\bibfield{author}{\bibinfo{person}{Ming Xie}, \bibinfo{person}{Guodong Long},
  \bibinfo{person}{Tao Shen}, \bibinfo{person}{Tianyi Zhou},
  \bibinfo{person}{Xianzhi Wang}, \bibinfo{person}{Jing Jiang}, {and}
  \bibinfo{person}{Chengqi Zhang}.} \bibinfo{year}{2020}\natexlab{}.
\newblock \showarticletitle{Multi-center federated learning}.
\newblock \bibinfo{journal}{\emph{arXiv preprint arXiv:2005.01026}}
  (\bibinfo{year}{2020}).
\newblock


\bibitem[Xiong et~al\mbox{.}(2023)]%
        {cvpr2023feddm}
\bibfield{author}{\bibinfo{person}{Yuanhao Xiong}, \bibinfo{person}{Ruochen
  Wang}, \bibinfo{person}{Minhao Cheng}, \bibinfo{person}{Felix Yu}, {and}
  \bibinfo{person}{Cho-Jui Hsieh}.} \bibinfo{year}{2023}\natexlab{}.
\newblock \showarticletitle{Feddm: Iterative distribution matching for
  communication-efficient federated learning}. In
  \bibinfo{booktitle}{\emph{CVPR}}.
\newblock


\bibitem[Xu et~al\mbox{.}(2022a)]%
        {cvpr2022fedcorr}
\bibfield{author}{\bibinfo{person}{Jingyi Xu}, \bibinfo{person}{Zihan Chen},
  \bibinfo{person}{Tony~QS Quek}, {and} \bibinfo{person}{Kai Fong~Ernest
  Chong}.} \bibinfo{year}{2022}\natexlab{a}.
\newblock \showarticletitle{FedCorr: Multi-Stage Federated Learning for Label
  Noise Correction}. In \bibinfo{booktitle}{\emph{CVPR}}.
\newblock


\bibitem[Xu et~al\mbox{.}(2023)]%
        {iclr2023fedpac}
\bibfield{author}{\bibinfo{person}{Jian Xu}, \bibinfo{person}{Xinyi Tong},
  {and} \bibinfo{person}{Shao-Lun Huang}.} \bibinfo{year}{2023}\natexlab{}.
\newblock \showarticletitle{Personalized Federated Learning with Feature
  Alignment and Classifier Collaboration}. In \bibinfo{booktitle}{\emph{ICLR}}.
\newblock


\bibitem[Xu et~al\mbox{.}(2022b)]%
        {ieeetii2022safe}
\bibfield{author}{\bibinfo{person}{Xiaolong Xu}, \bibinfo{person}{Haoyuan Li},
  \bibinfo{person}{Zheng Li}, {and} \bibinfo{person}{Xiaokang Zhou}.}
  \bibinfo{year}{2022}\natexlab{b}.
\newblock \showarticletitle{Safe: Synergic data filtering for federated
  learning in cloud-edge computing}.
\newblock \bibinfo{journal}{\emph{IEEE Transactions on Industrial Informatics}}
  (\bibinfo{year}{2022}).
\newblock


\bibitem[Yang et~al\mbox{.}(2021)]%
        {corr2020cucb}
\bibfield{author}{\bibinfo{person}{Miao Yang}, \bibinfo{person}{Ximin Wang},
  \bibinfo{person}{Hongbin Zhu}, \bibinfo{person}{Haifeng Wang}, {and}
  \bibinfo{person}{Hua Qian}.} \bibinfo{year}{2021}\natexlab{}.
\newblock \showarticletitle{Federated learning with class imbalance reduction}.
  In \bibinfo{booktitle}{\emph{EUSIPCO}}.
\newblock


\bibitem[Yang et~al\mbox{.}(2019)]%
        {tist2019fmlconcept}
\bibfield{author}{\bibinfo{person}{Qiang Yang}, \bibinfo{person}{Yang Liu},
  \bibinfo{person}{Tianjian Chen}, {and} \bibinfo{person}{Yongxin Tong}.}
  \bibinfo{year}{2019}\natexlab{}.
\newblock \showarticletitle{Federated machine learning: Concept and
  applications}. In \bibinfo{booktitle}{\emph{ACM TIST}}.
\newblock


\bibitem[Yang et~al\mbox{.}(2020)]%
        {yang2020noiselabel}
\bibfield{author}{\bibinfo{person}{Seunghan Yang}, \bibinfo{person}{Hyoungseob
  Park}, \bibinfo{person}{Junyoung Byun}, {and} \bibinfo{person}{et al.}}
  \bibinfo{year}{2020}\natexlab{}.
\newblock \showarticletitle{Robust Federated Learning with Noisy Labels}.
\newblock \bibinfo{journal}{\emph{arXiv preprint arXiv:2012.01700}}
  (\bibinfo{year}{2020}).
\newblock


\bibitem[Yao et~al\mbox{.}(2019)]%
        {yao2019fedmmd}
\bibfield{author}{\bibinfo{person}{Xin Yao}, \bibinfo{person}{Tianchi Huang},
  \bibinfo{person}{Chenglei Wu}, \bibinfo{person}{Rui-Xiao Zhang}, {and}
  \bibinfo{person}{Lifeng Sun}.} \bibinfo{year}{2019}\natexlab{}.
\newblock \showarticletitle{Federated learning with additional mechanisms on
  clients to reduce communication costs}.
\newblock \bibinfo{journal}{\emph{arXiv preprint arXiv:1908.05891}}
  (\bibinfo{year}{2019}).
\newblock


\bibitem[Yao and Sun(2020)]%
        {icip2020fedcl}
\bibfield{author}{\bibinfo{person}{Xin Yao} {and} \bibinfo{person}{Lifeng
  Sun}.} \bibinfo{year}{2020}\natexlab{}.
\newblock \showarticletitle{Continual local training for better initialization
  of federated models}. In \bibinfo{booktitle}{\emph{IEEE ICIP}}.
\newblock


\bibitem[Ye et~al\mbox{.}(2020)]%
        {ye2020augmentation}
\bibfield{author}{\bibinfo{person}{Mang Ye}, \bibinfo{person}{Jianbing Shen},
  \bibinfo{person}{Xu Zhang}, \bibinfo{person}{Pong~C Yuen}, {and}
  \bibinfo{person}{Shih-Fu Chang}.} \bibinfo{year}{2020}\natexlab{}.
\newblock \showarticletitle{Augmentation invariant and instance spreading
  feature for softmax embedding}.
\newblock \bibinfo{journal}{\emph{IEEE TPAMI}} (\bibinfo{year}{2020}).
\newblock


\bibitem[Yi et~al\mbox{.}(2022)]%
        {icml2022qsfl}
\bibfield{author}{\bibinfo{person}{Liping Yi}, \bibinfo{person}{Wang Gang},
  {and} \bibinfo{person}{Liu Xiaoguang}.} \bibinfo{year}{2022}\natexlab{}.
\newblock \showarticletitle{QSFL: A Two-Level Uplink Communication Optimization
  Framework for Federated Learning}. In \bibinfo{booktitle}{\emph{ICML}}.
\newblock


\bibitem[Yin et~al\mbox{.}(2021)]%
        {csur2021privacyflsurvey}
\bibfield{author}{\bibinfo{person}{Xuefei Yin}, \bibinfo{person}{Yanming Zhu},
  {and} \bibinfo{person}{Jiankun Hu}.} \bibinfo{year}{2021}\natexlab{}.
\newblock \showarticletitle{A comprehensive survey of privacy-preserving
  federated learning: A taxonomy, review, and future directions}.
\newblock \bibinfo{journal}{\emph{ACM CSUR}} (\bibinfo{year}{2021}).
\newblock


\bibitem[Yoon et~al\mbox{.}(2022)]%
        {icml2022bhfl}
\bibfield{author}{\bibinfo{person}{Jaehong Yoon}, \bibinfo{person}{Geon Park},
  \bibinfo{person}{Wonyong Jeong}, {and} \bibinfo{person}{Sung~Ju Hwang}.}
  \bibinfo{year}{2022}\natexlab{}.
\newblock \showarticletitle{Bitwidth Heterogeneous Federated Learning with
  Progressive Weight Dequantization}. In \bibinfo{booktitle}{\emph{ICML}}.
\newblock


\bibitem[Yoon et~al\mbox{.}(2021)]%
        {iclr2021fedmix}
\bibfield{author}{\bibinfo{person}{Tehrim Yoon}, \bibinfo{person}{Sumin Shin},
  \bibinfo{person}{Sung~Ju Hwang}, {and} \bibinfo{person}{Eunho Yang}.}
  \bibinfo{year}{2021}\natexlab{}.
\newblock \showarticletitle{Fedmix: Approximation of mixup under mean augmented
  federated learning}. In \bibinfo{booktitle}{\emph{ICLR}}.
\newblock


\bibitem[Yoshida et~al\mbox{.}(2020)]%
        {gc2020mab}
\bibfield{author}{\bibinfo{person}{Naoya Yoshida}, \bibinfo{person}{Takayuki
  Nishio}, \bibinfo{person}{Masahiro Morikura}, {and} \bibinfo{person}{Koji
  Yamamoto}.} \bibinfo{year}{2020}\natexlab{}.
\newblock \showarticletitle{Mab-based client selection for federated learning
  with uncertain resources in mobile networks}. In
  \bibinfo{booktitle}{\emph{IEEE GC Workshops}}.
\newblock


\bibitem[Yu et~al\mbox{.}(2020)]%
        {yu2020salvaging}
\bibfield{author}{\bibinfo{person}{Tao Yu}, \bibinfo{person}{Eugene
  Bagdasaryan}, {and} \bibinfo{person}{Vitaly Shmatikov}.}
  \bibinfo{year}{2020}\natexlab{}.
\newblock \showarticletitle{Salvaging federated learning by local adaptation}.
\newblock \bibinfo{journal}{\emph{arXiv preprint arXiv:2002.04758}}
  (\bibinfo{year}{2020}).
\newblock


\bibitem[Yue et~al\mbox{.}(2022)]%
        {icml2022ntk}
\bibfield{author}{\bibinfo{person}{Kai Yue}, \bibinfo{person}{Richeng Jin},
  \bibinfo{person}{Ryan Pilgrim}, \bibinfo{person}{Chau-Wai Wong},
  \bibinfo{person}{Dror Baron}, {and} \bibinfo{person}{Huaiyu Dai}.}
  \bibinfo{year}{2022}\natexlab{}.
\newblock \showarticletitle{Neural Tangent Kernel Empowered Federated
  Learning}. In \bibinfo{booktitle}{\emph{ICML}}.
\newblock


\bibitem[Zhang et~al\mbox{.}(2020b)]%
        {bigdata2020fairfl}
\bibfield{author}{\bibinfo{person}{Daniel~Yue Zhang}, \bibinfo{person}{Ziyi
  Kou}, {and} \bibinfo{person}{Dong Wang}.} \bibinfo{year}{2020}\natexlab{b}.
\newblock \showarticletitle{Fairfl: A fair federated learning approach to
  reducing demographic bias in privacy-sensitive classification models}. In
  \bibinfo{booktitle}{\emph{IEEE Big Data}}.
\newblock


\bibitem[Zhang et~al\mbox{.}(2020c)]%
        {zhang2020fedca}
\bibfield{author}{\bibinfo{person}{Fengda Zhang}, \bibinfo{person}{Kun Kuang},
  \bibinfo{person}{Zhaoyang You}, {and} \bibinfo{person}{et al.}}
  \bibinfo{year}{2020}\natexlab{c}.
\newblock \showarticletitle{Federated unsupervised representation learning}.
\newblock \bibinfo{journal}{\emph{arXiv preprint arXiv:2010.08982}}
  (\bibinfo{year}{2020}).
\newblock


\bibitem[Zhang et~al\mbox{.}(2018)]%
        {iclr2018mixup}
\bibfield{author}{\bibinfo{person}{Hongyi Zhang}, \bibinfo{person}{Moustapha
  Cisse}, \bibinfo{person}{Yann~N Dauphin}, {and} \bibinfo{person}{David
  Lopez-Paz}.} \bibinfo{year}{2018}\natexlab{}.
\newblock \showarticletitle{mixup: Beyond Empirical Risk Minimization}. In
  \bibinfo{booktitle}{\emph{ICLR}}.
\newblock


\bibitem[Zhang et~al\mbox{.}(2020a)]%
        {iot2020poisongan}
\bibfield{author}{\bibinfo{person}{Jiale Zhang}, \bibinfo{person}{Bing Chen},
  \bibinfo{person}{Xiang Cheng}, \bibinfo{person}{Huynh Thi~Thanh Binh}, {and}
  \bibinfo{person}{Shui Yu}.} \bibinfo{year}{2020}\natexlab{a}.
\newblock \showarticletitle{Poisongan: Generative poisoning attacks against
  federated learning in edge computing systems}.
\newblock \bibinfo{journal}{\emph{IEEE IoT}} (\bibinfo{year}{2020}).
\newblock


\bibitem[Zhang et~al\mbox{.}(2021a)]%
        {nips2021ktpfl}
\bibfield{author}{\bibinfo{person}{Jie Zhang}, \bibinfo{person}{Song Guo},
  \bibinfo{person}{Xiaosong Ma}, \bibinfo{person}{Haozhao Wang},
  \bibinfo{person}{Wenchao Xu}, {and} \bibinfo{person}{Feijie Wu}.}
  \bibinfo{year}{2021}\natexlab{a}.
\newblock \showarticletitle{Parameterized Knowledge Transfer for Personalized
  Federated Learning}.
\newblock \bibinfo{journal}{\emph{NeurIPS}}.
\newblock


\bibitem[Zhang et~al\mbox{.}(2022c)]%
        {icml2022fedlc}
\bibfield{author}{\bibinfo{person}{Jie Zhang}, \bibinfo{person}{Zhiqi Li},
  \bibinfo{person}{Bo Li}, \bibinfo{person}{Jianghe Xu},
  \bibinfo{person}{Shuang Wu}, \bibinfo{person}{Shouhong Ding}, {and}
  \bibinfo{person}{Chao Wu}.} \bibinfo{year}{2022}\natexlab{c}.
\newblock \showarticletitle{Federated Learning with Label Distribution Skew via
  Logits Calibration}. In \bibinfo{booktitle}{\emph{ICML}}.
\newblock


\bibitem[Zhang et~al\mbox{.}(2022f)]%
        {zhang2022learnable}
\bibfield{author}{\bibinfo{person}{Junwu Zhang}, \bibinfo{person}{Mang Ye},
  {and} \bibinfo{person}{Yao Yang}.} \bibinfo{year}{2022}\natexlab{f}.
\newblock \showarticletitle{Learnable Privacy-Preserving Anonymization for
  Pedestrian Images}.
\newblock \bibinfo{journal}{\emph{arXiv preprint arXiv:2207.11677}}
  (\bibinfo{year}{2022}).
\newblock


\bibitem[Zhang et~al\mbox{.}(2022e)]%
        {cvpr2022fedftg}
\bibfield{author}{\bibinfo{person}{Lin Zhang}, \bibinfo{person}{Li Shen},
  \bibinfo{person}{Liang Ding}, \bibinfo{person}{Dacheng Tao}, {and}
  \bibinfo{person}{Ling-Yu Duan}.} \bibinfo{year}{2022}\natexlab{e}.
\newblock \showarticletitle{Fine-tuning global model via data-free knowledge
  distillation for non-iid federated learning}. In
  \bibinfo{booktitle}{\emph{CVPR}}.
\newblock


\bibitem[Zhang et~al\mbox{.}(2021b)]%
        {iclr2021fedfomo}
\bibfield{author}{\bibinfo{person}{Michael Zhang}, \bibinfo{person}{Karan
  Sapra}, \bibinfo{person}{Sanja Fidler}, \bibinfo{person}{Serena Yeung}, {and}
  \bibinfo{person}{Jose~M Alvarez}.} \bibinfo{year}{2021}\natexlab{b}.
\newblock \showarticletitle{Personalized federated learning with first order
  model optimization}. In \bibinfo{booktitle}{\emph{ICLR}}.
\newblock


\bibitem[Zhang et~al\mbox{.}(2021d)]%
        {access2021csfedavg}
\bibfield{author}{\bibinfo{person}{Wenyu Zhang}, \bibinfo{person}{Xiumin Wang},
  \bibinfo{person}{Pan Zhou}, \bibinfo{person}{Weiwei Wu}, {and}
  \bibinfo{person}{Xinglin Zhang}.} \bibinfo{year}{2021}\natexlab{d}.
\newblock \showarticletitle{Client selection for federated learning with
  non-iid data in mobile edge computing}.
\newblock \bibinfo{journal}{\emph{IEEE Access}} (\bibinfo{year}{2021}).
\newblock


\bibitem[Zhang et~al\mbox{.}(2022a)]%
        {icml2022cefedavg}
\bibfield{author}{\bibinfo{person}{Xinwei Zhang}, \bibinfo{person}{Xiangyi
  Chen}, \bibinfo{person}{Mingyi Hong}, \bibinfo{person}{Steven Wu}, {and}
  \bibinfo{person}{Jinfeng Yi}.} \bibinfo{year}{2022}\natexlab{a}.
\newblock \showarticletitle{Understanding Clipping for Federated Learning:
  Convergence and Client-Level Differential Privacy}. In
  \bibinfo{booktitle}{\emph{ICML}}.
\newblock


\bibitem[Zhang et~al\mbox{.}(2022b)]%
        {icml2022pfedbayes}
\bibfield{author}{\bibinfo{person}{Xu Zhang}, \bibinfo{person}{Yinchuan Li},
  \bibinfo{person}{Wenpeng Li}, \bibinfo{person}{Kaiyang Guo}, {and}
  \bibinfo{person}{Yunfeng Shao}.} \bibinfo{year}{2022}\natexlab{b}.
\newblock \showarticletitle{Personalized Federated Learning via Variational
  Bayesian Inference}. In \bibinfo{booktitle}{\emph{ICML}}.
\newblock


\bibitem[Zhang et~al\mbox{.}(2021c)]%
        {cf2021shufflefl}
\bibfield{author}{\bibinfo{person}{Yuhui Zhang}, \bibinfo{person}{Zhiwei Wang},
  \bibinfo{person}{Jiangfeng Cao}, \bibinfo{person}{Rui Hou}, {and}
  \bibinfo{person}{Dan Meng}.} \bibinfo{year}{2021}\natexlab{c}.
\newblock \showarticletitle{ShuffleFL: Gradient-preserving federated learning
  using trusted execution environment}. In \bibinfo{booktitle}{\emph{ACM CF}}.
\newblock


\bibitem[Zhang et~al\mbox{.}(2022d)]%
        {icml2022neurotoxin}
\bibfield{author}{\bibinfo{person}{Zhengming Zhang}, \bibinfo{person}{Ashwinee
  Panda}, \bibinfo{person}{Linyue Song}, \bibinfo{person}{Yaoqing Yang},
  \bibinfo{person}{Michael Mahoney}, \bibinfo{person}{Prateek Mittal},
  \bibinfo{person}{Ramchandran Kannan}, {and} \bibinfo{person}{Joseph
  Gonzalez}.} \bibinfo{year}{2022}\natexlab{d}.
\newblock \showarticletitle{Neurotoxin: Durable Backdoors in Federated
  Learning}. In \bibinfo{booktitle}{\emph{ICML}}.
\newblock


\bibitem[Zhao et~al\mbox{.}(2021)]%
        {ihmmsec2021federatedreverse}
\bibfield{author}{\bibinfo{person}{Chen Zhao}, \bibinfo{person}{Yu Wen},
  \bibinfo{person}{Shuailou Li}, \bibinfo{person}{Fucheng Liu}, {and}
  \bibinfo{person}{Dan Meng}.} \bibinfo{year}{2021}\natexlab{}.
\newblock \showarticletitle{Federatedreverse: A detection and defense method
  against backdoor attacks in federated learning}. In
  \bibinfo{booktitle}{\emph{IH\&MMSec}}.
\newblock


\bibitem[Zheng et~al\mbox{.}(2021)]%
        {ijcai2021fedmetacredit}
\bibfield{author}{\bibinfo{person}{Wenbo Zheng}, \bibinfo{person}{Lan Yan},
  \bibinfo{person}{Chao Gou}, {and} \bibinfo{person}{Fei-Yue Wang}.}
  \bibinfo{year}{2021}\natexlab{}.
\newblock \showarticletitle{Federated meta-learning for fraudulent credit card
  detection}. In \bibinfo{booktitle}{\emph{IJCAI}}.
\newblock


\bibitem[Zhu et~al\mbox{.}(2021b)]%
        {zhu2021noniidflsurvey}
\bibfield{author}{\bibinfo{person}{Hangyu Zhu}, \bibinfo{person}{Jinjin Xu},
  \bibinfo{person}{Shiqing Liu}, {and} \bibinfo{person}{Yaochu Jin}.}
  \bibinfo{year}{2021}\natexlab{b}.
\newblock \showarticletitle{Federated learning on non-IID data: A survey}.
\newblock \bibinfo{journal}{\emph{Neurocomputing}} (\bibinfo{year}{2021}).
\newblock


\bibitem[Zhu et~al\mbox{.}(2021a)]%
        {icml2021fedgen}
\bibfield{author}{\bibinfo{person}{Zhuangdi Zhu}, \bibinfo{person}{Junyuan
  Hong}, {and} \bibinfo{person}{Jiayu Zhou}.} \bibinfo{year}{2021}\natexlab{a}.
\newblock \showarticletitle{Data-free knowledge distillation for heterogeneous
  federated learning}. In \bibinfo{booktitle}{\emph{ICML}}.
\newblock


\bibitem[Zhuang et~al\mbox{.}(2020)]%
        {acmmm2020fedreid}
\bibfield{author}{\bibinfo{person}{Weiming Zhuang}, \bibinfo{person}{Yonggang
  Wen}, \bibinfo{person}{Xuesen Zhang}, \bibinfo{person}{Xin Gan},
  \bibinfo{person}{Daiying Yin}, \bibinfo{person}{Dongzhan Zhou},
  \bibinfo{person}{Shuai Zhang}, {and} \bibinfo{person}{Shuai Yi}.}
  \bibinfo{year}{2020}\natexlab{}.
\newblock \showarticletitle{Performance optimization of federated person
  re-identification via benchmark analysis}. In \bibinfo{booktitle}{\emph{ACM
  MM}}.
\newblock


\end{thebibliography}

\end{document}